%% file: main.tex
\documentclass[sigconf]{acmart}

\usepackage{graphicx}
\usepackage{amsmath}
\usepackage{amsthm}
\usepackage{booktabs}
\usepackage{algorithm}
\usepackage{algpseudocode}
\usepackage{caption}
\usepackage{subcaption}
\usepackage{xcolor}
\usepackage{amsfonts}
\usepackage{tabularx}
\usepackage{bbm}
\usepackage{enumitem}
\usepackage{multirow}
\usepackage{mathtools}
\usepackage[normalem]{ulem}

\newtheorem{definition}{Definition}
\newtheorem{theorem}{Theorem}
\newtheorem{assumption}{Assumption}

\newcommand{\sysname}{FFML}

\makeatletter
\def\@opargbegintheorem#1#2#3{\trivlist
   \item[]{\bfseries #1\ #2\ (#3)} \itshape}
\makeatother

\makeatletter
\newcommand{\multiline}[1]{%
  \begin{tabularx}{\dimexpr\linewidth-\ALG@thistlm}[t]{@{}X@{}}
    #1
  \end{tabularx}
}
\makeatother

\newcommand{\dtoprule}{\specialrule{1pt}{0pt}{0.4pt}%
            \specialrule{0.3pt}{0pt}{\belowrulesep}%
            }
\newcommand{\dbottomrule}{\specialrule{0.3pt}{0pt}{0.4pt}%
            \specialrule{1pt}{0pt}{\belowrulesep}%
            }

\AtBeginDocument{%
  \providecommand\BibTeX{{%
    \normalfont B\kern-0.5em{\scshape i\kern-0.25em b}\kern-0.8em\TeX}}}

\acmPrice{15.00}

\copyrightyear{2021} 
\acmYear{2021} 
\setcopyright{rightsretained} 
\acmConference[KDD '21]{Proceedings of the 27th ACM SIGKDD Conference on Knowledge Discovery and Data Mining}{August 14--18, 2021}{Virtual Event, Singapore}
\acmBooktitle{Proceedings of the 27th ACM SIGKDD Conference on Knowledge Discovery and Data Mining (KDD '21), August 14--18, 2021, Virtual Event, Singapore}
\acmDOI{10.1145/3447548.3467389}
\acmISBN{978-1-4503-8332-5/21/08}

\begin{document}
\fancyhead{}

\title{Fairness-Aware Online Meta-learning}


% \author{anonymous}
% \email{anonymous}
% \affiliation{%
%   \institution{anonymous}
%   \streetaddress{}
%   \city{anonymous}
%   \state{}
%   \country{}
%   \postcode{}
% }

\author{Chen Zhao}
\email{chen.zhao@utdallas.edu}
\affiliation{%
  \institution{The University of Texas at Dallas}
  \streetaddress{800 W Campbell Rd}
  \city{Richardson}
  \state{Texas}
  \country{USA}
  \postcode{75080}
}

\author{Feng Chen}
\email{feng.chen@utdallas.edu}
\affiliation{%
  \institution{The University of Texas at Dallas}
  \streetaddress{800 W Campbell Rd}
  \city{Richardson}
  \state{Texas}
  \country{USA}
  \postcode{75080}
}

\author{Bhavani Thuraisingham}
\email{bhavani.thuraisingham@utdallas.edu}
\affiliation{%
  \institution{The University of Texas at Dallas}
  \streetaddress{800 W Campbell Rd}
  \city{Richardson}
  \state{Texas}
  \country{USA}
  \postcode{75080}
}


% \renewcommand{\shortauthors}{Chen, et al.}

\begin{abstract}
    \input{abstract}

\end{abstract}




\maketitle

% \vspace{-8mm}
\section{Introduction}
    \input{intro}

% \vspace{-3mm}
\section{Related Work}
    \input{relatedworks}

\section{Preliminaries}
    \input{preliminaries}

% \vspace{-3mm}
\section{Methodology}
    \input{keypart}

\section{Analysis}
\label{sec:analysis}
    \input{analysis}

% \section{Fairness-Aware Constraints}
%     \input{fairness}

\section{Experiments}
\label{sec:experiments}
    \input{experiments}

\section{Results}
\label{sec:results}

\input{results}

\section{Conclusion and Future Work}
    \input{conclusion}

\section*{Acknowledgement}
This work is supported by the National Science Foundation (NSF) under Grant Number \#1815696 and \#1750911.

% \newpage
\bibliographystyle{ACM-Reference-Format}
\bibliography{kdd2021}

\newpage
\newpage
\appendix

\section{Sketch of Proof of Theorem \ref{theorem1}}
\label{App:proofT1}
    \input{App_proofT1}

% \section{Proof of Lemma \ref{lemma}}
% \label{App:proofLemma}
%     \input{App_proofLemma}

\section{Sketch of Proof of Theorem \ref{theorem2}}
\label{App:proofT2}
    \input{App_proofT2}
    
\section{Additional Experiment Details}
\label{App:ExpDetails}
    \input{App_ExpDetails}

\end{document}

%% file: abstract.tex
% To enhance learning of unseen tasks, researches improve intelligent system by building experience and train machine learning models from historical tasks. 
In contrast to offline working fashions,
% in which a model is trained based on entire data at once
% , online learning are devised to learn models incrementally from data in a sequential manner. There are two distinct research paradigms that have attracted considerable attention. 
two research paradigms are devised for online learning:
(1) Online Meta Learning (OML) \cite{Finn-ICML-2019,Yao-2020-NeurIPS,Anusha-2019-ICLR}
learns good priors over model parameters (or learning to learn) in a sequential setting where tasks are revealed one after another.
% It extended the meta-learning setting onto online learning. 
Although it provides a sub-linear regret bound, such techniques completely ignore the importance of learning with fairness which is a significant hallmark of human intelligence. (2) Online Fairness-Aware Learning \cite{Bechavod-2019-NeurIPS,Stephen-2018-NeurIPS,Vishakha-2020-AAAI}. This setting captures many classification problems for which fairness is a concern. 
But it aims to attain zero-shot generalization without any task-specific adaptation.
% assumes that past experience is not available to be used to acquire a prior over model parameters or a learning-to-learn procedure at each round. 
This therefore limits the capability of a model to adapt onto newly arrived data. To overcome such issues and bridge the gap, in this paper for the first time we proposed a novel online meta-learning algorithm, namely \sysname{}, which is under the setting of unfairness prevention. The key part of \sysname{} is to 
learn good priors of an online fair classification model’s primal and dual parameters  that are associated with the model's accuracy and fairness, respectively.
% \sout{determine a good primal-dual parameter-pair in stead of find a single parameter at each round.} 
The problem is formulated in the form of a bi-level convex-concave optimization. Theoretic analysis provides sub-linear upper bounds $O(\log T)$ for loss regret and $O(\sqrt{T\log T})$ for violation of cumulative fairness constraints. Our experiments demonstrate the versatility of \sysname{} by applying it to classification on three real-world datasets and show substantial improvements over the best prior work on the tradeoff between fairness and classification accuracy.

%% file: intro.tex
% \begin{figure}[!t]
%     \centering
%     \includegraphics[width=\linewidth]{others/motivation-KDD.png}
%     \vspace{-8mm}
%     \caption{An example of bank loan approval during COVID-19 pandemic is given where a sequence of data batches from a non-stationary distribution are collected one after another over timeline.
%     % Filled black/white represents protected attributes (\textit{i.e.} race) and blue/red indicates positive/negative decisions. Different shapes indicate non-stationary data distribution. 
%     In each batch, decision-makings are biased on the protected attribute (\textit{e.g.} race). Our goal is to learn a fair model online that it efficiently returns predictions with bias-control and quickly adapts to unseen data.
%     } 
%     \label{fig:motivation}
%     \vspace{-5mm}
% \end{figure}

It’s no secret that bias is present everywhere in our society, such as  recruitment, loan qualification, recidivism, \textit{etc}. The manifestation of bias can be often as fraught as race or gender. Because of this, machine learning algorithms have several examples of training models, many of which have received strong criticism for exhibiting unfair bias. 
% For example, 
% % \cite{Barr-Google-Gorillas-2015} reported that a picture of two African Americans was automatically tagged as ``Gorillas'' by \textit{Google Photos}. 
% a 2016 study \cite{Ingold-Amazon-2016} found the data-driven system developed by \textit{Amazon} 
% % that used to determine the neighborhoods in which to offer free same-day delivery 
% is highly biased and unfair to African American communities due to the stark disparities in the demographic makeup of neighborhoods. 
Critics have voiced that human bias potentially has an influence on nowadays technology, which leads to outcomes with unfairness.

With biased input, the main goal of training an unbiased model in machine learning is to make the output fair. Group-fairness, also known as statistic parity, ensures the equality of a predictive utility across different sub-populations. In other words, the predictions are statistically independent on protected variables, \textit{e.g.}, race or gender. In the real world, data with bias are likely available only sequentially and also from a non-stationary task distribution. 
For example, a recent news \cite{Miller-2020-NYTimes} by \textit{New York Times} reports that systematic algorithms become increasingly discriminative to African Americans in bank loan during COVID-19 pandemic. These algorithms are built up from a sequence of data batches collected one after another over time, where in each batch, decision-makings are biased on the protected attribute (\textit{e.g.} race). 
To learn a fair model over time and make it efficiently and quickly adapt to unseen data,
online learning \cite{Hannan-1957-online} are devised to learn models incrementally from data in a sequential manner and models can be updated instantly and efficiently when new training data arrives \cite{Hoi-2018-survey}.
 
Two distinct research paradigms in online learning have attracted attentions in recent years. Online meta-learning \cite{Finn-ICML-2019} learns priors over model parameters in a sequential setting not only to master the batch of data at hand but also the learner becomes proficient with quick adaptation at learning new arrived tasks in the future. Although such techniques achieve sub-linear loss regret, it completely ignores the significance of learning with fairness, which is a crucial hallmark of human intelligence.

On the other hand, fairness-aware online learning captures supervised learning problems for which fairness is a concern. It either compels the algorithms satisfy common fairness constraints at each round \cite{Bechavod-2019-NeurIPS} or defines a fairness-aware loss regret where learning and fairness interplay with each other \cite{Vishakha-2020-AAAI}. However, neither of these settings is ideal for studying continual lifelong learning where past experience is used to learn priors over model parameters, and hence existing methods lack adaptability to new tasks.

With the aim of connecting the fields of online fairness-aware learning and online meta-learning, we introduce a new problem statement, that is \textit{fairness-aware online meta-learning with long-term constraints}, where the definition of long-term constraints \cite{OGDLC-2012-JMLR} indicates the sum of cumulative fairness constraints. From a global perspective, we allow the learner to make decisions at some rounds which may not belong to the fairness domain due to non-stationary aspects of the problem, but the overall sequence of chosen decisions must obey the fairness constraints at the end by a vanishing convergence rate.

To this end, technically we propose a novel online learning algorithm, namely \textit{follow-the-fair-meta-leader} (\sysname{}). At each round, we determine model parameters by formulating a problem composed by two main levels: online fair task-level learning and meta-level learning. Each level of the bi-level problem is embedded within each other with two parts of parameters: primal parameters $\boldsymbol{\theta}$ regarding model accuracy and dual parameters $\boldsymbol{\lambda}$ adjusting fairness notions.
% by updating a primal-dual meta-pair constrained by fairness notions at each round. 
% We hence refer this new optimization problem as \textit{online meta-learning with long-term constraints}, where the definition of long-term constraint is first proposed
% as the sum of constraints 
% in \cite{OGDLC-2012-JMLR}.
% More concretely, at each round $t$, we see each sub-problem as a constrained bi-level optimization problem where an inner-level optimization is embedded within an outer-level one.
Therefore, in stead of learning primary parameters only at round $t\in[T]$, an agent learns a meta-solution pair $(\boldsymbol{\theta}_{t+1},\boldsymbol{\lambda}_{t+1})$ across all existing tasks by optimizing a convex-concave problem and extending the gradient based approach for variational inequality. Furthermore, when a new task arrives at $t+1$, the primal variable $\boldsymbol{\theta}_{t+1}$ and the dual variable $\boldsymbol{\lambda}_{t+1}$ are able to quickly adapted to it and the overall model regret grows sub-linearly in $T$.
% respectively respond for adjusting accuracy and fairness level. 
% An overview of \sysname{} is presented in Figure \ref{fig:overview}.
We then analyze \sysname{} with theoretic proofs demonstrating it enjoys a $O(\log T)$ regret guarantee and $O(\sqrt{T\log T})$ bound for violation of long-term fairness constraints when competing with the best meta-learner in hindsight.
The main contributions are summarized:
\begin{itemize}[leftmargin=*]
    % \vspace{-2.3mm}
    \item To the best of our knowledge, for the first time a fairness-aware online meta-learning problem is proposed. To solve the problem efficiently, we propose a novel algorithm \textit{follow-the-fair-meta-leader} (\sysname{}). Specifically, at each time, the problem is formulated as a constrained bi-level convex-concave optimization with respect to a primal-dual parameter pair for each level, where the parameter pair responds for adjusting accuracy and fairness notion adaptively.
    \item Theoretically grounded analysis justifies the efficiency and effectiveness of the proposed method by demonstrating a $O(\log T)$ bound for loss regret and $O(\sqrt{T\log T})$ for violation of fairness constraints, respectively. 
    \item We validate the performance of our approach with state-of-the-art techniques on real-world datasets. Our results demonstrate the proposed approach is not only capable of mitigating biases but also achieves higher efficiency compared with the state-of-the-art algorithms.
\end{itemize}

%% file: relatedworks.tex
\textbf{Fairness-aware online learning} problems assume individuals arrive one at a time and the goal of such algorithms is to train predictive models free from biases. 
% Standard predictive models may discriminate groups of entities because (1) data bias comes from data being collected from different sources, or (2) dependence on sensitive attributes was identified in the data mining community. 
% Based on the taxonomy by tasks, fairness-aware in terms of equality across different sub-groups can be typically categorized to classification \cite{Feldman-KDD-2015,Hardt-NIPS-2016,Zafar-AISTATS-2017}, regression \cite{Calders-ICDM-2013,Berk-FATML-2018,Zhao-ICDM-2019}, clustering \cite{Gondek-KDD-2005}, and recommendation \cite{singh2018} works.
To require fairness guarantees at each round, \cite{Bechavod-2019-NeurIPS} has partial and bandit feedback and makes distributional assumptions. Due to the the trade-off between the loss regret and fairness in terms of an unfairness tolerance parameter, a fairness-aware regret \cite{Vishakha-2020-AAAI} is devised 
and it provides a fairness guarantee held uniformly over time. Besides, in contrast to group fairness, online individual bias is governed by an unknown similarity metric \cite{Stephen-2018-NeurIPS}.
% Although techniques for unfairness prevention were well developed,
However, these methods are not ideal for continual lifelong learning with non-stationary task distributions, as they aim to obtain zero-shot generalization but fail to learn priors from past experience to support any task-specific adaptation.
% \sout{However, to the best of our knowledge, non of existing fairness-aware machine learning algorithms are ideal for studying continual lifelong learning where a sequence of batches of samples arrive one after another and past experience is important to learn priors over model parameters to improve adaptability to non-stationary new tasks.}

\textbf{Meta-learning} \cite{schmidhuber-1987-srl} addresses the issue of learning with fast adaptation, where a meta-learner learns knowledge transfer from history tasks onto unseen ones.
% Several recent approaches have made significant progress in meta-learning, which 
Approaches could be broadly classified into offline and online paradigms. In the offline fashion, existing meta-learning methods  generally assume that the tasks come from some fixed distribution \cite{Finn-NIPS-2018,Rusu-ICLR-2019,zhao-2020-pdfm,Zhao-ICDM-2019,Zhao-ICKG-1-2020,Zhao-ICKG-2-2020,wang-2021-WWW}, whereas it is more realistic that methods are expected to work for non-stationary task distributions.
% nearest neighbors based methods \cite{Vinyals-NIPS-2016-(MatchingNet),Snell-NIPS-2017-(ProtoNet)}, recurrent network based methods \cite{Ravi-ICLR-2017}, and and gradient-based methods \cite{Finn-NIPS-2018,Rusu-ICLR-2019,iMAML-2019-NIPS,zhao-2020-pdfm}. 
% Despite their early success in the few-shot classification, little attention has been paid to studies of how fairness can be generalized to new tasks. \cite{zhao-2020-pdfm} extended from MAML \cite{Finn-ICML-2017-(MAML)} controls bias and further ensures fairness adaption in meta-testing phase, which resorts to duality theories in optimization. 
% Existing meta-learning methods generally assume that the tasks come from some fixed distribution, whereas it is more realistic that methods are expected to work for non-stationary task distributions. 
To this end, FTML \cite{Finn-ICML-2019} can be considered as an application of MAML \cite{Finn-ICML-2017-(MAML)} in the setting of online learning. OSML \cite{Yao-2020-NeurIPS} disentangles the whole meta-learner as a meta-hierarchical graph with multiple structured knowledge blocks. As for reinforcement learning, MOLe \cite{Anusha-2019-ICLR} used expectation maximization to learn mixtures of neural network models. A major drawback of aforementioned methods is that it immerses in minimizing objective functions but ignores the fairness of prediction. 
In our work, we deal with online meta learning subject to group fairness constraints and formulate the problem as a constrained bi-level optimization problem, 
in which the proposed algorithm enjoys both regret guarantee and an upper bound on violation of fairness constraints.

% \textbf{Online learning} is a paradigm in which an agent continually learns as it is interacting with the world and the goal is to generate a sequence of parameters at each round to minimize the cumulative regret \cite{Hoi-2018-survey}. 
% For online convex optimization with constraints, when the constraints are complex, the computational burden of the projection may be too high. To circumvent this dilemma, \cite{OGDLC-2012-JMLR} refers the problem as \textbf{online learning with long-term constraints} and relaxes the output through a simpler close-form projection. Furthermore, several close works aim to improve the bound of regret guarantee and violation of constraints by proposing an adaptive stepsize version \cite{AdpOLC-2016-ICML}, stochastic
% constraints \cite{Yu-2017-NIPS}, clipping constraints into non-negative orthant \cite{GenOLC-2018-NeurIPS}. Although aforementioned techniques achieve state-of-the-art theoretic guarantees, they do not generally consider how past experience can accelerate adaptation to a new task. Again, in this paper, we study group fairness by casting it as an optimization problem of online meta-learning with long-term fairness constraints. We allow the learner to make decisions at some iterations which may not satisfy the fairness domain, but the overall sequence of chosen decisions must obey the constraints at the end by a vanishing convergence rate.

\textbf{Online convex optimization with long term constra- ints.} For online convex optimization with long-term constraints, a projection operator is typically applied to update parameters to make them feasible at each round. However, when the constraints are complex, the computational burden of the projection may be too high. To circumvent this dilemma, \cite{OGDLC-2012-JMLR} relaxes the output through a simpler close-form projection. Thereafter, several close works aim to improve the theoretic guarantees
% bound of regret guarantee and violation of constraints 
by modifying stepsizes to a adaptive version \cite{AdpOLC-2016-ICML}, adjusting to stochastic constraints \cite{Yu-2017-NIPS}, and clipping constraints into a non-negative orthant \cite{GenOLC-2018-NeurIPS}. 
Although such techniques achieve state-of-the-art theoretic guarantees, 
%for solving single-level online convex optimization problems, 
they are not directly applicable to bi-level online convex optimization with long-term constraints.   

In this paper, we study the problem of online fairness-aware meta learning to deal with non-stationary task distributions. Our proposed approach is designed based on the bridging of the above three areas. In particular, we connect the first two areas to formulate the problem as a bi-level online convex optimization problem with long-term fairness constraints, and develop a novel learning algorithm based on generalization of the primal-dual optimization techniques designed in the third area. 

% Although such techniques achieve state-of-the-art theoretic guarantees, they \feng{can only address single-level online convex optimization problems} and do not generally consider how past experience can accelerate adaptation to a new task. 
%\feng{(or learning to learn)}. 
%As a result, these techniques are designed to solve 
%Again, 
% In this paper, we study group fairness by casting it as a bi-level optimization problem of online meta-learning with long-term fairness constraints. 
% To address, in this paper, model parameters are chosen through additional task-level update procedures. The problem hence takes the form of a bi-level optimization of online meta-learning with long-term fairness constraints. We allow the learner to make decisions at some iterations which may not satisfy the fairness domain, but the overall sequence of chosen decisions must obey the constraints at the end by a vanishing convergence rate.

% \vspace{-5mm}

%% file: preliminaries.tex
% We begin by introducing some basic mathematical terms and notations, and then define our problem.
\subsection{Notations}
An index set of a sequence of tasks is defined as $[T]=\{1,...,T\}$. Vectors are denoted by lower case bold face letters, \textit{e.g.} the primal variables $\boldsymbol{\theta}\in\Theta$ and the dual variables $\boldsymbol{\lambda}\in\mathbb{R}^m_+$ where their $i$-th entries are $\theta_i,\lambda_i$. Vectors with subscripts of task indices, such as $\boldsymbol{\theta}_t,\boldsymbol{\lambda}_t$ where $t\in[T]$, indicate model parameters for the task at round $t$. The Euclidean $\ell_2$-norm of $\boldsymbol{\theta}$ is denoted as $||\boldsymbol{\theta}||$. Given a differentiable function $\mathcal{L}(\boldsymbol{\theta,\lambda}):\Theta\times\mathbb{R}^m_+\rightarrow\mathbb{R}$, the gradient at $\boldsymbol{\theta}$ and $\boldsymbol{\lambda}$ is denoted as $\nabla_{\boldsymbol{\theta}} \mathcal{L}(\boldsymbol{\theta,\lambda})$ and $\nabla_{\boldsymbol{\lambda}} \mathcal{L}(\boldsymbol{\theta,\lambda})$, respectively. Scalars are denoted by lower case italic letters, \textit{e.g.} $\eta>0$. Matrices are denoted by capital italic letters. 
% The base learner is a mapping  $\mathcal{A}lg(\cdot):\boldsymbol{\theta}\in\mathcal{B}\rightarrow\boldsymbol{\theta}'\in\mathbb{R}^d$. 
$\prod_\mathcal{B}$ is the projection operation to the set $\mathcal{B}$. $[\boldsymbol{u}]_+$ denotes the projection of the vector $\boldsymbol{u}$ on the nonnegative orthant in $\mathbb{R}^m_+$, namely $[\boldsymbol{u}]_+ = (\max\{0,u_1\}),...,\max\{0,u_m\})$. An example of a function $f(\cdot)$ taking two variables $\boldsymbol{\theta,\phi}$ separated with a semicolon (\textit{i.e.} $f(\boldsymbol{\theta;\phi})$) indicates that $\boldsymbol{\theta}$ is initially assigned with $\boldsymbol{\phi}$. Some important notations are listed in Table \ref{tab:notation}.

\begin{table}[!t]
% \footnotesize
    \centering
    \caption{Important notations and corresponding descriptions.}
    \begin{tabular}{l|p{6cm}}
        \hline
        \textbf{Notations} & \textbf{Descriptions}  \\
        \hline
        $T$ & Total number of learning tasks\\
        % \hline
        $t$ & Indices of tasks\\
        % \hline
        $\mathcal{D}^S, \mathcal{D}^V, \mathcal{D}^Q$ & Support set, validation set, and query set of data $\mathcal{D}$\\
        % \hline
        $\boldsymbol{\theta, \lambda}$ & Meta primal and dual parameters\\
        % \hline
        $\boldsymbol{\theta}_t, \boldsymbol{\lambda}_t$ & Model primal and dual parameters of task $t$\\
        % \hline
        $f_t(\cdot)$ & Loss function of at round $t$  \\
        % \hline
        $g(\cdot)$ & Fairness function \\
        % \hline
        $m$ & Total number of fairness notions\\
        % \hline
        $i$ & Indices of fairness notions\\
        % \hline
        $\mathcal{A}lg(\cdot)$ & Base learner\\
        % \hline
        $\mathcal{U}$ & Task buffer\\
        % \hline
        $k$ & Indices of past tasks in $\mathcal{U}$\\
        % \hline
        $\mathcal{B}$ & Relaxed primal domain \\
        % \hline
        $\prod_{\mathcal{B}}$ & Projection operation to domain $\mathcal{B}$\\
        % \hline
        $\eta_1, \eta_2$ & Learning rates\\
        % \hline
        $\delta$ & Augmented constant\\
        \hline
        
    \end{tabular}
    \label{tab:notation}
% \vspace{-5mm}
\end{table}

\subsection{Fairness-Aware Constraints}
Intuitively, an attribute affects the target variable if one depends on the other. Strong dependency indicates strong effects. In general, group fairness criteria used for evaluating and designing machine learning models focus on the relationships between the protected attribute and the system output. The problem of group unfairness prevention can be seen as a constrained optimization problem. For simplicity, we consider one binary protected attribute (\textit{e.g.} white and black) in this work. However, our ideas can be easily extended to many protected attributes with multiple levels.

Let $\mathcal{Z=X\times Y}$ be the data space, where $\mathcal{X} = \mathcal{E} \cup\mathcal{S}$. Here $\mathcal{E} \subset \mathbb{R}^d$ is an input space, $\mathcal{S} = \{0,1\}$ is a protected space, and $\mathcal{Y} = \{0,1\}$ is an output space for binary classification. 
Given a task (batch) of samples
% At each round $t\in[T]$, for each task $k\in[t-1]$, we let 
$\{\mathbf{e}_{i}, y_{i}, s_{i}\}_{i=1}^n \in(\mathcal{E\times Y\times S})$ 
% be the corresponding task data and 
where $n$ is the number of datapoints, 
% In referencing fairness, we are using same amount of training examples irrespective of class label, with the assumption that all tasks are 2-way. 
a fine-grained measurement to ensure fairness in class label prediction is to design fair classifiers by controlling the decision boundary covariance (DBC) \cite{Zafar-AISTATS-2017}.
\begin{definition}[Decision Boundary Covariance \cite{Zafar-AISTATS-2017,Lohaus-2020-ICML}]
The Decision Boundary Covariance (i.e. DBC) is defined as the covariance between the protected variables $\mathbf{s}=\{s_i\}_{i=1}^n$ and the signed distance from the feature vectors to the decision boundary.
% $d_{\boldsymbol{\theta}} (\mathbf{e}) = \{d_{\boldsymbol{\theta}} (\mathbf{e}_i)\}_{i=1}^n$, 
% where $\mathbf{\alpha}$ is the decision boundary parameter.
% To get rid of indicator functions, 
A linear approximated form of DBC takes
% in \cite{Lohaus-2020-ICML},
\begin{align}
\label{dbc definition}
    \text{DBC} = 
    % (\mathbf{s}, d_{\boldsymbol{\theta}}(\mathbf{e})) = \mathbb{E} \big[ (\mathbf{s}-\mathbf{\Bar{s}})d_{\boldsymbol{\theta}}(\mathbf{e}) \big]
    % \frac{1}{n}\sum_{\{\mathbf{e}_{i}, y_{i}, s_{i}\}_{i=1}^n \in(\mathcal{E\times Y\times S})}
    \mathbb{E}_{(\mathbf{e},y,s)\in\mathcal{Z}}\Big[\frac{1}{\hat{p}_1(1-\hat{p}_1)}\Big(\frac{s+1}{2}-\hat{p}_1\Big)h(e,\boldsymbol{\theta})\Big]
\end{align}
where 
% $\mathbf{\Bar{s}} = (\frac{1}{n}\sum_{i=1}^n s_i)\mathbbm{1}$.
$\hat{p}_1$ is an empirical estimate of $p_1$ and $p_1=\mathbb{P}_{(\mathbf{e},y,s)\in\mathcal{Z}}(s=1)$ is the proportion of samples in group $s=1$, and $h:\mathbb{R}^d\times\Theta\rightarrow\mathbb{R}$ is a real valued function taking $\boldsymbol{\theta}$ as parameters.
\end{definition}
Therefore, parameters $\boldsymbol{\theta}$ in the domain of a task is feasible if it satisfies the fairness constraint $g(\boldsymbol{\theta})\leq 0$. More concretely, $g(\boldsymbol{\theta})$ is defined by DBC in Eq.(\ref{dbc definition}), \textit{i.e.}
\begin{align}
\label{eq:DBC constraint}
    g(\boldsymbol{\theta}) = \Big|\text{DBC} \Big| -\epsilon
\end{align}
where $|\cdot|$ is the absolute function and $\epsilon>0$ is the fairness relaxation determined by empirical analysis. 
% To formalize the supervised classification problem in the context of meta-learning definitions, a cross-entropy loss function is used to describe the adapted loss over a support set for each task.

% \sout{In general, fairness-aware machine learning can be sub-categorized into group fairness \cite{Calders-ICDM-2013,zhao-2020-pdfm,Zafar-AISTATS-2017} and individual fairness \cite{Zemel-ICML-2013,Dwork-2011-CoRR} problems. In this paper, we only focus on the group fairness.} As for group fairness constraints, two wildly used fair notion families are demographic parity \cite{calders-2009-icdmw} and equality of opportunity (\textit{a.k.a} equal opportunity) \cite{Hardt-NIPS-2016}. Notice that demographic parity and equality of opportunity are quite similar from a mathematical point of view \cite{Lohaus-2020-ICML}, and hence results and analysis on one notion can often be readily extended to the other one. In this work, according to \cite{Lohaus-2020-ICML}, the fairness notion introduced in Definition \ref{dbc definition} is% 

For group fairness constraints, two wildly used fair notion families are demographic parity \cite{calders-2009-icdmw} and equality of opportunity (\textit{a.k.a} equal opportunity) \cite{Hardt-NIPS-2016}. Notice that demographic parity and equality of opportunity are quite similar from a mathematical point of view \cite{Lohaus-2020-ICML}, and hence results and analysis on one notion can often be readily extended to the other one. According to \cite{Lohaus-2020-ICML}, the DBC fairness notion introduced in Definition \ref{dbc definition} is an empirical version of demographic parity. In this paper, we consider DBC as the fairness-aware constraint, but our proposed approach supports equality of opportunity and other fairness notions that are smooth functions.

% \vspace{-7mm}
\subsection{Settings and Problem Formulation}
\label{sec:settings and prob formulation}
% A general form for online convex optimization (OCO) is as follows: at each time $t\in[T]$,  we choose a model parameter $\boldsymbol{\theta}_t$ in its convex domain $\Theta$. Then we receive a loss function $f_t:\Theta\rightarrow\mathbb{R}$ drawn from a family of convex functions which need not be a fixed distribution, and could even be chosen adversarially over time. 
The goal of online learning is to generate a sequence of model parameters $\{\boldsymbol{\theta}_t\}_{t=1}^T$ that perform well on the loss sequence $\{f_t:\Theta\rightarrow\mathbb{R}\}_{t=1}^T$, \textit{e.g.} cross-entropy for classification problems.
Here, we assume $\Theta\subseteq\mathbb{R}^d$ is a compact and convex subset of the $d$-dimensional Euclidean space with non-empty interior. In a general setting of online learning, there is no constraint on how the sequence of loss functions is generated. In particular, the standard objective is to minimize the regret of Eq.(\ref{eq:OCO-regret}) defined as the difference between the cumulative losses that have incurred over time and the best performance achievable in hindsight. The solution to it is called Hannan consistent \cite{Cambridge-book-2006} if the upper bound on the worst case regret of an algorithm is sublinear in $T$. 
% \vspace{-2mm}
\begin{align}
\label{eq:OCO-regret}
    Regret_T = \sum^T_{t=1} f_t(\boldsymbol{\theta}_t) - \min_{\boldsymbol{\theta}\in\Theta}\sum^T_{t=1} f_t(\boldsymbol{\theta})
\end{align}
% \vspace{-2.5mm}
% In online convex optimization, we are usually interested in an upper bound on the worst case regret of an algorithm. 
% The solution to Eq.(\ref{eq:OCO-regret}) is called Hannan consistent \cite{Cambridge-book-2006} if the upper bound on the worst case regret of an algorithm, $Regret_T$, is sublinear in $T$. 
% To this end, the most straightforward approach to determine $\boldsymbol{\theta}_{t+1}$ at time $t+1$ is to use at any time the optimal decision in hindsight. Formally,

% \begin{align}
%     \boldsymbol{\theta}_{t+1} = \arg\min_{\boldsymbol{\theta}\in\Theta} \sum_{k=1}^t f_k(\boldsymbol{\theta})
% \end{align}
% This flavor of strategy is known as ``Follow the Leader" (FTL, \cite{Hannan-1957-online}) in machine learning. However, it is not hard to see that this simple strategy fails miserably in a worst-case sense that the predictions of FTL may vary wildly from one iteration to the next. This motivates the modification of the basic FTL strategy in order to stabilize the prediction. 
% % By adding a regularization term, we obtain the RFTL (Regularized Follow the Leader) algorithm in this work.

To control bias and especially ensure group fairness across different sensitive sub-populations, \textbf{fairness notions are considered as constraints added on optimization problems}. A projection operator is hence typically applied to the updated variables in order to make them feasible at each round \cite{OGDLC-2012-JMLR,GenOLC-2018-NeurIPS,AdpOLC-2016-ICML}.

In this paper, we consider a general sequential setting where an agent is faced with tasks $\{\mathcal{D}_t\}_{t=1}^T$ one after another. Each of these tasks corresponds to a batch of samples from a fixed but unknown non-stationary distribution. The goal for the agent is to minimize the regret under the summation of fair constraints, namely \textit{long-term constraints}:
\begin{align}
\label{eq: problem-formulation}
    \min_{\boldsymbol{\theta}_1,...,\boldsymbol{\theta}_T\in\mathcal{B}} \quad &Regret_T = \sum_{t=1}^T f_t(\mathcal{A}lg_t(\boldsymbol{\theta}_t, \mathcal{D}_t^{S}),\mathcal{D}_t^V) \\ 
    &- \min_{\boldsymbol{\theta}\in\Theta} \quad \sum_{t=1}^T f_t(\mathcal{A}lg_t(\boldsymbol{\theta}, \mathcal{D}_t^{S}), \mathcal{D}_t^V) \nonumber\\
    \text{subject to} \quad &\sum_{t=1}^T g_i(\mathcal{A}lg_t(\boldsymbol{\theta}_t, \mathcal{D}_t^{S}), \mathcal{D}_t^V) \leq O(T^\gamma), \forall i\in[m] \nonumber
\end{align}

where $\gamma\in(0,1)$; $\mathcal{D}_t^S\subset\mathcal{D}_t$ is the support set and $\mathcal{D}^{V}_t\subseteq\mathcal{D}_t$ is a subset of task $\mathcal{D}_t$ that is used for evaluation; $\mathcal{A}lg(\cdot)$ is the base learner which corresponds to one or multiple gradient steps \cite{Finn-ICML-2017-(MAML)} of a Lagrangian function, which will be introduced in the following sections. For the sake of simplicity, we will use one gradient step gradient throughout this work, but more steps are applicable. Different from traditional online learning settings, the long-term constraint violation $\sum_{t=1}^T g_i(\cdot): \mathcal{B}\rightarrow\mathbb{R}, \forall i\in[m]$ is required to be bounded sublinear in $T$. In order to facilitate our analysis, at each round, $\boldsymbol{\theta}_t$ is originally chosen from its domain $\Theta$, where it can be written as an intersection of a finite number of convex constraints that $\Theta=\{\boldsymbol{\theta}\in\mathbb{R}^d:g_i(\boldsymbol{\theta})\leq 0, i=1,...,m\}$. In order to lower the computational complexity and accelerate the online processing speed, inspired by \cite{OGDLC-2012-JMLR}, we relax the domain to $\mathcal{B}$, where $\Theta\subseteq\mathcal{B}=R\mathbb{K}$ with $\mathbb{K}$ being the unit $\ell_2$ ball centered at the origin, and $R\triangleq \max\{r>0: r=||\mathbf{x}-\mathbf{y}||, \forall \mathbf{x},\mathbf{y}\in\Theta\}$. With such relaxation, we allow the learner to make decisions at some rounds which do not belong to the domain $\Theta$, but the overall sequence of chosen decisions must obey the constraints at the end by vanishing convergence rate. 
% Besides, this relaxation allows even when $\Theta$ is non-convex due to the convexity of $\mathcal{B}$. 
Following \cite{GenOLC-2018-NeurIPS} and \cite{Finn-ICML-2019}, we assume
% focus on a finite-horizon setting where 
the number of rounds $T$ is known in advance.

\textbf{Remarks}: 
% In order to allow the comparator to adapt each task at hand with fairness and efficiency, we propose a novel online meta-learning setting with cumulative fairness constraints where each task is embeded with a learning-to-learn procedure. 
The new regret defined in Eq.(\ref{eq: problem-formulation}) differs from the settings of online learning with long-term constraints. Minimizing Eq.(\ref{eq: problem-formulation}) is embeded with a bi-level optimization problem.
In contrast to the regret considered in FTML \cite{Finn-ICML-2019}, both loss regret and violation of long-term fairness constraints in Eq.(\ref{eq: problem-formulation}) are required to be bounded sublinearly in $T$. 

%% file: keypart.tex
In order to minimize the regret constrained with fairness notions in Eq.(\ref{eq: problem-formulation}), the overall protocol for the setting is:
\begin{itemize}[leftmargin=*]
    \item \textbf{Step 1}: At round $t$, task $t$ and model parameters defined by $\mathcal{D}_t, \boldsymbol{\theta}_t$ are chosen.
    \item \textbf{Step 2}: The learning agent incurs loss $f_t(\mathcal{A}lg_t(\boldsymbol{\theta}_t))$ and fairness $g_i(\mathcal{A}lg_t(\boldsymbol{\theta}_t)), \forall i\in[m]$.
    \item \textbf{Step 3}: The update procedure is learned from prior experience, and it is used to determine model parameters $\boldsymbol{\theta}_{t+1}$ fairly through an optimization algorithm.
    \item \textbf{Step 4}: The next predictors are updated and advance to the next round $t+1$.
\end{itemize}

\subsection{Follow the Fair Meta Leader (\sysname{})}
In the protocol, the key step is to find a good meta parameters $\boldsymbol{\theta}$ at each round (Step 3). At round $t$, when the task $\mathcal{D}_t$ comes, the main goal incurred is to determine the meta parameters $\boldsymbol{\theta}_{t+1}$ for the next round. Specifically, the most intuitive way to find a good $\boldsymbol{\theta}_{t+1}$ is to optimize it over past seen tasks from 1 to $t$. We hence consider a setting where the agent can perform some local task-specific updates to the model before it is deployed and evaluated onto each task at each round. 

The problem of learning meta parameters $\boldsymbol{\theta}$ at each round, therefore, is embedded with another optimization problem of finding model-parameters in a task-specific level. Here, the base learner $\mathcal{A}lg_k(\cdot)$ determines model-parameters such that the task loss $f_k:\Theta\rightarrow\mathbb{R}$ is minimized subject to all constraints $g_i(\boldsymbol{\theta}_k)\leq 0, i=1,2,...,m$, where $k\in[t]$ is the index of previous tasks. 

The optimization problem is formulated with two nested levels, \textit{i.e.} an outer and an inner level, and one supports another. \textit{The outer problem} takes the form:
\begin{align}
\label{eq:outer-problem}
    \boldsymbol{\theta}_{t+1}=\arg\min_{\boldsymbol{\theta}\in\mathcal{B}} \quad &\sum_{k=1}^{t} f_k(\boldsymbol{\theta}, \mathcal{D}^Q_k ;\mathcal{A}lg_k(\boldsymbol{\theta}, \mathcal{D}^S_k))\\
    \text{subject to} \quad &\sum_{k=1}^{t} g_i(\boldsymbol{\theta},\mathcal{D}_k^Q;\mathcal{A}lg_k(\boldsymbol{\theta}, \mathcal{D}_k^S)) \leq 0, \forall i\in[m] \nonumber
\end{align}
where \textit{the inner problem} is defined as:
\begin{align}
\label{eq:inner-problem}
    \mathcal{A}lg_k(\boldsymbol{\theta}, \mathcal{D}^S_k) = &\arg\min_{\boldsymbol{\theta}_k\in\mathcal{B}} \quad f_k(\boldsymbol{\theta}_k, \mathcal{D}_k^S; \boldsymbol{\theta}) \\
    &\text{subject to}\quad g_i(\boldsymbol{\theta}_k, \mathcal{D}^S_k;\boldsymbol{\theta})\leq 0, \forall i\in[m] \nonumber
\end{align}

where $\mathcal{D}_k^S,\mathcal{D}_k^Q\subset\mathcal{D}_k$ are support and query mini-batches, which are independently sampled without replacements, \textit{i.e.} $\mathcal{D}_k^S\cap\mathcal{D}_k^Q=\emptyset$. 
In the following section, we introduce our proposed algorithm. In stead of optimizing primal parameters only, it efficiently deals with the bi-level optimization problem of Eq.(\ref{eq:outer-problem})(\ref{eq:inner-problem}) by approximating a sequence of a pair of primal-dual meta parameters $(\boldsymbol{\theta,\lambda})$ where the pair respectively responds for adjusting accuracy and fairness level.

\begin{figure}[t!]
    \centering
    \includegraphics[width=\linewidth]{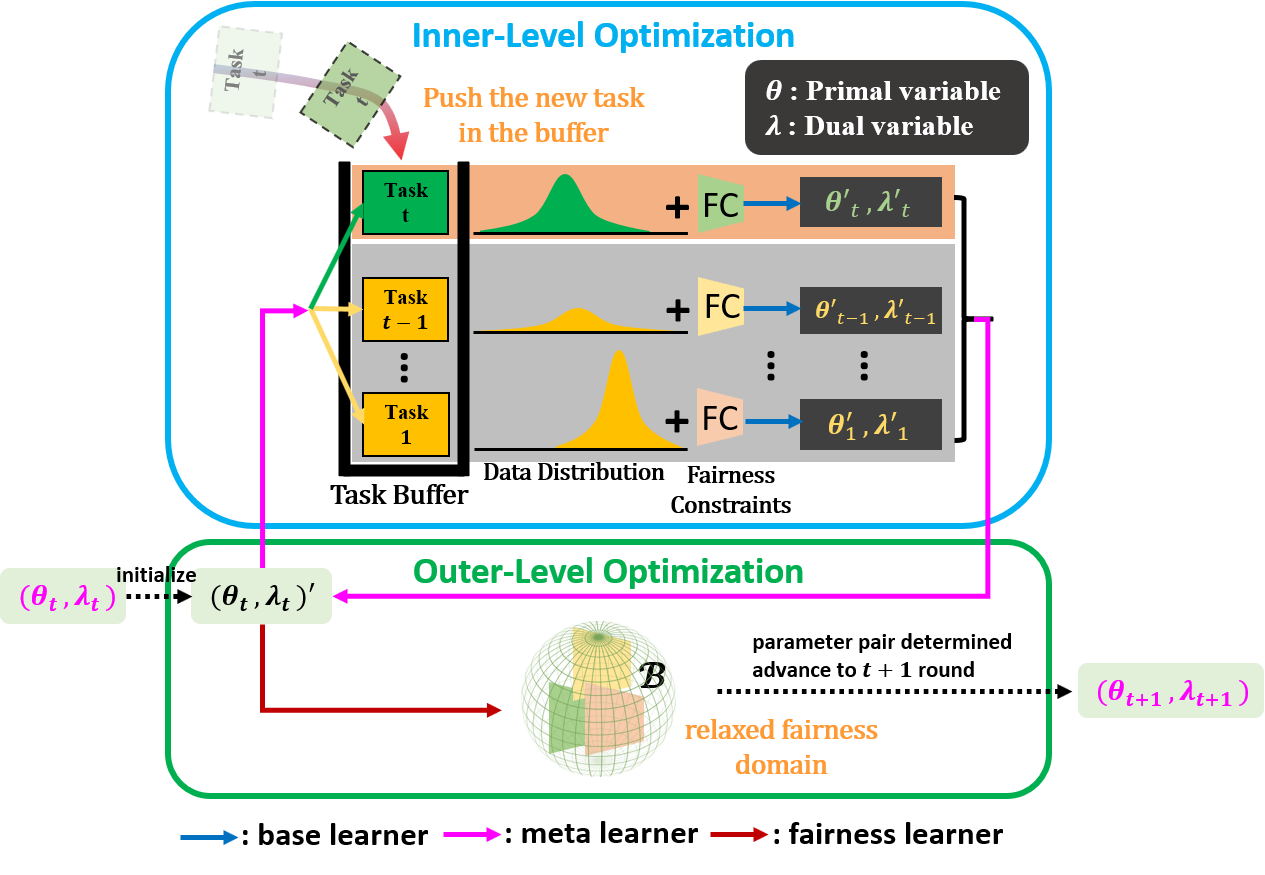}
    % \vspace{-8mm}
    \caption{An overview of update procedure stated in Step 3. At round $t$, new task is added in the the buffer. The parameter pair $\boldsymbol{(\theta}_t,\boldsymbol{\lambda}_t)$ are iteratively updated with fairness constraints through a bi-level optimization in which the inner and the outer interplay each other.}
    \label{fig:overview}
    % \vspace{-3mm}
\end{figure}

\subsection{An Efficient Algorithm}
The proposed Algorithm \ref{alg:PDRFTML} is composed of two levels where each responds for the inner and outer problems stated in Eq.(\ref{eq:inner-problem}) and Eq.(\ref{eq:outer-problem}), respectively. The output of the outer problem are used to define the inner objective function and vice versa. A parameter pair is therefore iteratively updated between the two levels.  

To solve the inner level problem stated in Eq.(\ref{eq:inner-problem}), we first consider the following Lagrangian function and omit $\mathcal{D}$ for the sake of brevity:
% \vspace{-8mm}
\begin{align}
\label{eq:inner-Lagrangian}
    \mathcal{L}_k(\boldsymbol{\theta}_k,\boldsymbol{\lambda}_k;\boldsymbol{\theta,\lambda}) = f_k(\boldsymbol{\theta}_k;\boldsymbol{\theta}) + \sum_{i=1}^m \lambda_{k,i} g_i(\boldsymbol{\theta}_k;\boldsymbol{\theta})
\end{align}
% \vspace{-5mm}

\begin{algorithm}[!t]
\caption{The \sysname{} Algorithm}
\label{alg:PDRFTML}
\begin{flushleft}
    \textbf{Require}: Learning rates $\eta_1, \eta_2$, some constant $\delta$.
\end{flushleft}
\begin{algorithmic}[1]
\State Randomly initialize primal and dual meta-parameters, $(\boldsymbol{\theta}_1, \boldsymbol{\lambda}_1)\in (\Theta\times\mathbb{R}_+^m)$
\State Initialize the task buffer $\mathcal{U}\leftarrow[]$
\For{$t=1,2,...,T$}
    \State Sample $\mathcal{D}_t^V\subseteq\mathcal{D}_t$ from task $t$
    \State \multiline{% 
        Evaluate performance of task $t$ using $\mathcal{D}_t^V$ and $\boldsymbol{\theta}_{t}$ }
    \State Append $\mathcal{D}_t$ to task buffer $\mathcal{U}\leftarrow\mathcal{U}+[\mathcal{D}_t]$
    \If{$t\neq 1$}
        \State Initialize $\boldsymbol{\theta}^{N_{iter}=0}\leftarrow\boldsymbol{\theta}_{t}$ and $\boldsymbol{\lambda}^{N_{iter}=0}\leftarrow\boldsymbol{\lambda}_{t}$
        \While{$N_{iter} = 1,2,...$}
            \For{each task $\mathcal{D}_k$ in $\mathcal{U}$ where $k\in[t]$}
                \State $\boldsymbol{\theta}_k\leftarrow\boldsymbol{\theta}^{N_{iter}-1}$; $\boldsymbol{\lambda}_k\leftarrow\boldsymbol{\lambda}^{N_{iter}-1}$
                \State Sample datapoints $\mathcal{D}_k^S\subset\mathcal{D}_k$
                \State \multiline{% 
                    Compute adapted task-level primal and dual pair $(\boldsymbol{\theta}'_k,\boldsymbol{\lambda}'_k)$ using Eq.($\ref{eq:task-level primal-dual}$) and (9).}
                \State Sample datapoints $\mathcal{D}_k^Q\subset\mathcal{D}_k$
                \State \multiline{% 
                    Compute $f_k(\boldsymbol{\theta}'_k, \mathcal{D}_k^Q)$ and $g_i(\boldsymbol{\theta}'_k, \mathcal{D}_k^Q), \forall i\in[m]$ using $\mathcal{D}_k^Q$.}
            \EndFor
            \State \multiline{% 
                Update meta-level primal-dual parameter pair ($\boldsymbol{\theta}^{N_{iter}}, \boldsymbol{\lambda}^{N_{iter}}$) using Eq.(\ref{eq:meta-level primal-dual}) and (12).}
        \EndWhile
        \State \multiline{% 
            Set meta-parameters $(\boldsymbol{\theta}_t, \boldsymbol{\lambda}_t)\leftarrow(\boldsymbol{\theta}^{N_{iter}},\boldsymbol{\lambda}^{N_{iter}})$}
    \EndIf
    \State $(\boldsymbol{\theta}_{t+1}, \boldsymbol{\lambda}_{t+1})\leftarrow(\boldsymbol{\theta}_t, \boldsymbol{\lambda}_t)$
    % \State Find a good pair meta-parameters for $\mathcal{D}_{t+1}$ that $(\boldsymbol{\theta}_{t+1}, \boldsymbol{\lambda}_{t+1})\leftarrow\textsc{Algorithm\ref{alg:meta-update}}(\mathcal{D}_t, \mathcal{U},\boldsymbol{\theta}_t, \boldsymbol{\lambda}_t)$
    
    % \State Sample batch of dataset $\mathcal{D}_t^V\subseteq\mathcal{D}_t$.
    % \State \multiline{% 
        % Record $loss_{\mathcal{D}_t} = f_t(\mathcal{A}lg(\boldsymbol{\theta}_{t+1}, \mathcal{D}_t^S), \mathcal{D}_t^V)$ and $fairness_{\mathcal{D}_t} = g_i(\mathcal{A}lg(\boldsymbol{\theta}_{t+1}, \mathcal{D}_t^S), \mathcal{D}_t^V), i=1,2,...,m$.}
    % \State $(\boldsymbol{\theta}_t, \boldsymbol{\lambda}_t)\leftarrow(\boldsymbol{\theta}_{t+1}, \boldsymbol{\lambda}_{t+1})$
    
\EndFor
\end{algorithmic}
\end{algorithm}

where $\boldsymbol{\theta}_k\in\mathcal{B}$ is the task-level primal variable initialized with the meta-level primal variable $\boldsymbol{\theta}$, and $\boldsymbol{\lambda}_k\in\mathbb{R}^m_+$ is the corresponding dual variable initialized with $\boldsymbol{\lambda}$, which is used to penalize the violation of constraints. Here, for the purpose of optimization with simplicity, constraints of Eq.(\ref{eq:inner-problem}) are approximated with the cumulative one shown in Eq.(\ref{eq:inner-Lagrangian}). To optimize, we update the task-level variables through a base learner $\mathcal{A}lg_k(\cdot):\boldsymbol{\theta}_k\in\mathcal{B}\rightarrow\boldsymbol{\theta}'_k\in\mathbb{R}^d$. One example for the learner is updating with one gradient step using the pre-determined stepsize $\eta_1>0$ \cite{Finn-ICML-2019}. Notice that for multiple gradient steps, $\boldsymbol{\theta}'_k$ and $\boldsymbol{\lambda}'_k$ interplay each other for updating.
% \vspace{-5mm}
\begin{align}
\label{eq:task-level primal-dual}
    &\boldsymbol{\theta}'_k =  \mathcal{A}lg_k(\boldsymbol{\theta}_k;\boldsymbol{\theta}) \triangleq \boldsymbol{\theta}_k - \eta_1\nabla_{\boldsymbol{\theta}}\mathcal{L}_k(\boldsymbol{\theta}_k,\boldsymbol{\lambda}_k;\boldsymbol{\theta,\lambda})\\
    &\boldsymbol{\lambda}'_k =\Big[ \boldsymbol{\lambda}_k + \eta_1\nabla_{\boldsymbol{\lambda}}\mathcal{L}_k(\boldsymbol{\theta}'_k,\boldsymbol{\lambda}_k;\boldsymbol{\theta,\lambda}) \Big]_+
\end{align}
Next, to solve the outer level problem, 
% a straightforward approach \cite{Abernethy-2012-TIT} is to introduce a projection free and self concordant barrier function for the convex domain $\Theta$ and add it to the objective function in Eq.(\ref{eq:outer-problem}). The main limitation is that it requires computing the inversion of the Hessian matrix of the objective function and hence suffers high complexity. 
% Furthermore, an alternative approach for online convex optimization with long term constraints is to introduce a penalty term in the loss function that penalizes the violation of constraints \cite{OGDLC-2012-JMLR}. The analysis shows that in order to obtain $O(T^{1/2})$ regret bound, linear bound on the long term violation of the constraints is unavoidable.
the intuition behind our approach stems from the observation that the constrained optimization problem is equivalent to a convex-concave optimization problem with respect to the outer-level primal variable $\boldsymbol{\theta}$ and dual variable $\boldsymbol{\lambda}$. We hence consider the following augmented Lagrangian function:
\begin{align}
\label{eq:outer-Lagrangian}
    \mathcal{L}_t(\boldsymbol{\theta,\lambda})=
    \frac{1}{t}\sum_{k=1}^{t} \Bigg\{ f_k(\boldsymbol{\theta};\boldsymbol{\theta}'_k) 
    + \sum_{i=1}^m \Big(\lambda_i g_i(\boldsymbol{\theta};\boldsymbol{\theta}'_k) - \frac{\delta\eta_2}{2}\lambda_i^2 \Big)\Bigg\} 
\end{align}

where $\delta > 0$ and $\eta_2>0$ are some constant and stepsize whose values will be decided by the analysis. 
% $\mathcal{F}(\cdot)$ is an appropriate meta-regularizer (\textit{e.g.} Euclidean norm $||\boldsymbol{\theta}||$) ensuring the stability of the proposed algorithm and $\zeta>0$ is its trade-off parameter. 
Besides, the augmented term on the dual variable is devised to prevent $\boldsymbol{\lambda}$ from being too large \cite{OGDLC-2012-JMLR}.
To optimize the meta-parameter pair in the outer problem, the update rule follows
\begin{align}
\label{eq:meta-level primal-dual}
    \boldsymbol{\theta}_{t+1} &= \prod_\mathcal{B}(\boldsymbol{\theta}_{t} - \eta_2\nabla_{\boldsymbol{\theta}}\mathcal{L}_t(\boldsymbol{\theta}_{t},\boldsymbol{\lambda}_{t})) \nonumber\\
    &= \arg\min_{\boldsymbol{y}\in\mathcal{B}} ||\boldsymbol{y}-(\boldsymbol{\theta}_{t} - \eta_2\nabla_{\boldsymbol{\theta}}\mathcal{L}_t(\boldsymbol{\theta}_{t},\boldsymbol{\lambda}_{t}))||^2   \\
    \boldsymbol{\lambda}_{t+1} &= \Big[ \boldsymbol{\lambda}_{t} + \eta_2\nabla_{\boldsymbol{\lambda}}\mathcal{L}_t(\boldsymbol{\theta}_t,\boldsymbol{\lambda}_{t}) \Big]_+
\end{align}

where $\prod_\mathcal{B}$ is the projection operation to the relaxed domain $\mathcal{B}$ that is introduced in Sec.\ref{sec:settings and prob formulation}. This approximates the true desired projection with a simpler closed-form.

We detail the iterative update procedure in Algorithm \ref{alg:PDRFTML}. At round $t\in[T]$, we first evaluate the performance on the new arrived task $\mathcal{D}_t$ using $\boldsymbol{\theta}_t$ (line 5) and $\mathcal{D}_t$ is added to the task buffer $\mathcal{U}$ (line 6). As for the bi-level update, each task-level parameters $\boldsymbol{\theta}_k,\boldsymbol{\lambda}_k$ from the buffer are initialized with the meta-level ones (line 8). In line 13, task-level parameters are updated using support data. Query loss and fairness for each task are further computed in line 15 and they are used to optimize meta parameters in line 17. An overview of the update procedure is described in Figure \ref{fig:overview}.

\begin{sloppypar}
Different from techniques devised to solve online learning problems with long-term constraints, at each round \sysname{} finds a good primal-dual parameter pair by learning from prior experience through dealing with a bi-level optimization problem. In order to ensure bias-free predictions, objective functions in both inner and outer levels subject to fairness constraints. Besides, since we generalize the traditional primal-dual update scheme onto a bi-level optimization problem, conventional theoretic guarantees cannot be applied. We hence demonstrate analysis of \sysname{} in the following section. 
\end{sloppypar}

%% file: analysis.tex
To analysis, in this paper, we first make following assumptions as in \cite{Finn-ICML-2019} and \cite{OGDLC-2012-JMLR}. These assumptions are commonly used in meta learning and online learning settings. Examples where these assumptions hold include logistic regression and $L2$ regression over a bounded domain. As for constraints, a family of fairness notions, such as linear relaxation based DDP (Difference of Demographic Parity) including Eq.(\ref{eq:DBC constraint}), are applicable as discussed in \cite{Lohaus-2020-ICML}.
\begin{assumption}[Convex domain]
\label{assmp1}
    The convex set $\Theta$ is non-empty, closed, bounded, and can be described by $m$ convex functions as $\Theta = \{\boldsymbol{\theta}:g_i(\boldsymbol{\theta})\leq 0, \forall i\in[m]\}$
\end{assumption}

\begin{assumption}
\label{assmp2}
    Both the loss functions $f_t(\cdot), \forall t$ and constraint functions $g_i(\cdot), \forall i\in[m]$ satisfy the following assumptions 
    \begin{enumerate}[leftmargin=*]
        \item 
        \begin{sloppypar}
        (Lipschitz continuous) $f_t(\boldsymbol{\theta}), \forall t$ and $g_i(\boldsymbol{\theta}),\forall i$ are Lipschitz continuous in $\mathcal{B}$, that is, $\forall \boldsymbol{\theta,\phi}\in\mathcal{B}$,  $||f_t(\boldsymbol{\theta})-f_t(\boldsymbol{\phi})||\leq L_f||\boldsymbol{\theta-\phi}||, ||g_i(\boldsymbol{\theta})-g_i(\boldsymbol{\phi})||\leq L_g||\boldsymbol{\theta-\phi}||$, and $G = \max\{L_f,L_g\}$, and 
        \end{sloppypar}
        \begin{align*}
            &F = \max_{t\in[T]}\max_{\boldsymbol{\theta,\phi}\in\mathcal{B}} f_t(\boldsymbol{\theta})-f_t(\boldsymbol{\phi})\leq 2L_f R \\
            &D = \max_{i\in[m]}\max_{\boldsymbol{\theta}\in\mathcal{B}}g_i(\boldsymbol{\theta})\leq L_g R
        \end{align*}
        % $F = \max_{t\in[T]}\max_{\boldsymbol{\theta,\phi}\in\mathcal{B}} f_t(\boldsymbol{\theta})-f_t(\boldsymbol{\phi})\leq 2L_f R$, $D = \max_{i\in[m]}\max_{\boldsymbol{\theta}\in\mathcal{B}}g_i(\boldsymbol{\theta})\leq L_g R$.
        \item (Lipschitz gradient) $f_t(\boldsymbol{\theta}), \forall t$ are $\beta_f$-smooth and $g_i(\boldsymbol{\theta}),\forall i$ are $\beta_g$-smooth, that is, $\forall \boldsymbol{\theta,\phi}\in\mathcal{B}$, $||\nabla f_t(\boldsymbol{\theta})-\nabla f_t(\boldsymbol{\phi})||\leq \beta_f||\boldsymbol{\theta-\phi}||,\\ ||\nabla g_i(\boldsymbol{\theta})-\nabla g_i(\boldsymbol{\phi})||\leq \beta_g||\boldsymbol{\theta-\phi}||$, and $H=\max\{\beta_f,\beta_g\}$.
        \item (Lipschitz Hessian) Twice-differentiable functions $f_t(\boldsymbol{\theta}), \forall t$ and $g_i(\boldsymbol{\theta}),\forall i$ have $\rho_f$ and $\rho_g$- Lipschitz Hessian, respectively. That is, $\forall \boldsymbol{\theta,\phi}\in\mathcal{B}$, $||\nabla^2 f_t(\boldsymbol{\theta})-\nabla^2 f_t(\boldsymbol{\phi})||\leq \rho_f||\boldsymbol{\theta-\phi}||,\\ ||\nabla^2 g_i(\boldsymbol{\theta})-\nabla^2 g_i(\boldsymbol{\phi})||\leq \rho_g||\boldsymbol{\theta-\phi}||$.
    \end{enumerate}
\end{assumption}

\begin{assumption}[Strongly convexity]
\label{assmp3}
    \begin{sloppypar}
    Suppose $f_t(\boldsymbol{\theta}), \forall t$ and $g_i(\boldsymbol{\theta}),\forall i$ have strong convexity, that is, $\forall \boldsymbol{\theta,\phi}\in\mathcal{B}$, $||\nabla f_t(\boldsymbol{\theta})-\nabla f_t(\boldsymbol{\phi})||\geq \mu_f||\boldsymbol{\theta-\phi}||, ||\nabla g_i(\boldsymbol{\theta})-\nabla g_i(\boldsymbol{\phi})||\geq \mu_g||\boldsymbol{\theta-\phi}||$.
    \end{sloppypar}
\end{assumption}

We then analyze the proposed \sysname{} algorithm and use one-step gradient update as an example. Under above assumptions, we first target Eq.(\ref{eq:outer-Lagrangian}) and state:

\begin{theorem}
\label{theorem1}
    Suppose $f$ and $g:\Theta\times\mathbb{R}_+^m \rightarrow \mathbb{R}$ satisfy Assumptions \ref{assmp1}, \ref{assmp2} and \ref{assmp3}. The inner level update and the augmented Lagrangian function $\mathcal{L}_t(\boldsymbol{\theta, \lambda})$ are defined in Eq.(\ref{eq:task-level primal-dual})(7) and Eq.(\ref{eq:outer-Lagrangian}). Then, the function $\mathcal{L}_t(\boldsymbol{\theta, \lambda})$ is convex-concave with respect to the arguments $\boldsymbol{\theta}$ and $\boldsymbol{\lambda}$, respectively. Furthermore, as for $\mathcal{L}_t(\boldsymbol{\cdot,\lambda})$, if stepsize $\eta_1$ is selected as $\eta_1\leq\min\{\frac{\mu_f+\Bar{\lambda}m\mu_g}{8(L_f+\Bar{\lambda}mL_g)(\rho_f+\Bar{\lambda}m\rho_g)}, \frac{1}{2(\beta_f+\Bar{\lambda}m\beta_g)}\}$, then $\mathcal{L}_t(\boldsymbol{\cdot,\lambda})$ enjoys $\frac{9}{8}(\beta_f+\Bar{\lambda}m\beta_g)$-smooth and $\frac{1}{8}(\mu_f+\Bar{\lambda} m\mu_g)$-strongly convex, where $\Bar{\lambda}\geq 0$ is the mean value of $\boldsymbol{\lambda}$.
\end{theorem}
% Proof of Theorem \ref{theorem1} is given in Appendix \ref{App:proofT1}. 
We next present the key Theorem \ref{theorem2}. We state that \sysname{} enjoys sub-linear guarantee for both regret and long-term fairness constraints in the long run for Algorithm \ref{alg:PDRFTML}. 
% In order to better understand Theorem \ref{theorem2}, we first introduce Lemma \ref{lemma} and its analysis is modified and analogous to that developed in \cite{OGDLC-2012-JMLR}.
% \begin{lemma}
% \label{lemma}
% Let $\mathcal{L}_t(\cdot,\cdot)$ be the function defined in Eq.(\ref{eq:outer-Lagrangian}), which is convex in its first argument and concave in its second argument. Let $\boldsymbol{\theta}_t$ and $\boldsymbol{\lambda}_t, t\in[T]$ be the sequence of solution obtained by Algorithm \ref{alg:PDRFTML}. Then for any $(\boldsymbol{\theta,\lambda})\in\mathcal{B}\times\mathbb{R}^m_+$, we have
% \begin{align}
% \label{eq:lemma_bound}
%     &\sum_{t=1}^T \mathcal{L}_t(\boldsymbol{\theta}_t, \boldsymbol{\lambda})-\mathcal{L}_t(\boldsymbol{\theta}, \boldsymbol{\lambda}_t) \leq \frac{R^2+||\boldsymbol{\lambda}||^2}{2\eta_2} - \frac{\mu_f}{2}R^2\\
%     &+\sum_{t=1}^T\Big\{\frac{\eta_2}{2}\Big(4G^2\eta_1^2H^2(m+1)^2
%     % +2\zeta^2
%     +4m(D^2+\eta_1^2G^4)\Big)\nonumber \\
%     &+2\eta_2m(G^4\eta_1^2m^2+\delta^2\eta_2^2)||\boldsymbol{\lambda}_t||^2
%     +2\eta_2G^2\eta_1^2H^2(m+1)^2||\boldsymbol{\lambda}_t||^4 \Big\} \nonumber
% \end{align}
% \end{lemma}

% By applying Lemma \ref{lemma}, the following Theorem \ref{theorem2} bounds the regret and the violation of the constraints in the long run for Algorithm \ref{alg:PDRFTML}.

\begin{theorem}
\label{theorem2}
    Set $\eta_2 = \mu_f/(t+1)$ and choose $\delta$ such that $\delta\geq\max\{4m(G^4\eta_1^2m^2+ \delta^2\eta_2^2),4G^2\eta_1^2H^2(m+1)^2\}$. If we follow the update rule in Eq.(\ref{eq:meta-level primal-dual})(10) and $\boldsymbol{\theta}^*$ being the optimal solution for $\min_{\boldsymbol{\theta}\in\Theta} \sum_{t=1}^T \\ f_t(\mathcal{A}lg_t(\boldsymbol{\theta}))$, we have upper bounds for both the regret on the loss and the cumulative constraint violation
    \begin{align*}
        &\sum_{t=1}^T \Big\{ f_t(\mathcal{A}lg_t(\boldsymbol{\theta}_t)) - f_t(\mathcal{A}lg_t(\boldsymbol{\theta}^*)) \Big\} \leq O(\log T) \\
        &\sum_{t=1}^T g_i(\mathcal{A}lg_t(\boldsymbol{\theta}_t)) \leq O(\sqrt{T\log T}), \forall i\in[m]
    \end{align*}
\end{theorem}

% A family of strongly convex functions are discussed in \cite{Jenatton-2016-ICML}. Algorithm \ref{alg:PDRFTML} and the resulting bound are useful for two reasons: (1) The co-update of the primal-dual pair in the inner problem implies penalties for task-level unfairness violations. (2) Algorithm \ref{alg:PDRFTML} can also upper bound the constraint violation for each round.
\noindent \textbf{Regret Discussion:} Under aforementioned assumptions and provable convexity of Eq.(\ref{eq:outer-Lagrangian}) in $\boldsymbol{\theta}$ (see Theorem \ref{theorem1}), Algorithm \ref{alg:PDRFTML} achieves sublinear bounds for both loss regret and violation of fairness constraints (see Theorem \ref{theorem2}) where $\lim_{T\rightarrow\infty} O(\cdot)/T=0$. Although such bounds are comparable with cutting-edge techniques of online learning with long-term constraints \cite{GenOLC-2018-NeurIPS,AdpOLC-2016-ICML} in the case of strongly convexity, in terms of online meta-learning paradigms, for the first time we bound loss regret and cumulative fairness constraints simultaneously.
For space purposes, proofs for all theorems are contained in the Appendix \ref{App:proofT1} and \ref{App:proofT2}.

%% file: experiments.tex
To corroborate our algorithm, we conduct extensive experiments
% \footnote{The link to the code and datasets are available at \url{https://github.com/charliezhaoyinpeng/ffml}.},
comparing \sysname{} with some popular baseline methods. We aim to answer the following questions: 
\begin{enumerate}[leftmargin=*]
    \item \textbf{Question 1: } Can \sysname{} achieve better performance on both fairness and classification accuracy compared with baseline methods?
    \item 
    \begin{sloppypar}
    \textbf{Question 2: } Can \sysname{} be successfully applied to non-stationary learning problems and achieve a bounded fairness as $t$ increases?
    \end{sloppypar}
    \item \textbf{Question 3: } How efficient is \sysname{} in task evaluation over time and how is the contribution for each component? 
\end{enumerate}
% (1) \textbf{Question 1: } Can \sysname{} achieve better performance on both fairness and classification accuracy compared with baseline methods? (2) \textbf{Question 2: } Can \sysname{} be successfully applied to non-stationary learning problems and achieve a bounded fairness as $t$ increases? (3) \textbf{Question 3: } How efficient is \sysname{} in task evaluation over time and how is the contribution for each component? 
% The link to the code and datasets are available at \url{https://bit.ly/2LwGSvk}.
% \begin{itemize}[leftmargin=*]
%     \item \textbf{Question 1: } Can \sysname{} achieve better performance on both fairness and classification accuracy compared with baseline methods?
%     \item \textbf{Question 2: } Can \sysname{} be successfully applied to non-stationary learning problems and achieve a bounded fairness as $t$ increases?
%     \item \textbf{Question 3: } How efficient is \sysname{} in terms of task evaluation over time with which component?
% \end{itemize}
\subsection{Datasets}
We use the following three publicly available datasets. Each dataset contains a sequence of tasks where the ordering of
tasks is selected at random. (1) \textbf{Bank Marketing} \cite{Moro-bank-marketing-dataset-2014} contains a total 41,188 subjects, each with 16 attributes and a binary label, which indicates whether the client has subscribed or not to a term deposit. We consider the marital status as the binary protected attribute.
% , which is discretized to indicate whether the client is married or not. 
% Since the dataset contains information of different months and dates, we combine them as task labels and thus the dataset contains 50 tasks.
The dataset contains 50 tasks and each corresponds to a date when the data are collected from April to December in 2013.
(2) \textbf{Adult} \cite{AdultDataSet-UCI-1994} is broken down into a sequence of 41 income classification tasks, each of which relates to a specific native country. The dataset totally 48,842 instances with 14 features and a binary label, which indicates whether a subject’s incomes is above or below 50K dollars. We consider gender, \textit{i.e.} male and female, as the protected attribute. (3) \textbf{Communities and Crime} \cite{CommunitiesandCrimeDataSet-UCI-1994} 
% includes crime records across the U.S. We convert this dataset by using each state as a different task. 
is split into 43 crime classification tasks where each corresponds to a state in the U.S.
Following the same setting in \cite{Slack-FAT-2019}, we convert the violent crime rate into binary labels based on whether the community is in the top $50\%$ crime rate within a state. Additionally, we add a binary protected attribute that receives a protected label if African-Americans are the highest or second highest population in a community in terms of percentage racial makeup. 

\subsection{Evaluation Metrics}
% To evaluate the proposed techniques for fairness learning, we introduced three classic evaluation metrics to measure data biases. 
Three popular evaluation metrics are introduced that each allows quantifying the extent of bias taking into account the protected attribute.

\textbf{Demographic Parity} (DP) \cite{Dwork-2011-CoRR} and \textbf{Equalized Odds} (EO) \cite{Hardt-NIPS-2016} can be formalized as
\begin{align*}
    \text{DP} = \frac{P(\hat{Y}=1|S=0)}{P(\hat{Y}=1|S=1)}; \quad \text{EO} = \frac{P(\hat{Y}=1|S=0,Y=y)}{P(\hat{Y}=1|S=1,Y=y)}
\end{align*}
where $y\in\{0,1\}$. Equalized odds requires that $\hat{Y}$ have equal true positive rates and false positive rates between sub-groups. For both metrics, a value closer to 1 indicate fairness.

\textbf{Discrimination} \cite{Zemel-ICML-2013} measures the bias with respect to the protected attribute $S$ in the classification:
\begin{align*}
    \text{Disc} = \Bigg| {\frac{\sum_{i:s_i=1} \hat{y}_i}{\sum_{i:s_i=1} 1} - \frac{\sum_{i:s_i=0} \hat{y}_i}{\sum_{i:s_i=0} 1}} \Bigg|
\end{align*}
This is a form of statistical parity that is applied to the binary classification decisions. 
% It measures the difference in the proportion of positive classifications of individuals in the protected and unprotected groups. 
We re-scale values across all baseline methods into a range of $[0,1]$ and $Disc=0$ indicates there is no discrimination.

\begin{table*}[!t]
% \small
    \centering
    \caption{End task performance on real datasets over all baseline methods. Evaluation metrics with "$\uparrow$" indicates the bigger the better and "$\downarrow$" indicates the smaller the better. Best performance are labeled in bold.}
    % \vspace{-3mm}
    \setlength\tabcolsep{3.7pt}
    \begin{tabular}{c|c|c|c|c}
        \dtoprule
        \multirow{2}{*}{Dataset} & DP $\uparrow$ & EO $\uparrow$ & Disc $\downarrow$ & Acc($\%$) $\uparrow$ \\
        \cline{2-5}
         & \multicolumn{4}{c}{TWP / m-FTML\cite{Finn-ICML-2019} / OGDLC\cite{OGDLC-2012-JMLR} / AdpOLC\cite{AdpOLC-2016-ICML} / GenOLC\cite{GenOLC-2018-NeurIPS} / \sysname{}(Ours)} \\
        \hline
        Bank & 0.09/0.68/0.36/0.76/0.81/\textbf{0.97} & 0.11/0.65/0.35/0.72/0.78/\textbf{0.96} & 0.19/0.72/0.23/0.22/0.11/\textbf{0.07} & 52.14/53.69/52.36/54.89/\textbf{56.78}/52.55 \\
        \hline
        Adult & 0.05/0.54/0.41/0.60/0.78/\textbf{0.91}\ & 0.04/0.43/0.30/0.62/0.69/\textbf{0.87} & 0.32/0.69/0.21/0.19/0.12/\textbf{0.10} & 51.21/\textbf{67.91}/52.31/48.36/47.87/61.35 \\
        \hline
        Crime & 0.40/0.38/0.48/0.68/0.70/\textbf{0.74} & 0.38/0.29/0.39/0.43/0.64/\textbf{0.69} & 0.23/0.78/0.25/0.31/\textbf{0.15}/0.17 & 51.23/48.69/49.10/58.89/49.43/\textbf{59.57} \\
        \dbottomrule
    \end{tabular}
    \label{tab:endPerformance}
    % \vspace{-4mm}
\end{table*}

\subsection{Competing Methods}
% We consider the following state-of-the-art baseline methods:
% (1) \textbf{Train on everything (TOE)}: trains on all available data so far and trains a single predictive model. This model is directly tested without any specific adaptation.
(1) \textbf{Train with penalty (TWP)}: is an intuitive approach for online fair learning where loss functions at each round is penalized by the violation of fairness constraints. We then run the standard online gradient descent (OGD) algorithm to minimize the modified loss function.
(2) \textbf{m-FTML} \cite{Finn-ICML-2019}: the original FTML finds a sequence of meta parameters by simply applying MAML \cite{Finn-ICML-2017-(MAML)} at each round. To focus on fairness learning, this approach is applied to modified datasets by removing protected attributes. 
Notice that techniques (3)-(5) are proposed for online learning with long-term constraints and achieve state-of-the-art performance in theoretic guarantees. In order to fit bias-prevention and compare them to \sysname{}, we specify such constraints as DBC stated in Eq.(\ref{eq:DBC constraint}).
(3) \textbf{OGDLC} \cite{OGDLC-2012-JMLR}: updates parameters solely based on the on-going task.
(4) \textbf{AdpOLC} \cite{AdpOLC-2016-ICML}: improves OGDLC by modifying stepsizes to an adapted version.
(5) \textbf{GenOLC} \cite{GenOLC-2018-NeurIPS}: rectifies AdpOLC by square-clipping the constraints in place of $g_i(\cdot), \forall i$. Although OGDLC, AdpOLC and GenOLC are devised for online learning with long-term constraints, none of these state-of-the-arts considers inner-level adaptation. We note that, among the above baseline methods, m-FTML is a state-of-the-art one in the the area of online meta learning. The last three baselines, including OGDLC, AdpOLC, GenOLC are representative ones in the area of online optimization with long term constraints. 
% The two most relevant methods in the area of online fairness-aware learning are \cite{Vishakha-2020-AAAI} and \cite{Bechavod-2019-NeurIPS}, but the former is designed for fairness on resource allocation of multi-armed bandit problem and the latter is designed for online classification with partial feedback, both of which are not applicable to the standard online learning setting of group fairness as considered in this paper. 

\begin{figure}[!t]
% \captionsetup[subfigure]{aboveskip=-2pt,belowskip=-2pt}
\centering
    \begin{subfigure}[b]{0.235\textwidth}
        \includegraphics[width=\textwidth]{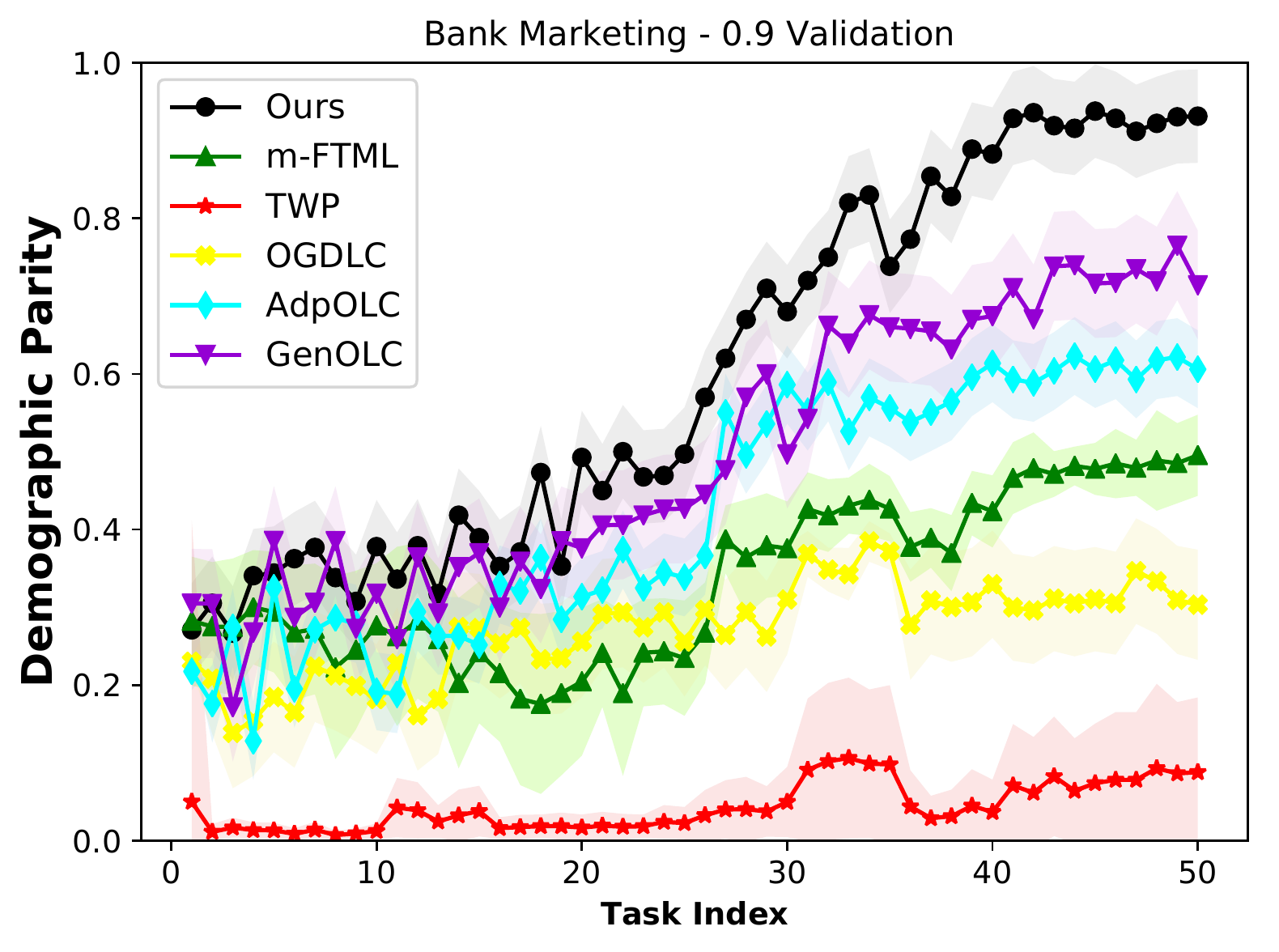}
        % \caption{}
    \end{subfigure}
    \begin{subfigure}[b]{0.235\textwidth}
        \includegraphics[width=\textwidth]{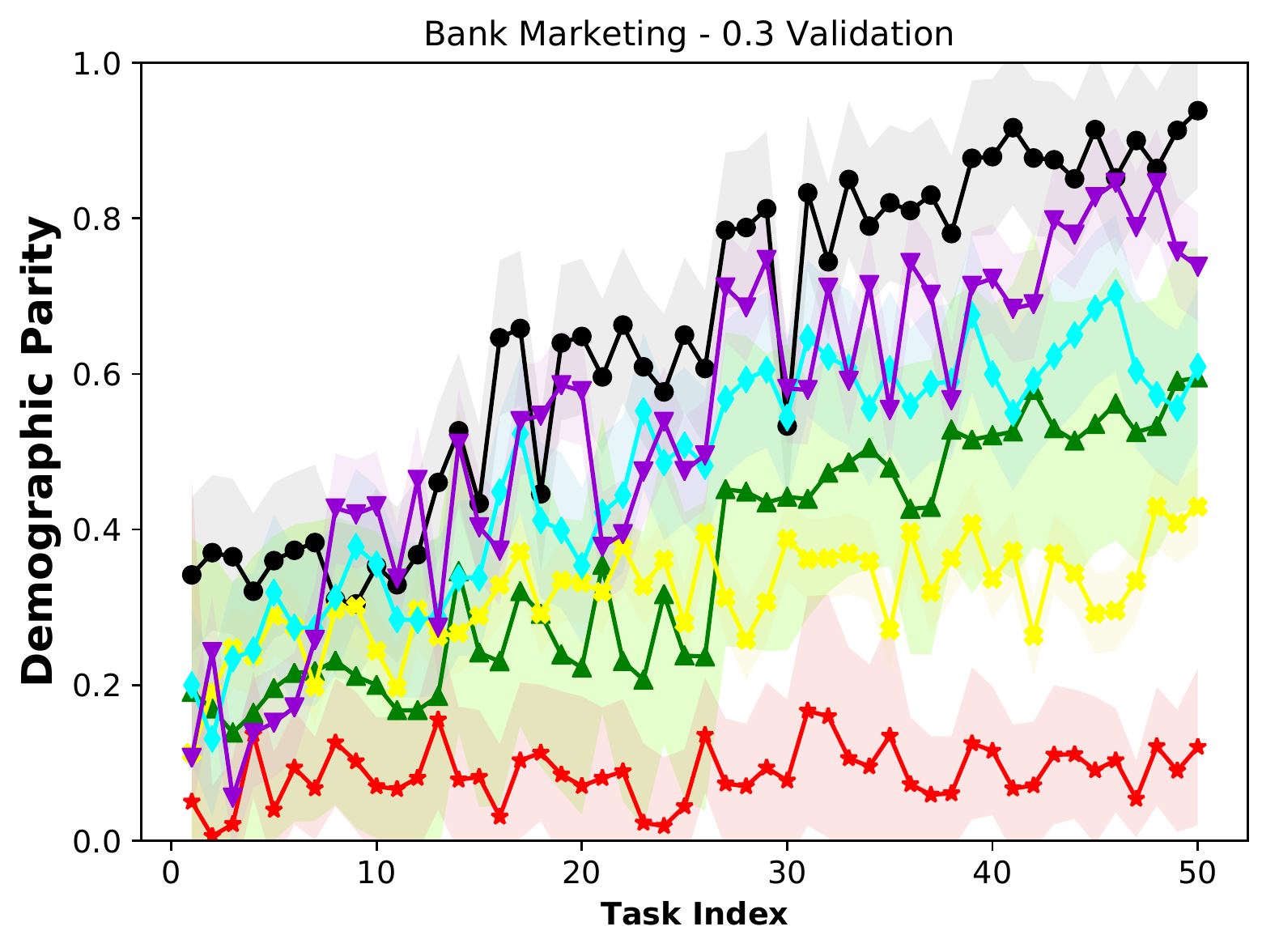}
        % \caption{}
    \end{subfigure}
    
    \begin{subfigure}[b]{0.235\textwidth}
        \includegraphics[width=\textwidth]{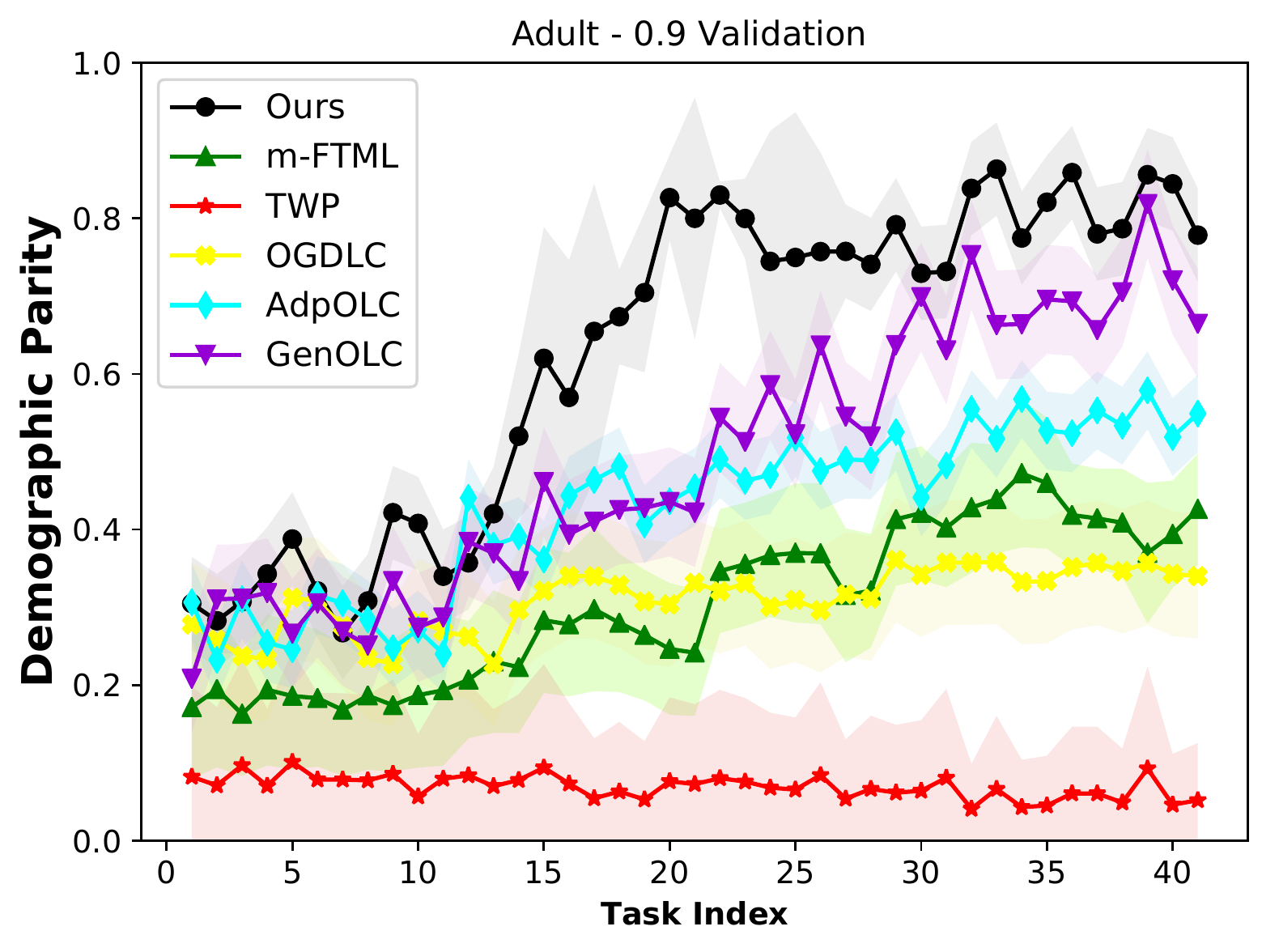}
        % \caption{}
    \end{subfigure}
    \begin{subfigure}[b]{0.235\textwidth}
        \includegraphics[width=\textwidth]{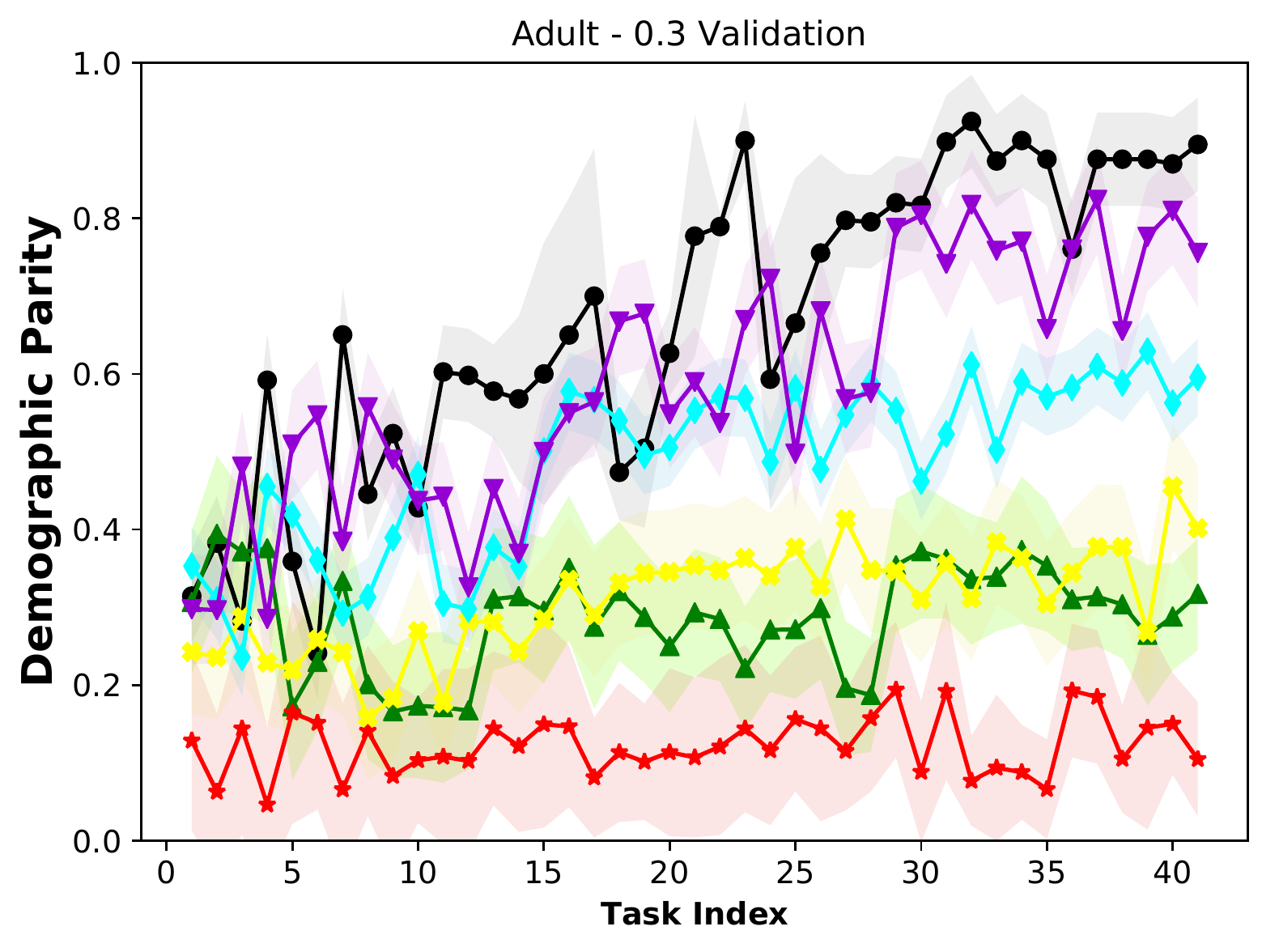}
        % \caption{}
    \end{subfigure}
    
    \begin{subfigure}[b]{0.235\textwidth}
        \includegraphics[width=\textwidth]{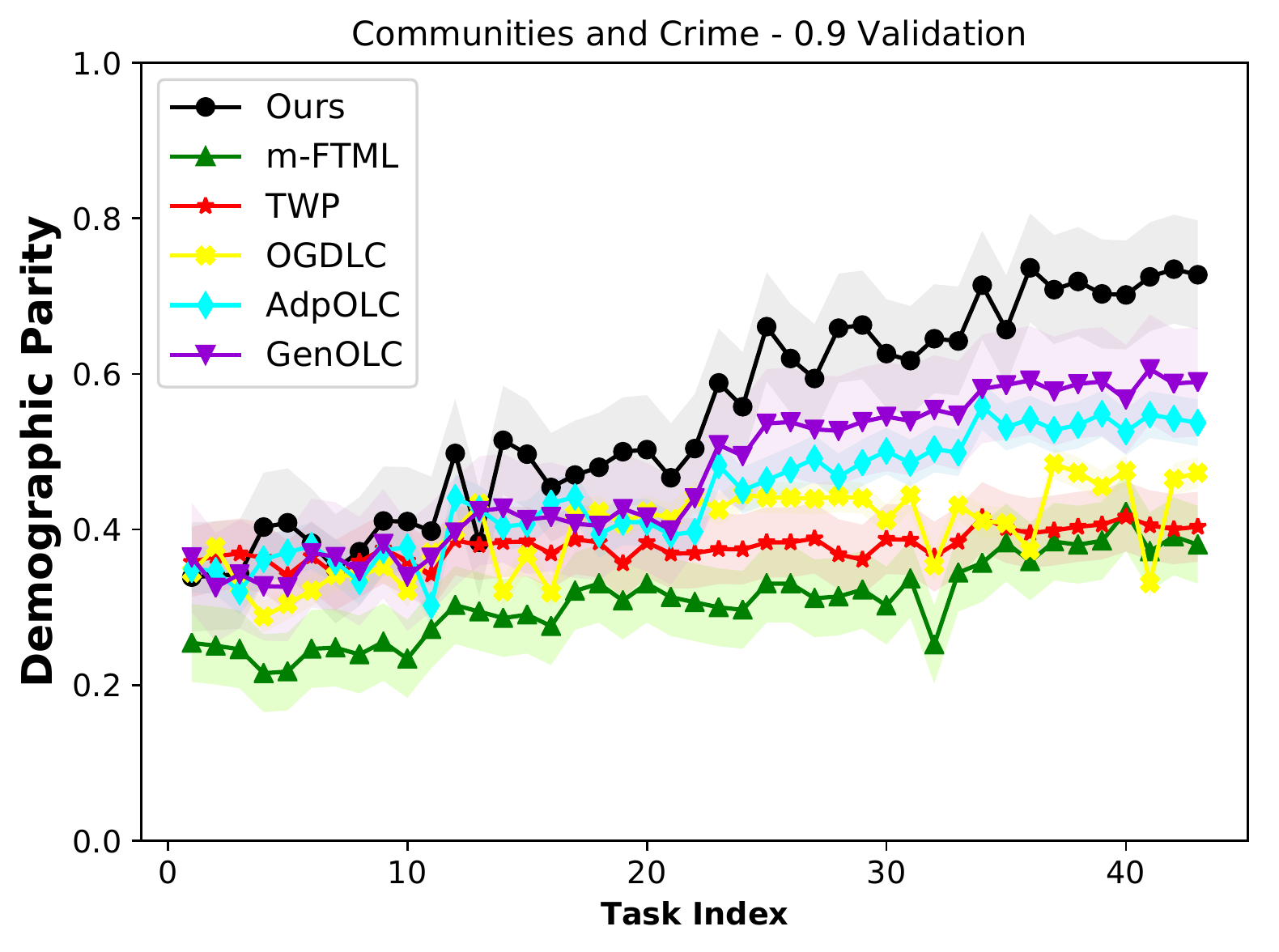}
        % \caption{}
    \end{subfigure}
    \begin{subfigure}[b]{0.235\textwidth}
        \includegraphics[width=\textwidth]{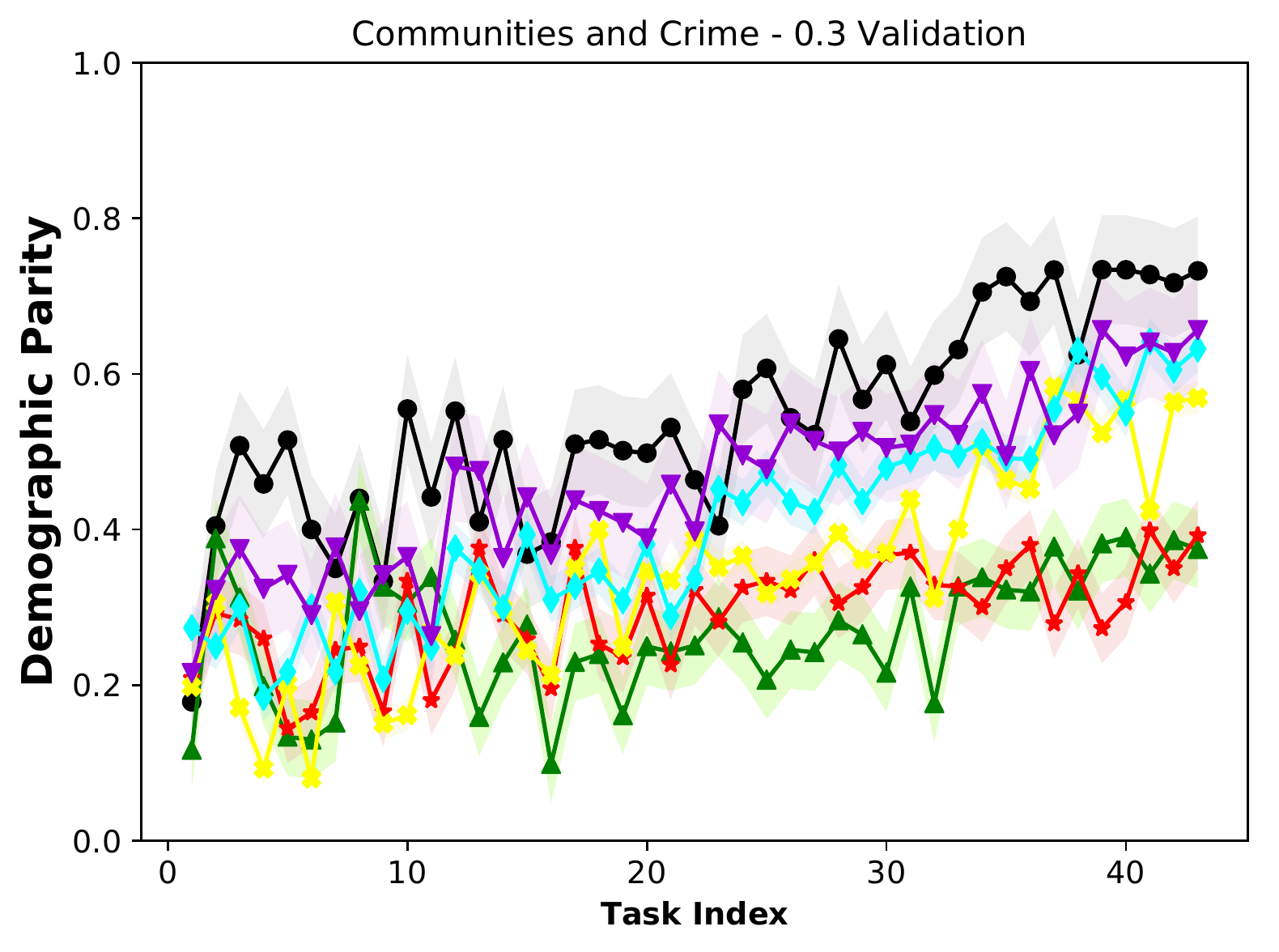}
        % \caption{}
    \end{subfigure}
    % \vspace{-3mm}
    \caption{Evaluation using fair metric DP at each round.}
    \label{fig:DP-over-tasks}
% \vspace{-5mm}
\end{figure}

\begin{figure}[!t]
% \captionsetup[subfigure]{aboveskip=-2pt,belowskip=-2pt}
\centering
    \begin{subfigure}[b]{0.235\textwidth}
        \includegraphics[width=\textwidth]{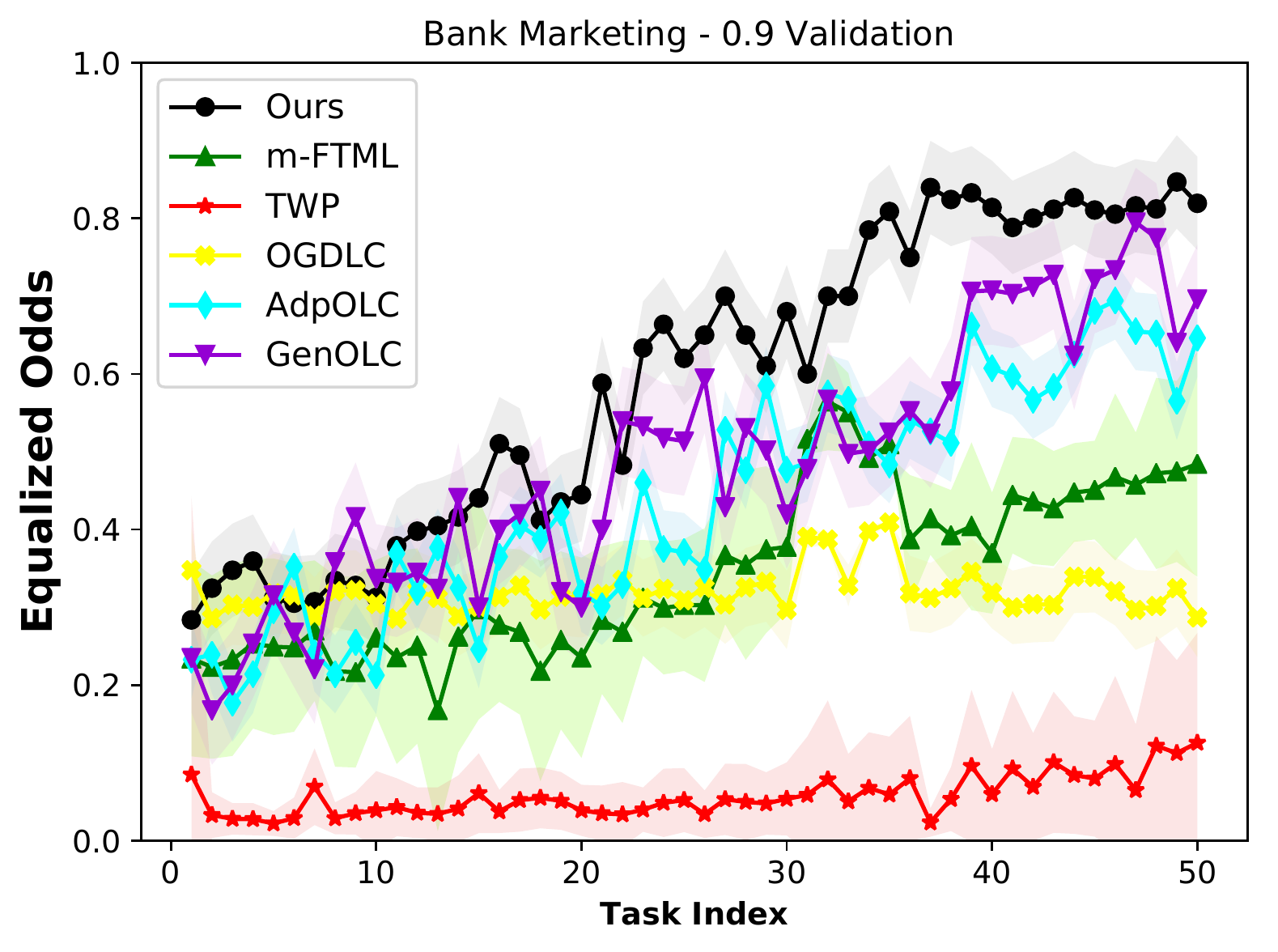}
        % \caption{}
    \end{subfigure}
    \begin{subfigure}[b]{0.235\textwidth}
        \includegraphics[width=\textwidth]{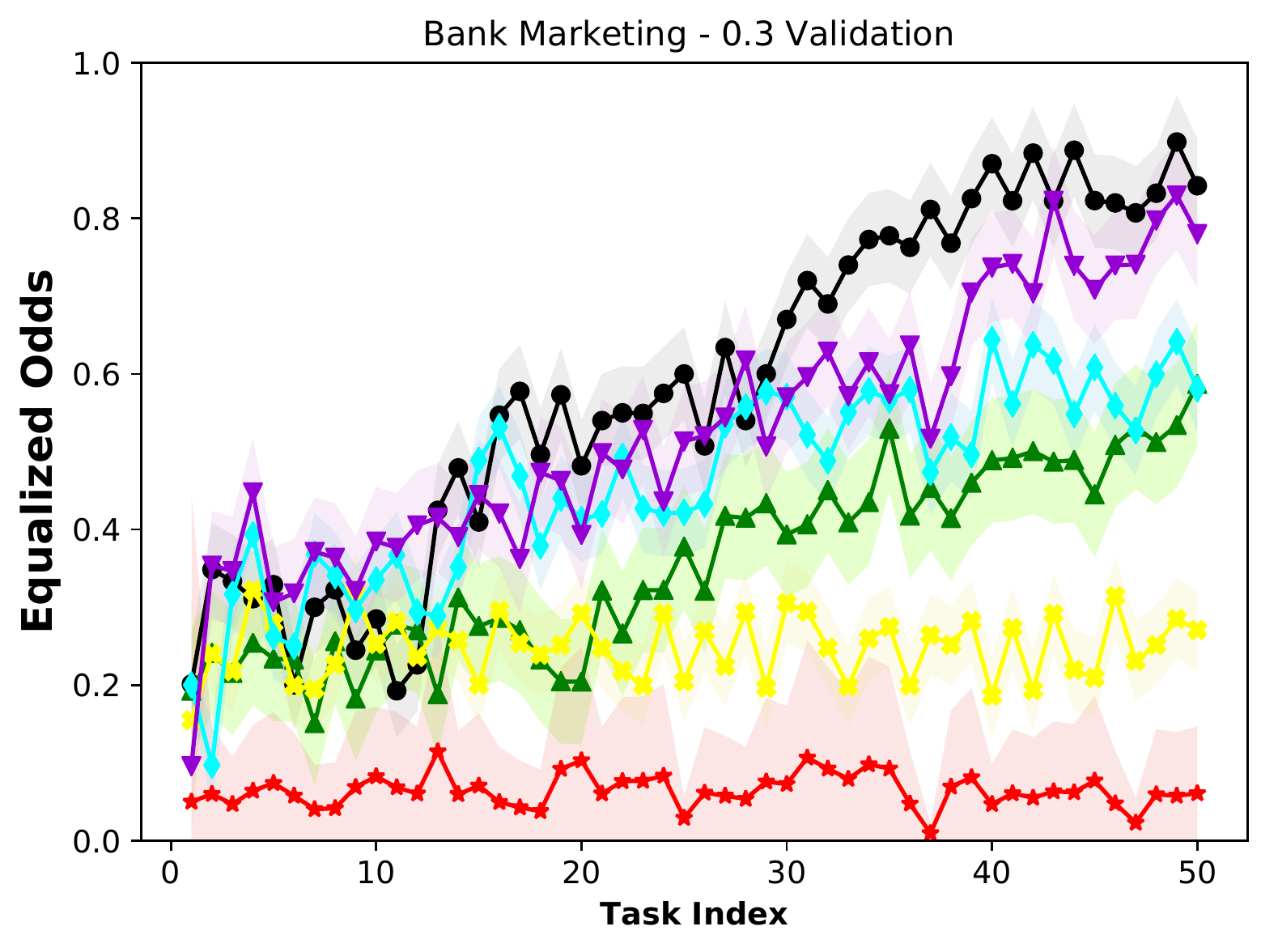}
        % \caption{}
    \end{subfigure}
    
    \begin{subfigure}[b]{0.235\textwidth}
        \includegraphics[width=\textwidth]{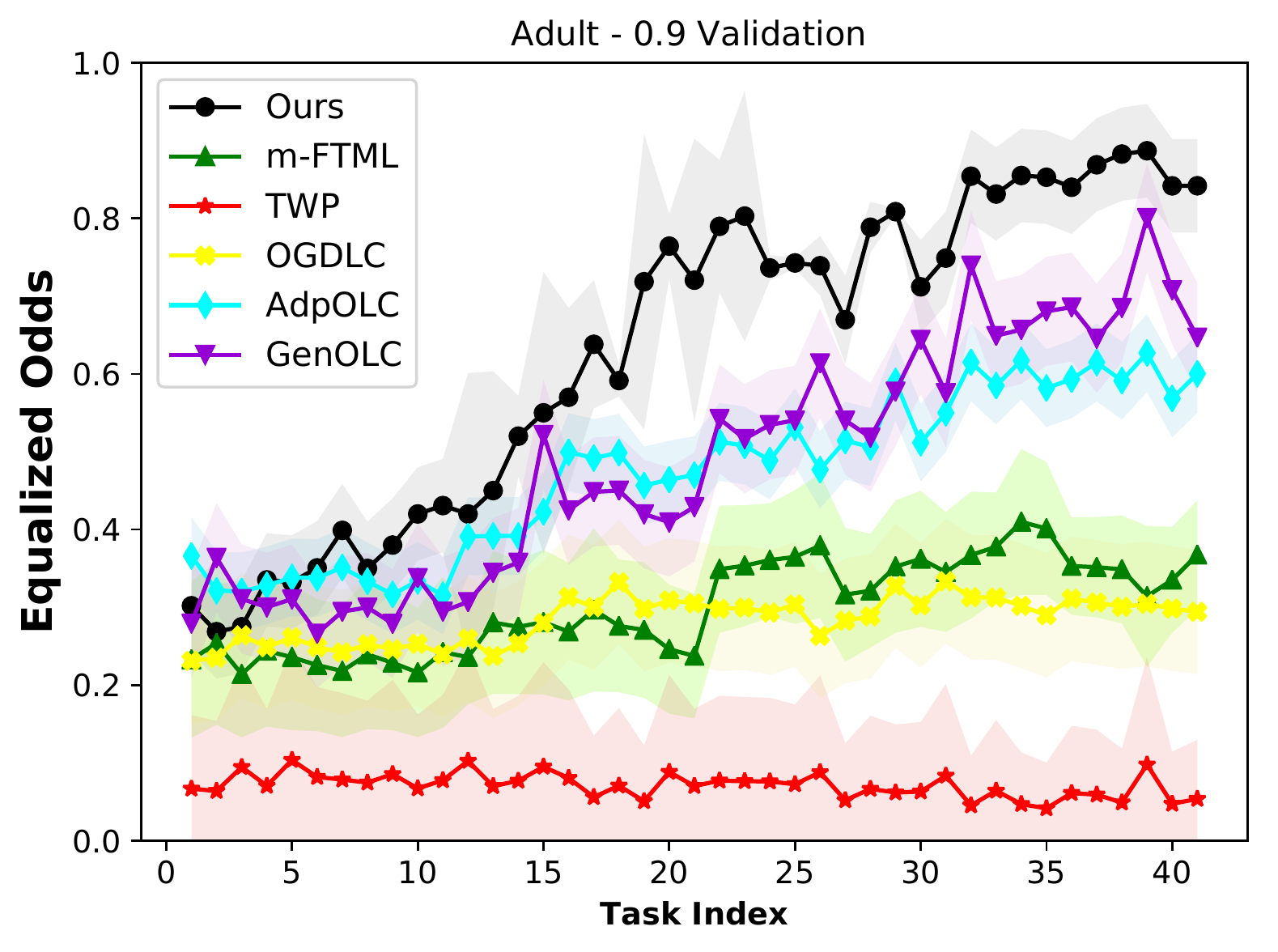}
        % \caption{}
    \end{subfigure}
    \begin{subfigure}[b]{0.235\textwidth}
        \includegraphics[width=\textwidth]{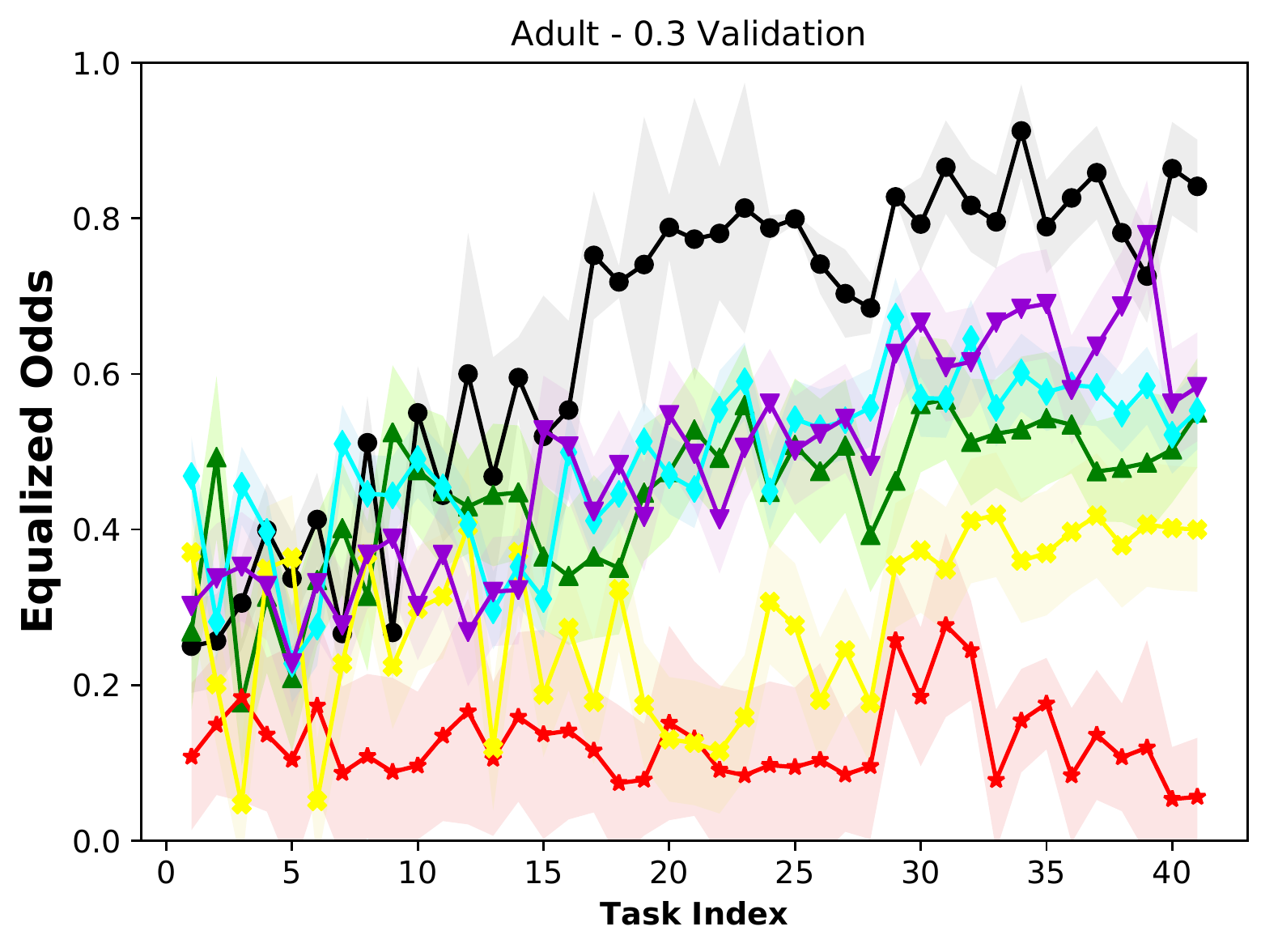}
        % \caption{}
    \end{subfigure}
    
    \begin{subfigure}[b]{0.235\textwidth}
        \includegraphics[width=\textwidth]{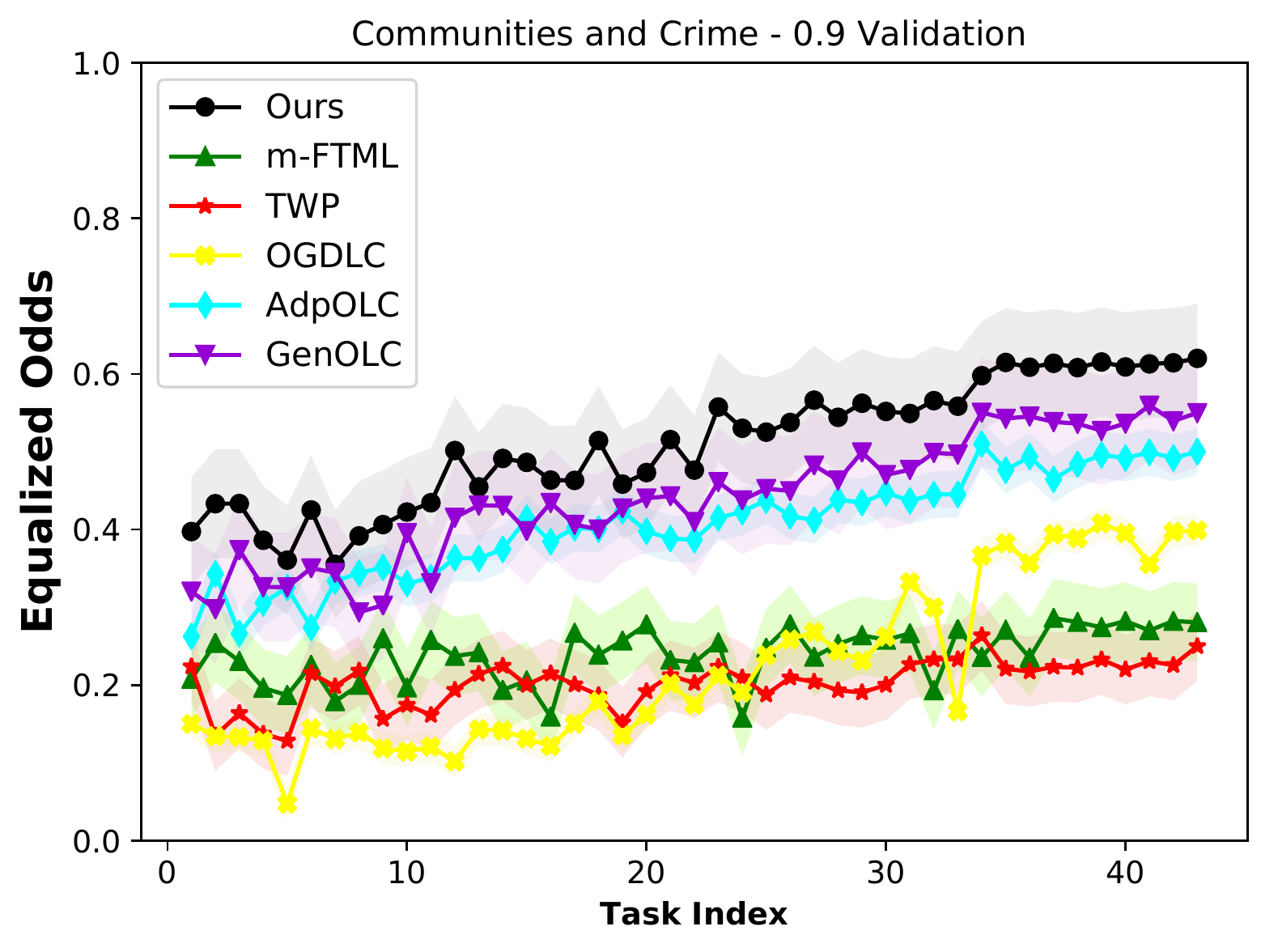}
        % \caption{}
    \end{subfigure}
    \begin{subfigure}[b]{0.235\textwidth}
        \includegraphics[width=\textwidth]{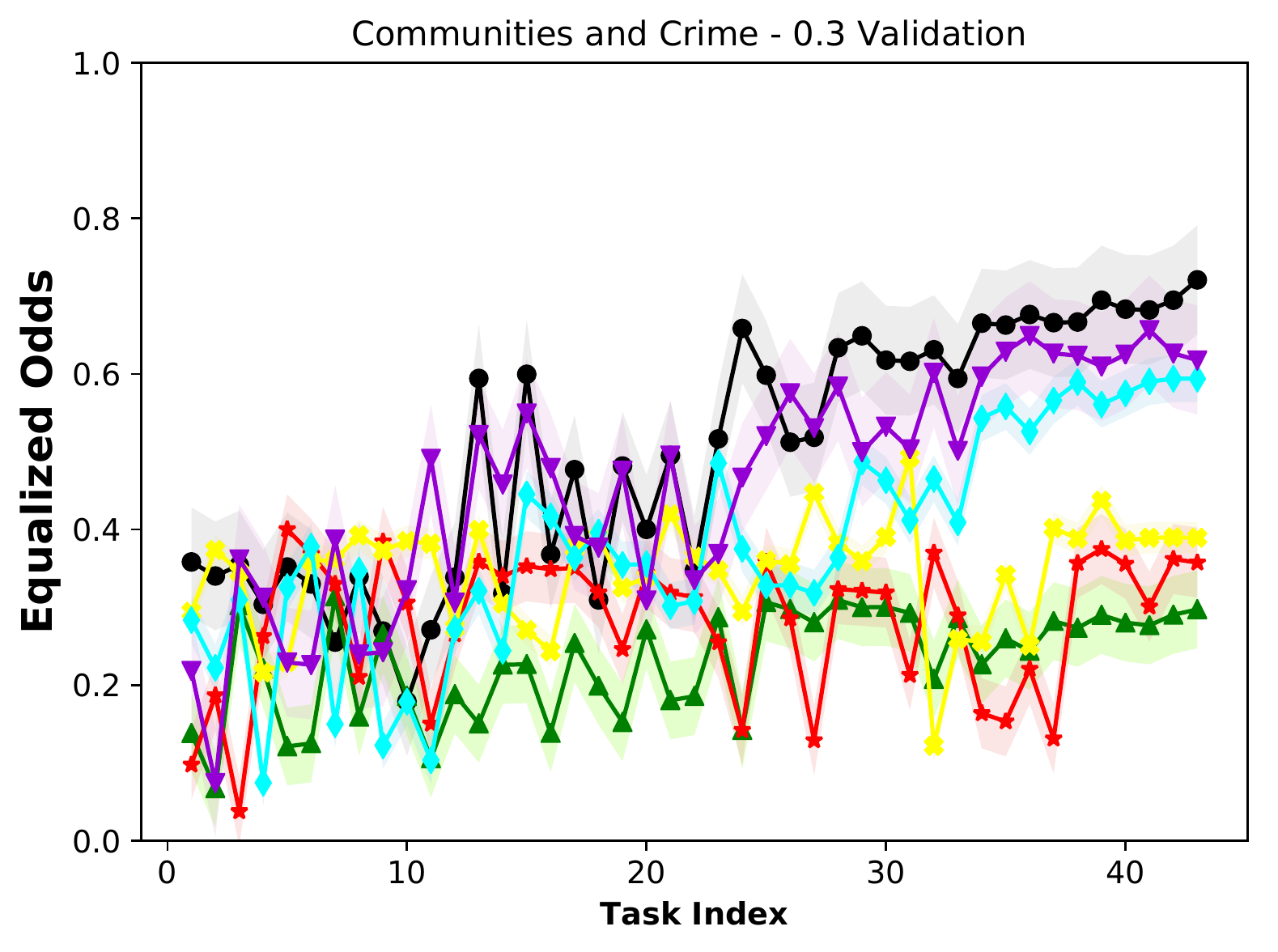}
        % \caption{}
    \end{subfigure}
    % \vspace{-3mm}
    \caption{Evaluation using fair metric EO at each round.}
    \label{fig:EOP-over-tasks}
% \vspace{-5mm}
\end{figure}

\subsection{Settings}

%  Here, we focus on the analysis of its empirical performance on deep learning models. 
% As demonstrated in our theoretical analysis in Section 5, our proposed method has nice theoretical properties for 

As discussed in Sec.\ref{sec:analysis}, the performance of our proposed method has been well justified theoretically for machine learning models, such as logistic regression and L2 regression, whose objectives that are strongly convex and smooth.
%In Sec.\ref{sec:analysis}, theoretically guarantees are derived under the assumption that the convexity always holds for all functions and variable domains. 
However, in machine learning and fairness studies, due to the non-linearity of neural networks, many problems have a non-convex landscape where theoretical analysis is challenging. Nevertheless, algorithms originally developed for convex optimization problems like gradient descent have shown promising results in practical non-convex settings \cite{Finn-ICML-2019}. Taking inspiration from these successes, we describe practical instantiations for the proposed online algorithm, and empirically evaluate the performance in Sec.\ref{sec:results}. 

% For simplicity, in our experiments, for each task we set the number of fairness constraints to one, \textit{i.e.} $m=1$. 
% Furthermore, at each round, $\mathcal{D}_t^S$, $\mathcal{D}_t^Q$ and $\mathcal{D}_t^V$ are sampled from $\mathcal{D}_t$ following the uniform sampling distribution, but other sampling distributions can be used if required. 
% Besides, note that the original MAML \cite{Finn-ICML-2017-(MAML)} is designed for quick adaption of few-shot learning. In our work, to study fairness, we consider control DBC as the optimization constraint. 
% In practice, we meta-train with support size of 100 for each sub-group, whereas evaluation may use hundreds of datapoints for some tasks.

% \vspace{-3mm}
\begin{figure*}[!t]
% \captionsetup[subfigure]{aboveskip=-2pt,belowskip=-2pt}
\centering
    \begin{subfigure}[b]{0.235\textwidth}
        \includegraphics[width=\textwidth]{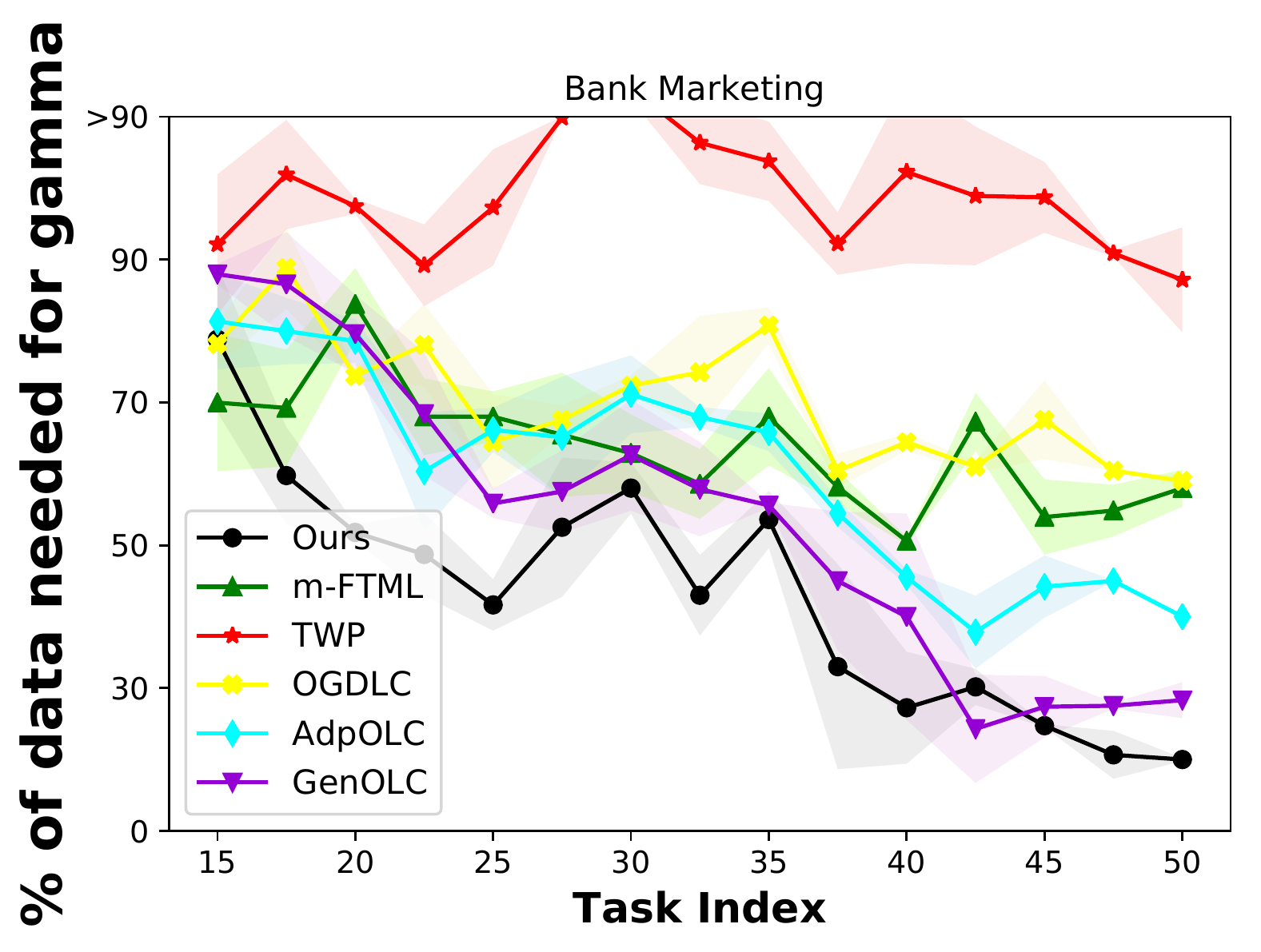}
        % \caption{}
    \end{subfigure}
    \begin{subfigure}[b]{0.235\textwidth}
        \includegraphics[width=\textwidth]{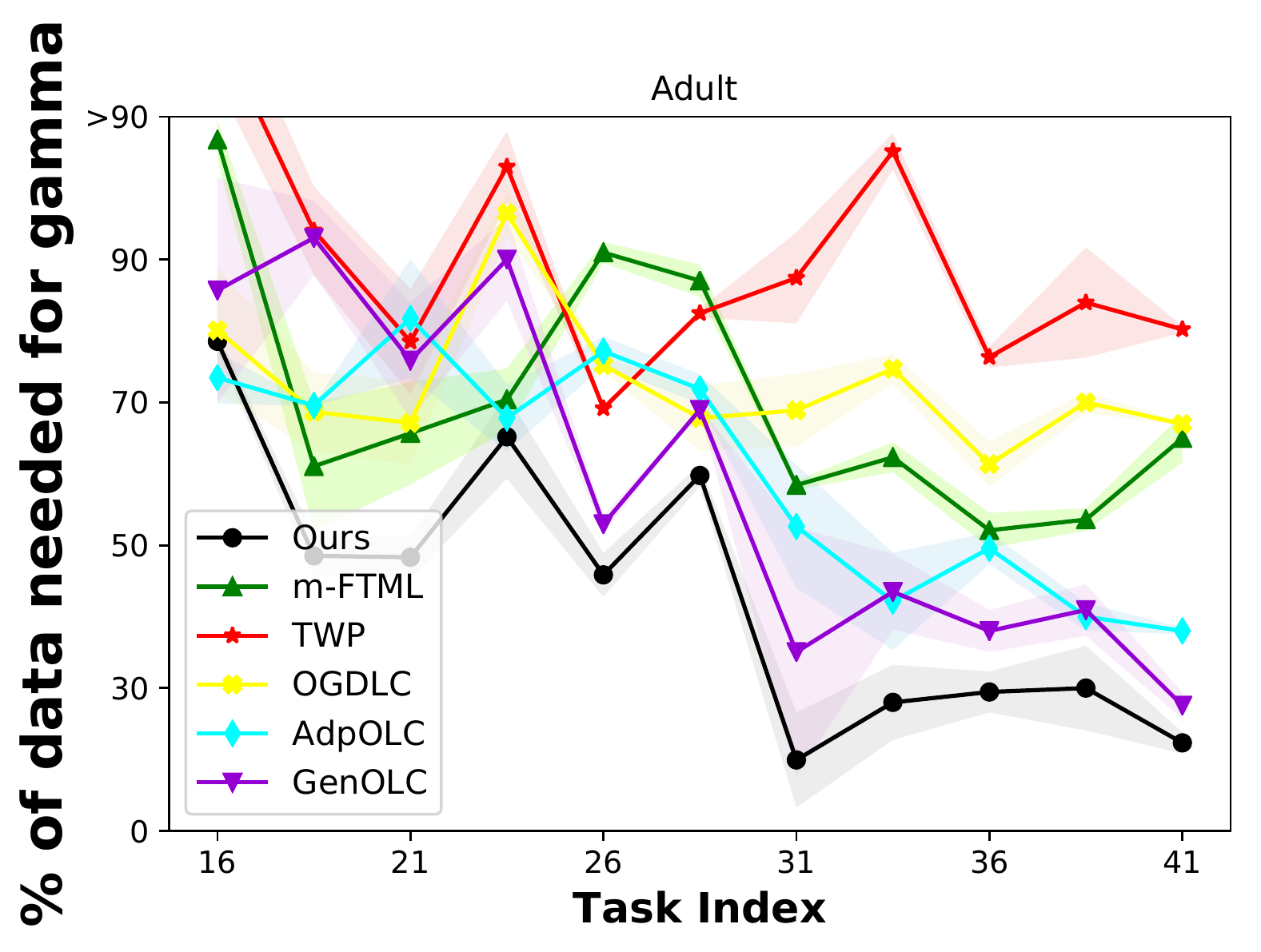}
        % \caption{}
    \end{subfigure}
    \begin{subfigure}[b]{0.235\textwidth}
        \includegraphics[width=\textwidth]{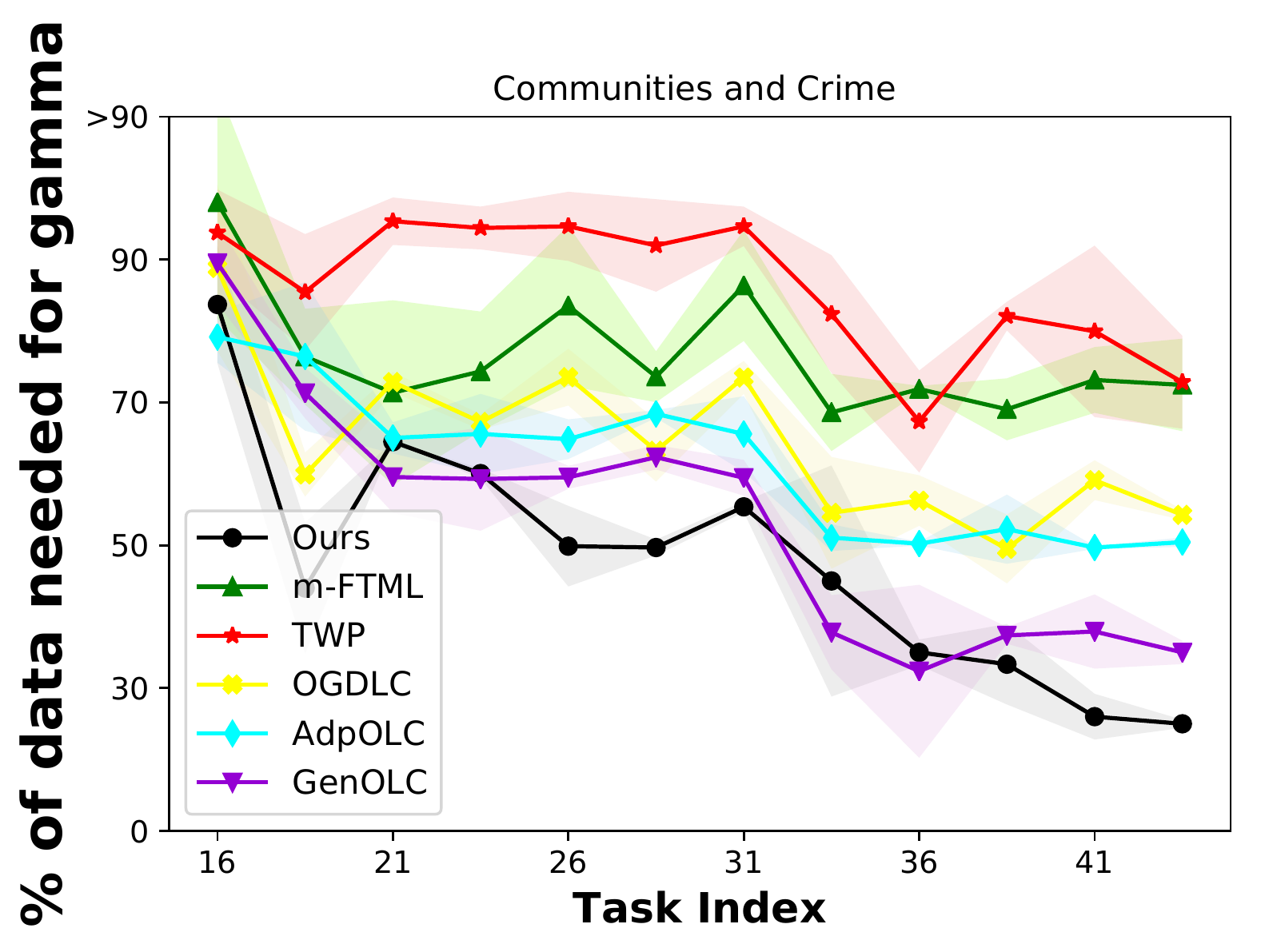}
        % \caption{}
    \end{subfigure}
    % \vspace{-8mm}
    \caption{Amount of data needed to learn each new task.}
    \label{fig:efficiency}
% \vspace{-5mm}
\end{figure*}

For each task we set the number of fairness constraints to one, \textit{i.e.} $m=1$. For the rest, we following the same settings as used in online meta learning~\cite{Finn-ICML-2019}. In particular, we meta-train with support size of 100 for each class, whereas $30\%$ or $90\%$ (hundreds of datapoints) of task samples for evaluation. All methods are completely online, \textit{i.e.}, all learning algorithms receive one task per-iteration. All the baseline models that are used to compare with our proposed approach share the same neural network architecture and parameter settings.
% Hyperparameters are selected by a held-out validation procedure. 
All the experiments are repeated 10 times with the same settings and the mean and standard deviation results are reported. 
% Due to the space limit, 
Details on the settings are given in Appendix \ref{App:ExpDetails}.

% When training, we use only one step gradient update (\textit{i.e.} $q=1$) and $k=10$ inner primal-dual updates with $2NK$ samples of query set ($N=1$ for regression setting and $N=2$ for classification), and a fixed primal and dual learning rate of $\gamma = 0.01$ and $\alpha = 0.01$. We use Adam as the meta-optimizer. Similarly, we set meta-learning rates of $\eta = 0.001$ and $\beta = 0.01$ used to update the meta-loss in the outer loop. 

% For all datasets, all the unprotected attributes are standardized to zero mean and unit variance and prepared for experiments. 

% Besides, taking few-shot learning into account, we set a meta batch-size of $8$ tasks and $4000$ meta-iterations for all datasets. Some key characteristics for all real data are listed in Table \ref{tab:key}.

% All baseline models used to compare with our proposed approach share the same neural network architecture and parameter settings. Hyperparameters are selected by a held-out validation procedure. All experiments are repeated 10 times with the same settings. Results shown with these methods in this paper are mean of experimental outputs.

%% file: results.tex
The following experimental results on each dataset are to answer all \textbf{Questions} given in Sec.\ref{sec:experiments}.
% and evaluates the effectiveness of the proposed approach and its competitors on a classification task. 
% For all baseline methods, wherever applicable, hyper-parameters were tuned via grid search. 
% We chose the models that were Pareto-optimal with regard to \textit{DBC} and all other evaluation metrics.

\subsection{End Task Performance}
In contrast to traditional machine learning paradigms where they often run a batch learning fashion, 
% online learning is a method for data arriving in a sequential order. 
online learning aims to learn and update the best predictor for future data at each round. 
In our experiments, we consider a sequential setting where a task comes one after another. To validate the effectiveness of the proposed algorithm, we first compare the end task performance. All methods stop at $t=T$ after seeing all tasks. The learned parameter pair $(\boldsymbol{\theta}_{T},\boldsymbol{\lambda}_{T})$ is further fine-tuned using the support set $\mathcal{D}_T^S$ which is sampled from the task $T$. The end task performance is hence evaluated on the validation set $\mathcal{D}_T^V$ using the adapted parameter $\boldsymbol{\theta}'_{T}$.

Consolidated and detailed performance of the different techniques over real-world data are listed in Table \ref{tab:endPerformance}. We evaluate performance across all competing methods on a scale of $90\%$ datapoints of the end task $T$ for each dataset. Best performance in each experimental unit are labeled in bold. We observe that as for bias-controlling, \sysname{} out-performs than other baseline methods. Specifically, \sysname{} has the highest scores in terms of the fairness metrics \textit{DP} and \textit{EO}, and the smallest value of \textit{Disc} close to zero signifies a fair prediction. Note that although \sysname{} returns a bit smaller predictive accuracy, this is due to the trade-off between losses and fairness.

\subsection{Performance Through Each Round}
% The goal of bias prevention in a online fashion with long-term constraints is to minimize the violation of unfairness notions over tasks. Note that, due to the relaxation of primal domain to $\mathcal{B}$, satisfaction of the fairness constraint at each round may not be guaranteed, but a sub-linear bound for the overall violation of cumulative constraints can be achieved. 
In order to take a closer look at the performance regarding bias-control in a non-stationary environment, at round $t\in[T]$, the parameter pair $(\boldsymbol{\theta}_{t},\boldsymbol{\lambda}_{t})$ inherited from the previous task $t-1$ are employed to evaluate the new task $t$. Inspired by \cite{Finn-ICML-2019}, we separately record the performance based on different amount (\textit{i.e.} $90\%$ and $30\%$) of validation samples.

Figure \ref{fig:DP-over-tasks} and \ref{fig:EOP-over-tasks} detail evaluation results across three real-world datasets at each round with respect to two wildly used fairness metrics \textit{DP} and \textit{EO}, respectively. Specifically, higher is better for all plots, while shaded regions show standard error computed using various random seeds. The learning curves show that with each new task added \sysname{} efficiently controls bias and substantially outperforms the alternative approaches in achieving the best fairness aware results represented by the highest $DP$ and $EO$ in final performance. GenOLC returns better results than AdpOLC and OGDLC since it applies both adaptive learning rates and squared clipped constraint term. However, two of the reasons giving rise to the performance of GenOLC inferior to \sysname{} is that our method (1) takes task-specific adaptation with respect to primal-dual parameter pair at inner loops, which further helps the task progress better as for fairness learning, (2) \sysname{} explicitly meta-trains and hence fully learns the structure across previous seen tasks. Although m-FTML shows an improvement in fairness, there is still substantial unfairness hidden in the data in the form of correlated attributes, which is consistent with \cite{Zemel-ICML-2013} and \cite{Lahoti-2019-ICDE}. As the most intuitive approach, theoretic analysis in \cite{OGDLC-2012-JMLR} shown the failure of using TWP is that the weight constant is fixed and independent from the sequences of solutions obtained so far.

Another observation is when evaluation data are reduced to $30\%$. Although the fairness performance becomes more fluctuant, \sysname{} remains the out-performance than other baseline methods through each round. This results from that in a limited amount of evaluated data our method stabilizes the results by learning parameters from all prior tasks so far at each round, and suggests that even better transfer can be accomplished through meta-learning.

\subsection{Task Learning Efficiency}
\label{sec:TaskLearningEfficiency}
To validate the learning efficiency of the proposed \sysname{}, we set a proficiency threshold $\boldsymbol{\gamma}$ for all methods at each round, where $\boldsymbol{\gamma} = (\gamma_1, \gamma_2)$ corresponds to the amount of data needed in $\mathcal{D}_t$ to achieve both a loss value $f_t(\boldsymbol{\theta}_t,\mathcal{D}_t^V)\leq\gamma_1$ and a \textit{DBC} value $g_t(\boldsymbol{\theta}_t,\mathcal{D}_t^V)\leq\gamma_2$ at the same time. We set $\gamma_1=0.0005$ and $\gamma_2=0.0001$ for all datasets. If less data is sufficient to reach the threshold, then priors learned from previous tasks are being useful and we have achieved positive transfer \cite{Finn-ICML-2019}. Through the results demonstrated in Figure \ref{fig:efficiency}, we observe while the baseline methods improve in efficiency over the course of learning as they see more tasks, they struggle to prevent negative transfer on each new task.

\subsection{Ablation Studies}
\begin{figure}[!h]
% \captionsetup[subfigure]{aboveskip=-2pt,belowskip=-2pt}
\centering
    \begin{subfigure}[b]{0.235\textwidth}
        \includegraphics[width=\textwidth]{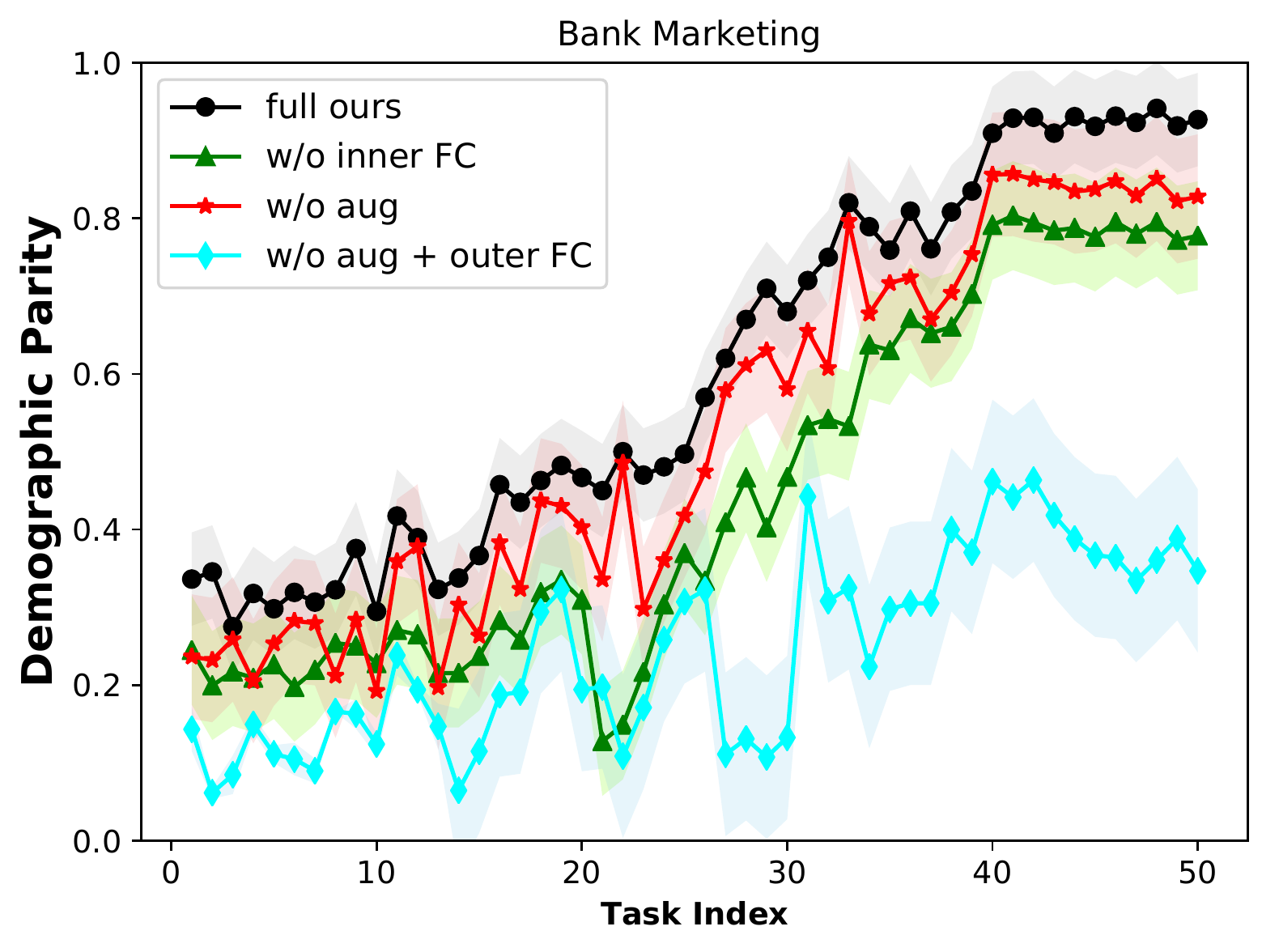}
        % \caption{}
    \end{subfigure}
    \begin{subfigure}[b]{0.235\textwidth}
        \includegraphics[width=\textwidth]{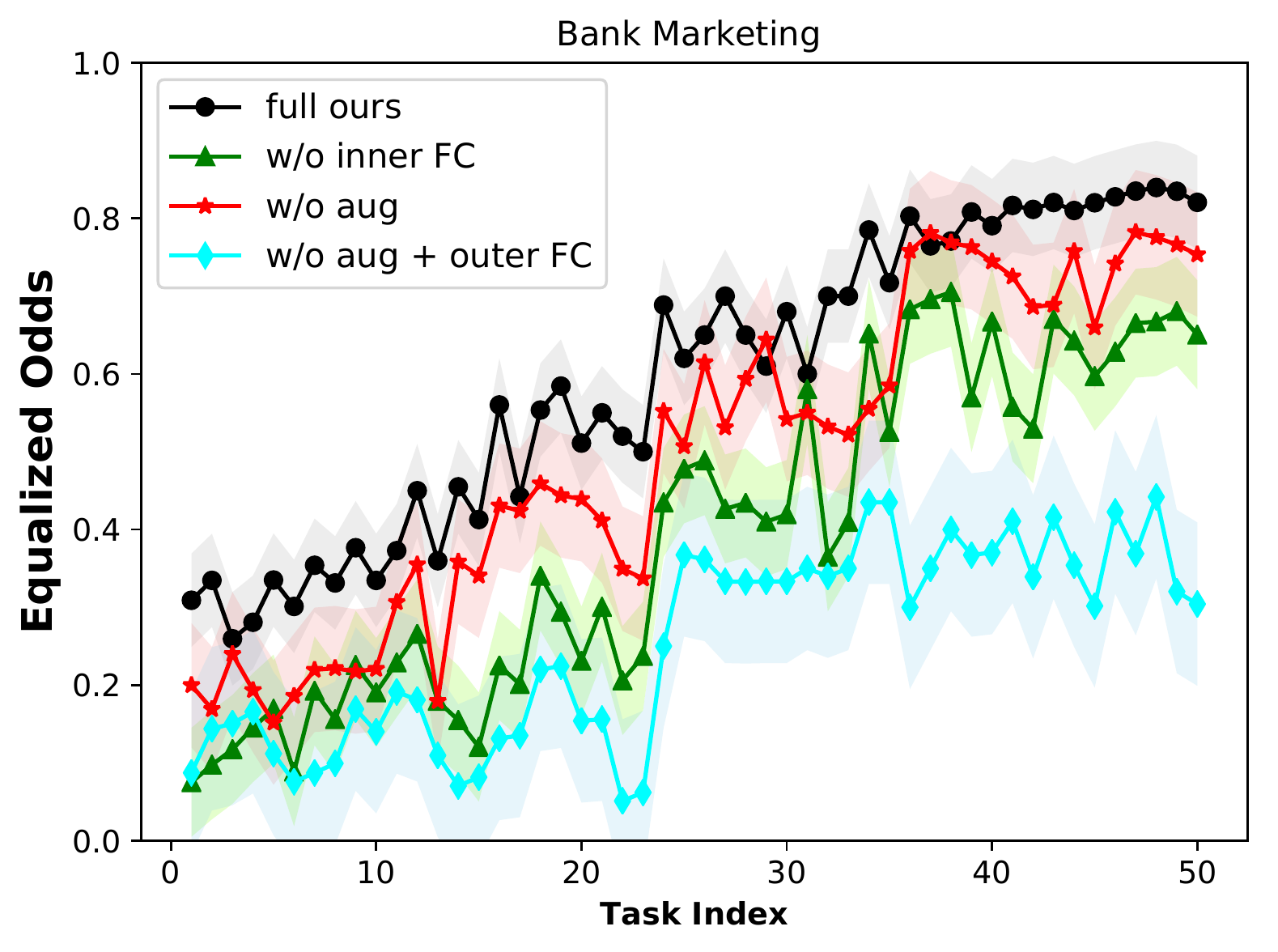}
        % \caption{}
    \end{subfigure}
    % \vspace{-8mm}
    \caption{Ablation study of our proposed models. (1) \textsf{w/o inner FC}: \sysname{} without inner fairness constraints; (2) \textsf{w/o aug}: \sysname{} without the augmented term in Eq.(\ref{eq:outer-Lagrangian}); (3) \textsf{w/o aug + outer FC}: \sysname{} without the augmented term and outer fairness constraints.}
    \label{fig:AS-bank}
% \vspace{-8mm}
\end{figure}

We conducted additional experiments to demonstrate the contributions of the three key technical components in \sysname{}:
the inner(task)-level fairness constraints (\textsf{inner FC}) in Eq.(\ref{eq:inner-problem}), the outer (meta)-level fairness constraints (\textsf{outer FC}) in Eq.(\ref{eq:inner-Lagrangian}), and the augmented term (\textsf{aug}) in Eq.(\ref{eq:outer-Lagrangian}).
%  $\delta\eta_2/2$
% that is used to prevent the dual parameters being to large
% In order to learn \sysname{}, fairness constraints in both inner (Eq.(\ref{eq:inner-problem})) and outer (Eq.(\ref{eq:outer-problem})) levels are indispensable. 
Particularly, \textsf{Inner FC} and \textsf{outer FC}  are used to regularize the task-level and metal-level loss functions, respectively, and 
%\sout{simply} considered as penalty terms added on corresponding \feng{task-level} loss functions,
%are used to regularize the meta-level loss function, 
%In the outer problem, loss functions with respect to query set are regularized with  
%\textsf{aug} in Eq.(\ref{eq:outer-Lagrangian}); 
\textsf{aug} is used to prevent the dual parameters being to large and hence stabilizes the learning with fairness in the outer problem.
The key findings in Figure \ref{fig:AS-bank} are (1) inner fairness constraints and the augmented term  can enhance bias control, and (2) outer update procedure plays more important role in \sysname{}. This is due to the close-form projection onto the relaxed domain $\mathcal{B}$ with respect to primal variable and clipped non-negative dual variable.
% \vspace{-3mm}

%% file: conclusion.tex
In this paper, we formulate the problem of online fairness-aware meta learning and present a novel algorithm, namely \sysname{}. 
% We conduct both theoretical and empirical analyses extensively to demonstrate the efficiency and effectiveness of \sysname{}. 
%which efficiently controls fairness in a online meta-learning paradigm. 
We claim that for the first time a fairness-aware online meta-learning framework is proposed. 
The goal of this model is to minimize both the loss regret and violation of long-term fairness constraints as $t$ increases, and to achieve sub-linear bound for them. 
Specifically, in stead of learning primal parameters only at each round, \sysname{} trains a meta-parameter pair including primal and dual variables, where the primal variable determines the predictive accuracy and the dual variable controls the level of satisfaction of model fairness. 
To determine the parameter pair at each round, we formulate the problem to a bi-level convex-concave optimization problem. 
Detailed theoretic analysis and corresponding proofs justify the efficiency and effectiveness of the proposed algorithm by demonstrating upper bounds for regret and violation of fairness constraints. Experimental evaluation based on three real-world datasets shows that our method out-performs than state-of-the-art online learning techniques with long-term fairness constraints in bias-controlling. 
It remains interesting if one can prove that fairness constraints are satisfied at each round without approximated projections onto the relaxed domain, and if one can explore learning when environment is changing over time.

%% file: App_proofT1.tex
\begin{proof}
Before giving the proof of Theorem \ref{theorem1}, let us define a function $\Tilde{\mathcal{L}}(\boldsymbol{\theta,\lambda}):(\mathcal{B}\times\mathbb{R}^m_+)\rightarrow\mathbb{R}$ with respect to the primal and dual variable $\boldsymbol{\theta}$ and $\boldsymbol{\lambda}$. Therefore, $\Tilde{\mathcal{L}}(\boldsymbol{\theta,\lambda})$ is considered as a single task function $k\in[t]$ from prior experience and it is stripped from Eq.(\ref{eq:outer-Lagrangian}).
\begin{align*}
    \Tilde{\mathcal{L}}(\boldsymbol{\theta,\lambda})&:= \mathcal{L}'\Big(\mathcal{A}lg(\boldsymbol{\theta}),\boldsymbol{\lambda}\Big)\\
    &=f(\mathcal{A}lg(\boldsymbol{\theta}))+\sum\nolimits_{i=1}^m\Big( \lambda_i g_i(\mathcal{A}lg(\boldsymbol{\theta})) - \frac{\delta\eta_2}{2}\lambda)_i^2\Big)
\end{align*}
where $\mathcal{A}lg(\boldsymbol{\theta})$ is defined in Eq.(\ref{eq:task-level primal-dual}). We hence first prove $\Tilde{\mathcal{L}}(\boldsymbol{\cdot,\lambda})$ is convex with respect to $\boldsymbol{\theta}$. Similar results can be easily achieved with respect to $\boldsymbol{\lambda}$.
We consider two arbitrary point $\boldsymbol{\theta,\phi}\in\mathcal{B}$ with respect to the primal variable:
\begin{align*}
    &||\nabla_{\boldsymbol{\theta}}\Tilde{\mathcal{L}}(\boldsymbol{\theta})-\nabla_{\boldsymbol{\phi}}\Tilde{\mathcal{L}}(\boldsymbol{\phi})||\\
    % =&||\nabla_{\boldsymbol{\theta}'}\mathcal{L}'(\mathcal{A}lg(\boldsymbol{\theta}))\nabla_{\boldsymbol{\theta}}\mathcal{A}lg(\boldsymbol{\theta})-\nabla_{\boldsymbol{\phi}'}\mathcal{L}'(\mathcal{A}lg(\boldsymbol{\phi}))\nabla_{\boldsymbol{\phi}}\mathcal{A}lg(\boldsymbol{\phi})||\\
    \leq& \underbrace{||\nabla_{\boldsymbol{\theta}'}\mathcal{L}'(\mathcal{A}lg(\boldsymbol{\theta}))(\nabla_{\boldsymbol{\theta}}\mathcal{A}lg(\boldsymbol{\theta})-\nabla_{\boldsymbol{\phi}}\mathcal{A}lg(\boldsymbol{\phi}))||}_{\textbf{First Term (FT)}} + \\
    &\underbrace{||\nabla_{\boldsymbol{\phi}}\mathcal{A}lg(\boldsymbol{\phi})(\nabla_{\boldsymbol{\theta}'}\mathcal{L}'(\mathcal{A}lg(\boldsymbol{\theta}))-\nabla_{\boldsymbol{\phi}'}\mathcal{L}'(\mathcal{A}lg(\boldsymbol{\phi})))||}_{\textbf{Second Term (ST)}}
\end{align*}
We then bound the first and second terms that:
\begin{align*}
    \text{FT} 
    \leq&\eta_1||\nabla_{\boldsymbol{\theta}'}\mathcal{L}'(\mathcal{A}lg(\boldsymbol{\theta}))||||(\nabla^2_{\boldsymbol{\theta}}f(\boldsymbol{\theta})-\nabla^2_{\boldsymbol{\phi}}f(\boldsymbol{\phi}))+\\
    &(\nabla^2_{\boldsymbol{\theta}}\sum\nolimits_{i=1}^m \lambda_i g_i(\boldsymbol{\theta})-\nabla^2_{\boldsymbol{\phi}}\sum\nolimits_{j=1}^m\lambda_j g_j(\boldsymbol{\phi}))||\\
    \leq &\eta_1(L_f+\Bar{\lambda}mL_g)(\rho_f+\Bar{\lambda}m\rho_g)||\boldsymbol{\theta}-\boldsymbol{\phi}||\\
    \text{ST}
    =&||(\boldsymbol{I}-\eta_1(\nabla^2_{\boldsymbol{\phi}}f(\boldsymbol{\phi})+\nabla^2_{\boldsymbol{\phi}}\sum\nolimits_{i=1}^m\lambda_i g_i(\boldsymbol{\phi})))\\
    &(\nabla_{\boldsymbol{\theta}'}\mathcal{L}'(\mathcal{A}lg(\boldsymbol{\theta}))-\nabla_{\boldsymbol{\phi}'}\mathcal{L}'(\mathcal{A}lg(\boldsymbol{\phi})))||\\
    % \leq&(1-\eta_1(\mu_f+\Bar{\lambda}m\mu_g))(\beta_f+\Bar{\lambda}m\beta_g)||\mathcal{A}lg(\boldsymbol{\theta})-\mathcal{A}lg(\boldsymbol{\phi})||\\
    \leq&(1-\eta_1(\mu_f+\Bar{\lambda}m\mu_g))^2(\beta_f+\Bar{\lambda}m\beta_g)||\boldsymbol{\theta}-\boldsymbol{\phi}||
\end{align*}
% \begin{align*}
%     \text{FT} \leq &\eta_1(L_f+\Bar{\lambda}mL_g)(\rho_f+\Bar{\lambda}m\rho_g)||\boldsymbol{\theta}-\boldsymbol{\phi}||\\
%     \text{ST}\leq&(1-\eta_1(\mu_f+\Bar{\lambda}m\mu_g))^2(\beta_f+\Bar{\lambda}m\beta_g)||\boldsymbol{\theta}-\boldsymbol{\phi}||
% \end{align*}
The 
% last 
inequality to bound ST is due to the Lemma 2 to 4 in \cite{Finn-ICML-2019}. 
% Here we approximate $h(\boldsymbol{\lambda})\approx \boldsymbol{\lambda}$. 
Together the upper bounds for the first and the second terms and choose step size
\begin{align*}
    \eta_1\leq\min\Big\{ \frac{\mu_f+\Bar{\lambda}m\mu_g}{8(L_f+\Bar{\lambda}mL_g)(\rho_f+\Bar{\lambda}m\rho_g)}, \frac{1}{2(\beta_f+\Bar{\lambda}m\beta_g)} \Big\}
\end{align*}
Then we have
\begin{align*}
    &||\nabla_{\boldsymbol{\theta}}\Tilde{\mathcal{L}}(\boldsymbol{\theta})-\nabla_{\boldsymbol{\phi}}\Tilde{\mathcal{L}}(\boldsymbol{\phi})|| 
    % \leq \Big\{\eta_1(L_f+\Bar{\lambda}mL_g)(\rho_f+\Bar{\lambda}m\rho_g)\\
    % &+(1-\eta_1(\mu_f+\Bar{\lambda}m\mu_g))^2(\beta_f+\Bar{\lambda}m\beta_g)\Big\}||\boldsymbol{\theta}-\boldsymbol{\phi}||\\
    % &\leq ( (\mu_f+\Bar{\lambda}m\mu_g)/8+(\beta_f+\Bar{\lambda}m\beta_g))||\boldsymbol{\theta}-\boldsymbol{\phi}||\\
    \leq \frac{9}{8}(\beta_f+\Bar{\lambda}m\beta_g)||\boldsymbol{\theta}-\boldsymbol{\phi}||
\end{align*}
Therefore $\Tilde{\mathcal{L}}(\cdot,\boldsymbol{\lambda})$ is $\frac{9}{8}(\beta_f+\Bar{\lambda}m\beta_g)$-smooth. Next to achieve lower bound for $\Tilde{\mathcal{L}}(\cdot,\boldsymbol{\lambda})$
\begin{align*}
    &||\nabla_{\boldsymbol{\theta}}\Tilde{\mathcal{L}}(\boldsymbol{\theta})-\nabla_{\boldsymbol{\phi}}\Tilde{\mathcal{L}}(\boldsymbol{\phi})||\\
    \geq& \underbrace{||\nabla_{\boldsymbol{\phi}}\mathcal{A}lg(\boldsymbol{\phi})(\nabla_{\boldsymbol{\theta}'}\mathcal{L}'(\mathcal{A}lg(\boldsymbol{\theta}))-\nabla_{\boldsymbol{\phi}'}\mathcal{L}'(\mathcal{A}lg(\boldsymbol{\phi})))||}_{\textbf{Third Term (TT)}} -\\
    &||\nabla_{\boldsymbol{\theta}'}\mathcal{L}'(\mathcal{A}lg(\boldsymbol{\theta}))(\nabla_{\boldsymbol{\theta}}\mathcal{A}lg(\boldsymbol{\theta})-\nabla_{\boldsymbol{\phi}}\mathcal{A}lg(\boldsymbol{\phi}))||
\end{align*}
The second term in the above inequality is the same as the FT. We hence bound the TT that:
\begin{align*}
    \text{TT}
    % &\geq (1-\eta_1(\beta_f+\Bar{\lambda}m\beta_g))(\mu_f+\Bar{\lambda}m\mu_g)||\mathcal{A}lg(\boldsymbol{\theta})-\mathcal{A}lg(\boldsymbol{\phi})|| \\
    &\geq (1-\eta_1(\beta_f+\Bar{\lambda}m\beta_g))^2(\mu_f+\Bar{\lambda}m\mu_g)||\boldsymbol{\theta}-\boldsymbol{\phi}||\\
    &\geq \frac{\mu_f+\Bar{\lambda}m\mu_g}{4}||\boldsymbol{\theta}-\boldsymbol{\phi}||
\end{align*}
Together TT and FT, we derive the lower bound
\begin{align*}
    ||\nabla_{\boldsymbol{\theta}}\Tilde{\mathcal{L}}(\boldsymbol{\theta})-\nabla_{\boldsymbol{\phi}}\Tilde{\mathcal{L}}(\boldsymbol{\phi})||
    % &\geq \Big( \frac{\mu_f+\Bar{\lambda}m\mu_g}{4} -\frac{\mu_f+\Bar{\lambda}m\mu_g}{8}\Big)||\boldsymbol{\theta}-\boldsymbol{\phi}||\\
    \geq\frac{\mu_f+\Bar{\lambda}m\mu_g}{8}||\boldsymbol{\theta}-\boldsymbol{\phi}||
\end{align*}
Thus, $\Tilde{\mathcal{L}}(\cdot,\boldsymbol{\lambda})$ is $\frac{1}{8}(\mu_f+\Bar{\lambda}m\mu_g)$-strongly convex. Since $\Tilde{\mathcal{L}}(\cdot,\boldsymbol{\lambda})$ is convex, summation of convex functions with non-negative weights preserves convexity and summation of strongly convex functions is strongly convex. Therefore, we complete the proof for $\mathcal{L}(\cdot,\boldsymbol{\lambda})$.
\end{proof}

%% file: App_proofT2.tex
In order to better understand Theorem \ref{theorem2}, we first introduce Lemma \ref{lemma} and its analysis is modified and analogous to that developed in \cite{OGDLC-2012-JMLR}.
\begin{lemma}
\label{lemma}
Let $\mathcal{L}_t(\cdot,\cdot)$ be the function defined in Eq.(\ref{eq:outer-Lagrangian}), which is convex in its first argument and concave in its second argument. Let $\boldsymbol{\theta}_t$ and $\boldsymbol{\lambda}_t, t\in[T]$ be the sequence of solution obtained by Algorithm \ref{alg:PDRFTML}. Then for any $(\boldsymbol{\theta,\lambda})\in\mathcal{B}\times\mathbb{R}^m_+$, we have
\begin{align}
\label{eq:lemma_bound}
    &\sum\nolimits_{t=1}^T \mathcal{L}_t(\boldsymbol{\theta}_t, \boldsymbol{\lambda})-\mathcal{L}_t(\boldsymbol{\theta}, \boldsymbol{\lambda}_t) \leq \frac{R^2+||\boldsymbol{\lambda}||^2}{2\eta_2} - \frac{\mu_f}{2}R^2\\
    &+\sum\nolimits_{t=1}^T\Big\{\frac{\eta_2}{2}\Big(4G^2\eta_1^2H^2(m+1)^2
    % +2\zeta^2
    +4m(D^2+\eta_1^2G^4)\Big)\nonumber \\
    &+2\eta_2m(G^4\eta_1^2m^2+\delta^2\eta_2^2)||\boldsymbol{\lambda}_t||^2
    +2\eta_2G^2\eta_1^2H^2(m+1)^2||\boldsymbol{\lambda}_t||^4 \Big\} \nonumber
\end{align}
\end{lemma}
\begin{proof}
Following the Assumption \ref{assmp3} and analysis of the Lemma 2 in \cite{OGDLC-2012-JMLR}, we derive
\begin{align*}
    &\mathcal{L}_t(\boldsymbol{\theta}_t,\boldsymbol{\lambda}) - \mathcal{L}_t(\boldsymbol{\theta},\boldsymbol{\lambda}_t) \leq \frac{1}{2\eta_2}\Big(||\boldsymbol{\theta}-\boldsymbol{\theta}_t||^2+||\boldsymbol{\lambda}-\boldsymbol{\lambda}_t||^2\\
    &-||\boldsymbol{\theta}-\boldsymbol{\theta}_{t+1}||^2-||\boldsymbol{\lambda}-\boldsymbol{\lambda}_{t+1}||^2 \Big) + \frac{\eta_2}{2}\Big(||\nabla_{\boldsymbol{\theta}}\mathcal{L}_t(\boldsymbol{\theta}_t,\boldsymbol{\lambda}_t)||^2\\
    &+\nabla_{\boldsymbol{\lambda}}\mathcal{L}_t|\boldsymbol{\theta}_t,\boldsymbol{\lambda}_t||^2\Big) - \frac{\mu_f}{2}||\boldsymbol{\theta}-\boldsymbol{\theta}_t||^2
\end{align*}
We bound $||\nabla_{\boldsymbol{\theta}}\mathcal{L}_t(\boldsymbol{\theta}_t,\boldsymbol{\lambda}_t)||^2\leq4G^2\eta_1^2 H^2(m+1)^2(1+||\boldsymbol{\lambda}_t||^4)$ and $\nabla_{\boldsymbol{\lambda}}\mathcal{L}_t|\boldsymbol{\theta}_t,\boldsymbol{\lambda}_t||^2\leq4m(D^2+G^4\eta_1^2m^2||\boldsymbol{\lambda}_t||^2+\delta^2\eta_2^2||\boldsymbol{\lambda}_t||^2+\eta_1^2G^4)$ using the inequality $(a_1+a_2+\cdots+a_n)^2\leq n(a_1^2+a_2^2+\cdots+a_n^2)$. 
% As for the regularization term $\mathcal{F}(\boldsymbol{\theta})$, it simply takes the Euclidean norm $||\boldsymbol{\theta}||$.
% \begin{align*}
%     \text{FT}
%     \leq &||\Big\langle\nabla_{\boldsymbol{\theta}'}f(\boldsymbol{\theta}'_t),\nabla_{\boldsymbol{\theta}}\mathcal{A}lg(\boldsymbol{\theta}_t)\Big\rangle+\sum\nolimits_{i=1}^m\lambda_{t,i}\Big\langle\nabla_{\boldsymbol{\theta}'}g_i(\boldsymbol{\theta}'_t),\\
%     &\nabla_{\boldsymbol{\theta}}\mathcal{A}lg(\boldsymbol{\theta}_t)\Big\rangle
%     % +\zeta
%     ||^2\\
%     \leq&\Big(G(1-\eta_1 H(1+\sum\nolimits_{i=1}^m\lambda_{t,i}))(1+\sum\nolimits_{i=1}^m\lambda_{t,i})
%     % +\zeta
%     \Big)^2\\
%     \leq&4G^2\eta_1^2 H^2(m+1)^2(1+||\boldsymbol{\lambda}_t||^4)
%     % +2\zeta^2
% \end{align*}
% and the second term
% \begin{align*}
%     \text{ST}
%     \leq&||\Big\{\Big\langle\nabla_{\boldsymbol{\theta}'}f(\boldsymbol{\theta}'),\nabla_{\boldsymbol{\lambda}}\mathcal{A}lg(\boldsymbol{\theta})\Big\rangle+\sum\nolimits_{i=1}^m\Big(g_i(\boldsymbol{\theta}')+\\
%     &\lambda_{t,i}\Big\langle\nabla_{\boldsymbol{\theta}'}g_i(\boldsymbol{\theta}'),\nabla_{\boldsymbol{\lambda}}\mathcal{A}lg(\boldsymbol{\theta})\Big\rangle-\delta\eta_2\lambda_{t,i}\Big)\Big\}||^2\\
%     \leq&\Big(\sum\nolimits_{i=1}^m(D-\lambda_{t,i}G^2\eta_1 m -\delta\eta_2\lambda_{t,i} -\eta_1G^2)\Big)^2\\
%     \leq&4m(D^2+G^4\eta_1^2m^2||\boldsymbol{\lambda}_t||^2+\delta^2\eta_2^2||\boldsymbol{\lambda}_t||^2+\eta_1^2G^4)
% \end{align*}
By adding the inequalities of FT and ST, and using the fact $||\boldsymbol{\theta}||\leq R$ and $t\geq 1$ we complete the proof.
\end{proof}
By applying Lemma \ref{lemma}, we now prove Theorem \ref{theorem2}.
\begin{proof}
By expanding Eq.(\ref{eq:lemma_bound}) using Eq.(\ref{eq:outer-Lagrangian}), and in short we use \textit{RHS} to substitute the right-hand side of the inequality of Eq.(\ref{eq:lemma_bound})
% \begin{align*}
%     &\sum\nolimits_{t=1}^T\frac{1}{t}\sum\nolimits_{k=1}^t\Big\{f_k(\mathcal{A}lg_k(\boldsymbol{\theta}_k))-f_k(\mathcal{A}lg_k(\boldsymbol{\theta}))\\
%     &+\sum\nolimits_{i=1}^m\Big(\lambda_i g_i(\mathcal{A}lg_k(\boldsymbol{\theta}_k))
%     -\lambda_{t,i} g_i(\mathcal{A}lg_k(\boldsymbol{\theta}))\Big)-\frac{\delta\eta_2}{2}\lambda_k^2\\
%     &+\frac{\delta\eta_2}{2}\lambda_{t,i}^2\Big\}
%     % +\zeta\sum_{t=1}^T(||\boldsymbol{\theta}_t||-||\boldsymbol{\theta}||)\\
%     \leq RHS
% \end{align*}
and set $\boldsymbol{\theta}=\boldsymbol{\theta}^*$. Following the Theorem 3.1 in \cite{Cambridge-book-2006}, we have
\begin{align*}
    &\sum\nolimits_{t=1}^T\Big\{f_t(\mathcal{A}lg_t(\boldsymbol{\theta}_t))-f_t(\mathcal{A}lg_t(\boldsymbol{\theta}^*))\Big\} \\
    &+\sum\nolimits_{i=1}^m\Big\{\lambda_i\sum\nolimits_{t=1}^T g_i(\mathcal{A}lg_t(\boldsymbol{\theta}_t))-\sum\nolimits_{t=1}^T\lambda_{t,i}g_i(\mathcal{A}lg_t(\boldsymbol{\theta}^*))\Big\}\\
    &-\frac{\delta\eta_2 T}{2}||\boldsymbol{\lambda}||^2+\frac{\delta\eta_2}{2}\sum\nolimits_{t=1}^T||\boldsymbol{\lambda}||^2
    % -\zeta TR\\
    \leq RHS
\end{align*}
Since $\delta\geq\max\{4m(G^4\eta_1^2m^2+ \delta^2\eta_2^2),4G^2\eta_1^2H^2(m+1)^2\}$, we can drop terms containing $||\boldsymbol{\lambda}_t||^2$ and $||\boldsymbol{\lambda}_t||^4$.
% , and obtain
% \begin{align*}
%     &\sum\nolimits_{t=1}^T\Big\{f_t(\mathcal{A}lg_t(\boldsymbol{\theta}_t))-f_t(\mathcal{A}lg_t(\boldsymbol{\theta}^*))\Big\}+\sum\nolimits_{i=1}^m\Big\{\lambda_i\sum\nolimits_{t=1}^T g_i(\mathcal{A}lg_t(\boldsymbol{\theta}_t))\\
%     &-\frac{\delta\eta_2 T}{2}\lambda_i^2-\frac{m}{2\eta_2}\lambda_i^2\Big\}\leq\sum\nolimits_{i=1}^m\sum\nolimits_{t=1}^T\lambda_{t,i}g_i(\mathcal{A}lg_t(\boldsymbol{\theta}^*))+\frac{R^2}{2\eta_2}-\frac{\mu_f}{2}R^2\\
%     &\sum\nolimits_{t=1}^T\frac{\eta_2}{2}\Big(4G^2\eta_1^2H^2(m+1)^2+2\zeta^2+4m(D^2+\eta_1^2G^4)\Big)
%     % +\zeta TR
% \end{align*}
By taking maximization for $\boldsymbol{\lambda}$ over $(0,+\infty)$, we get
\begin{align*}
    &\sum\nolimits_{t=1}^T\Big\{f_t(\mathcal{A}lg_t(\boldsymbol{\theta}_t))-f_t(\mathcal{A}lg_t(\boldsymbol{\theta}^*))\Big\}\\
    &+\sum\nolimits_{i=1}^m\Big\{\frac{\Big[\sum\nolimits_{t=1}^T g_i(\mathcal{A}lg_t(\boldsymbol{\theta}_t))\Big]^2_+}{2(\delta\eta_2 T+\frac{m}{\eta_2})}-\sum\nolimits_{t=1}^T\lambda_{t,i}g_i(\mathcal{A}lg_t(\boldsymbol{\theta}^*))\Big\}\\
    &\leq\frac{R^2}{2\eta_2}-\frac{\mu_f}{2}R^2\\
    % +\zeta TR\\
    &+\Big(4G^2\eta_1^2H^2(m+1)^2+2\zeta^2+4m(D^2+\eta_1^2G^4)\Big)\sum\nolimits_{t=1}^T\frac{\eta_2}{2}
\end{align*}
Since $g_i(\mathcal{A}lg(\boldsymbol{\theta}^*))\leq 0$ and $\lambda_{t,i}\geq 0, \forall i\in[m]$.
For $f_t(\boldsymbol{\theta})$ to be strongly convex, in order to have lower upper bounds for both objective regret and the long-term constraint, we need to use time-varying stepsize as the one used in \cite{GenOLC-2018-NeurIPS}, that is $\eta_2=\mu_f/(t+1)$. 
% \begin{align*}
%     &\sum\nolimits_{t=1}^T\Big\{f_t(\mathcal{A}lg_t(\boldsymbol{\theta}_t))-f_t(\mathcal{A}lg_t(\boldsymbol{\theta}^*))\Big\}+\sum\nolimits_{i=1}^m\frac{\Big[\sum\nolimits_{t=1}^T g_i(\mathcal{A}lg_t(\boldsymbol{\theta}_t))\Big]^2_+}{2(\delta\eta_2 T+\frac{m}{\eta_2})}\\
%     &\leq \frac{R^2}{\mu_f}+\Big(2G^2\eta_1^2H^2(m+1)^2+\zeta^2+2m(D^2+\eta_1^2G^4)\Big)\sum\nolimits_{t=1}^T\frac{\mu_f}{t+1}\\
%     &\leq O(\log T)
%     % &+\zeta TR 
% \end{align*}
Due to non-negative of $\frac{\Big[\sum\nolimits_{t=1}^T g_i(\mathcal{A}lg_t(\boldsymbol{\theta}_t))\Big]^2_+}{2(\delta\eta_2 T+\frac{m}{\eta_2})}$, we have
\begin{align*}
    \sum\nolimits_{t=1}^T\Big\{f_t(\mathcal{A}lg_t(\boldsymbol{\theta}_t))-f_t(\mathcal{A}lg_t(\boldsymbol{\theta}^*))\Big\} \leq O(\log T)
\end{align*}
According to the assumption, we have $\sum\nolimits_{t=1}^T\Big\{f_t(\mathcal{A}lg_t(\boldsymbol{\theta}_t))-\\f_t(\mathcal{A}lg_t(\boldsymbol{\theta}^*))\Big\}\geq-FT$. Therefore,
\begin{align*}
    \sum\nolimits_{t=1}^T g_i(\mathcal{A}lg_t(\boldsymbol{\theta}_t)) \leq O(\sqrt{T\log T}), \quad \forall i\in[m]
\end{align*}
\end{proof}

%% file: App_ExpDetails.tex
\subsection{Data Pre-processing}
In order to adapt online environment, all datasets are split into a sequence of tasks. However, numbers of data samples in each task may be small. Data augmentation is therefore used on the samples in the form of rotations of random degree. Specifically, for each data sample in a task, unprotected attributes are rotated for $n$ degrees, where $n$ is randomly selected from a range of $[1,360]$. Note that, for the new rotated data sample, its label and protected feature remain the same as before. Each task is enriched for a size of at least 2500. For all datasets, all the unprotected attributes are standardized to zero mean and unit variance and prepared for experiments.

\subsection{Implementation Details and Parameter Tuning}
Our neural network trained follows the same architecture used by \cite{Finn-ICML-2017-(MAML)}, which contains 2 hidden layers of size of 40 with ReLU activation functions. In the training stage, each gradient is computed using a batch size of 200 examples where each binary class contains 100 examples. 
For each dataset, we tune the folowing hyperparameters: (1) learning rates $\eta_1,\eta_2$ for updating inner and outer parameters in Eq.(\ref{eq:task-level primal-dual})(9) and (\ref{eq:meta-level primal-dual})(12), (2) task buffer size $|\mathcal{U}|$, (3) some positive constant $\delta$ used in the augmented term in Eq.(\ref{eq:outer-Lagrangian}), (4) inner gradient steps $N_{step}$, and (5) the number of outer iterations $N_{iter}$.
% , and (6) proficiency threshold $\boldsymbol{\gamma}=(\gamma_1,\gamma_2)$ stated in Sec.\ref{sec:TaskLearningEfficiency} where each corresponds for a value of loss and \textit{DBC}. 
Hyperparameter configurations for all datasets are summarized in Table \ref{tab:hyperparameters}.
% For Bank dataset, at each round $t$, the model is trained with $5$ gradient steps with inner step size $\eta_1= 0.01$, a meta batch-size of $32$ tasks with outer step size $\eta_2=0.01$, the constant for outer augmentation $\delta=50$, and task buffer size $|\mathcal{U}|=32$. All models were trained for at least $3000$ outer iterations.
\begin{table}[!h]
    \centering
    % \vspace{-3mm}
    \caption{Hyperparameter configurations of \sysname{}.}
    % \vspace{-3mm}
    % \setlength\tabcolsep{3.7pt}
    \begin{tabular}{c|cccccc}
        \hline
         & $\eta_1$ & $\eta_2$ & $|\mathcal{U}|$ & $\delta$ & $N_{step}$ & $N_{iter}$ \\
        \hline
        Bank & 0.01 & 0.01 & 32 & 50 & 5 & 3000 \\
        \hline
        Adult & 0.001 & 0.1 & 32 & 60 & 3 & 3000 \\
        \hline
        Crime & 0.001 & 0.05 & 32 & 50 & 5 & 3500 \\
        \hline
    \end{tabular}
    \label{tab:hyperparameters}
    % \vspace{-3mm}
\end{table}
% \textcolor{red}{Efficiency details}
Initial primal meta parameters $\boldsymbol{\theta}_1$ of all baseline methods and proposed algorithms are randomly chosen, which means we train all methods starting from a random point. The initial dual meta parameters $\boldsymbol{\lambda}_1$ is chosen from $\{0.0001, 0.001, 0.01, 0.1, 1.0, 10.0, 100.0, 1000.0,\\ 10000.0\}$. Parameters $\eta_1$ and $\eta_2$ control the inner and outer learning rates are chosen from $\{0.0001, 0.0005, 0.001, 0.005, 0.01, 0.05, 0.1, 0.5, \\1.0, 5.0, 10.0, 50.0, 100.0,500.0, 1000.0\}$.

\subsection{Additional Results}
\begin{figure}[!h]
\vspace{-3mm}
% \captionsetup[subfigure]{aboveskip=-2pt,belowskip=-2pt}
\centering
    \begin{subfigure}[b]{0.235\textwidth}
        \includegraphics[width=\textwidth]{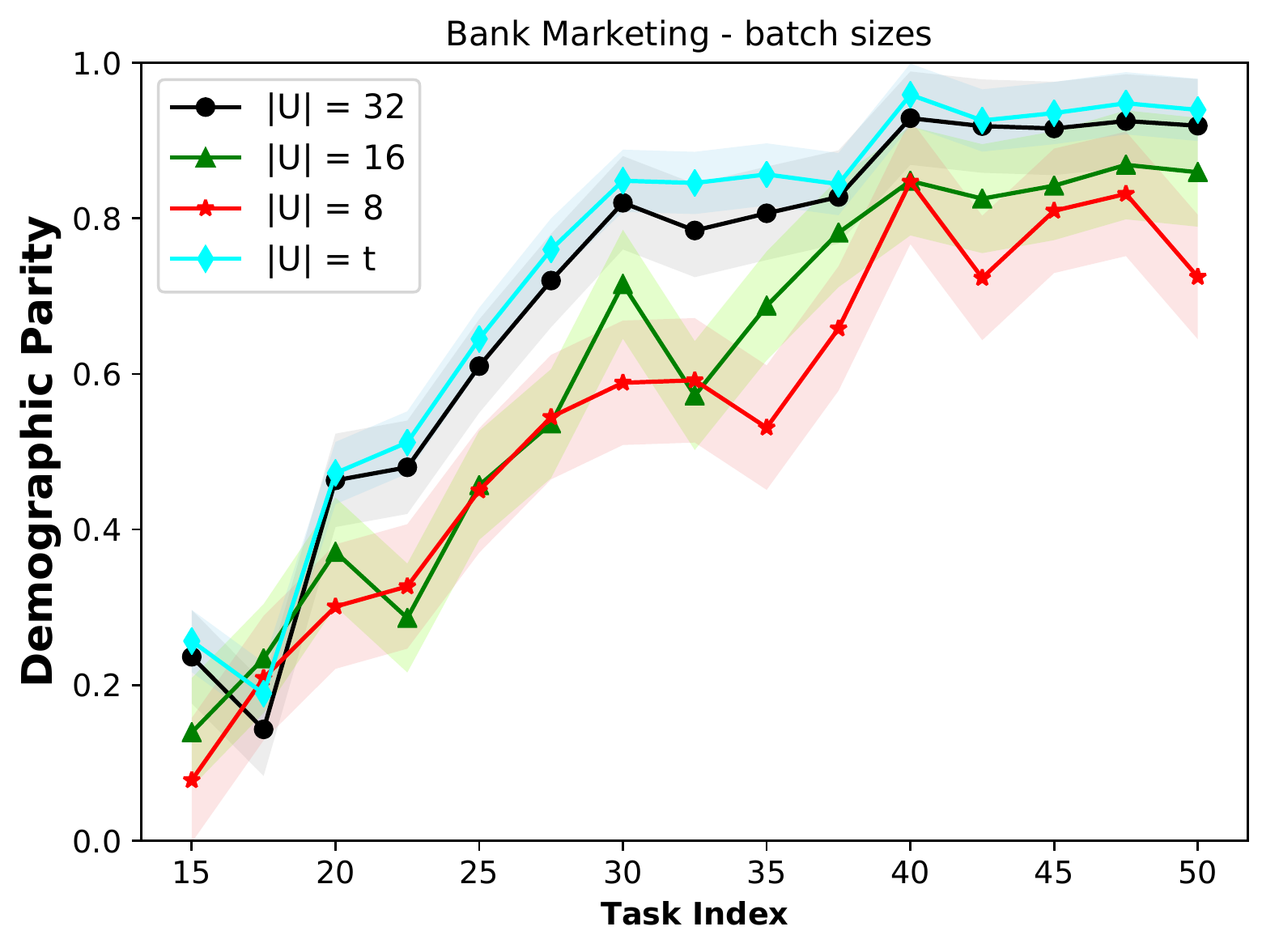}
        % \caption{}
    \end{subfigure}
    \begin{subfigure}[b]{0.235\textwidth}
        \includegraphics[width=\textwidth]{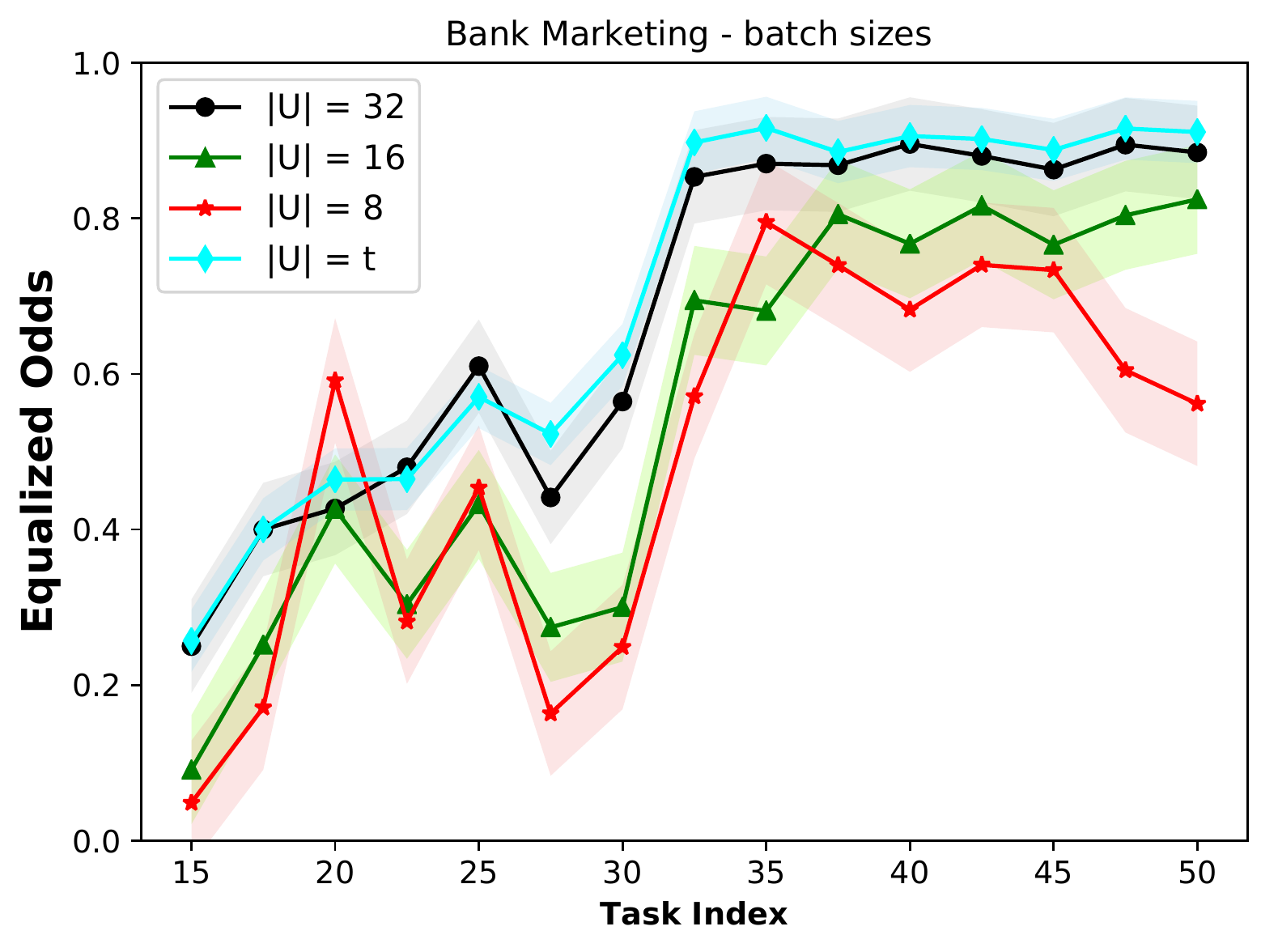}
        % \caption{}
    \end{subfigure}
    \vspace{-5mm}
    \caption{\sysname{} performance across various batch sizes.}
    \label{fig:batchsize}
% \vspace{-5mm}
\end{figure}

% Additionally, 
% in Algorithm \ref{alg:PDRFTML}, a meta-parameter pair at round $t$ is learned based on prior seen tasks. 
In order to reduce computational time in Algorithm \ref{alg:PDRFTML}, in stead of using all tasks, we approximate the results in practice with a fixed size of task buffer $\mathcal{U}$. In other words, at each round, we add the new task to $\mathcal{U}$ if $t\leq|\mathcal{U}|$. 
% The model is hence learned using all seen tasks. 
However, when $t>|\mathcal{U}|$, we stochastically sample a batch of tasks from seen tasks. 
% In our experiments, we set $|\mathcal{U}|=32$ for all experiments and datasets. 
% To validate working efficiency for our algorithm, 
Figure \ref{fig:batchsize} represents results based on the \textit{Bank} dataset with various batch sizes. 
% For other dataset, we defer additional results to Appendix \ref{App:ExpDetails}. 
Our empirical results indicate that although better performance is able to achieved with higher batch size, on the contrary more expensive the experiments will be.

\begin{figure}[!h]
\vspace{-3mm}
\captionsetup[subfigure]{aboveskip=-2pt,belowskip=-2pt}
\centering
    \begin{subfigure}[b]{0.235\textwidth}
        \includegraphics[width=\textwidth]{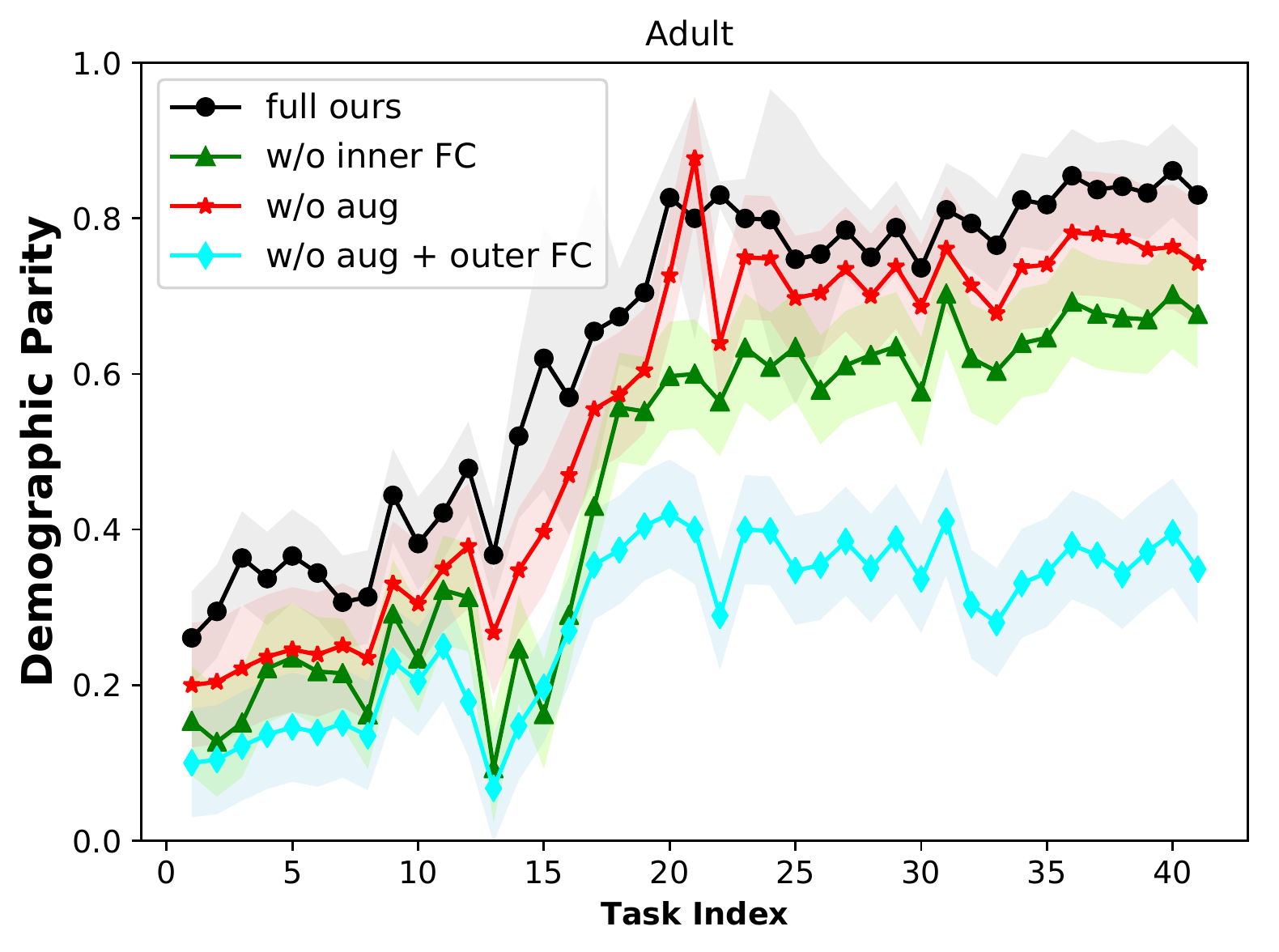}
        % \caption{}
    \end{subfigure}
    \begin{subfigure}[b]{0.235\textwidth}
        \includegraphics[width=\textwidth]{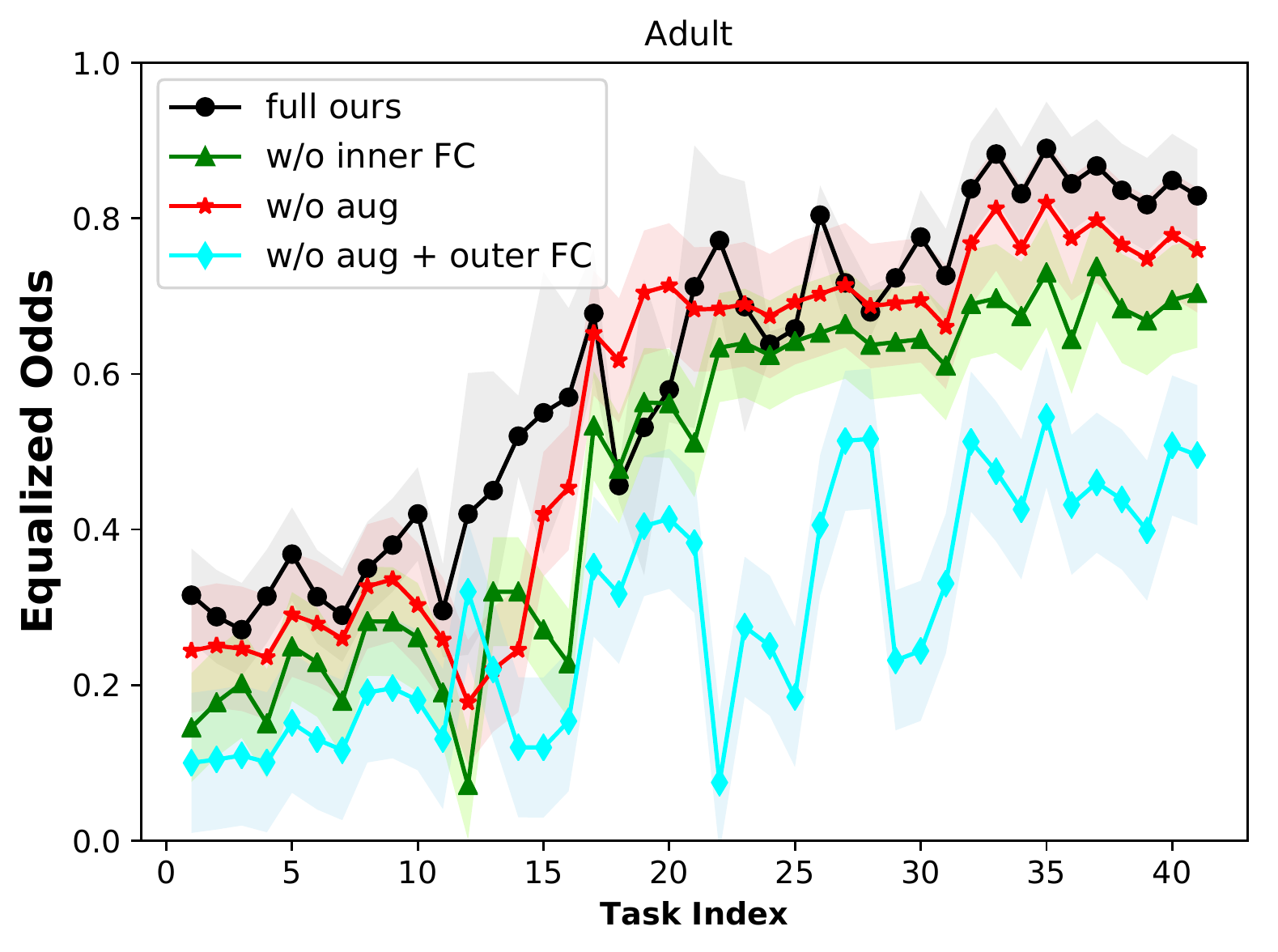}
        % \caption{}
    \end{subfigure}
    
    \begin{subfigure}[b]{0.235\textwidth}
        \includegraphics[width=\textwidth]{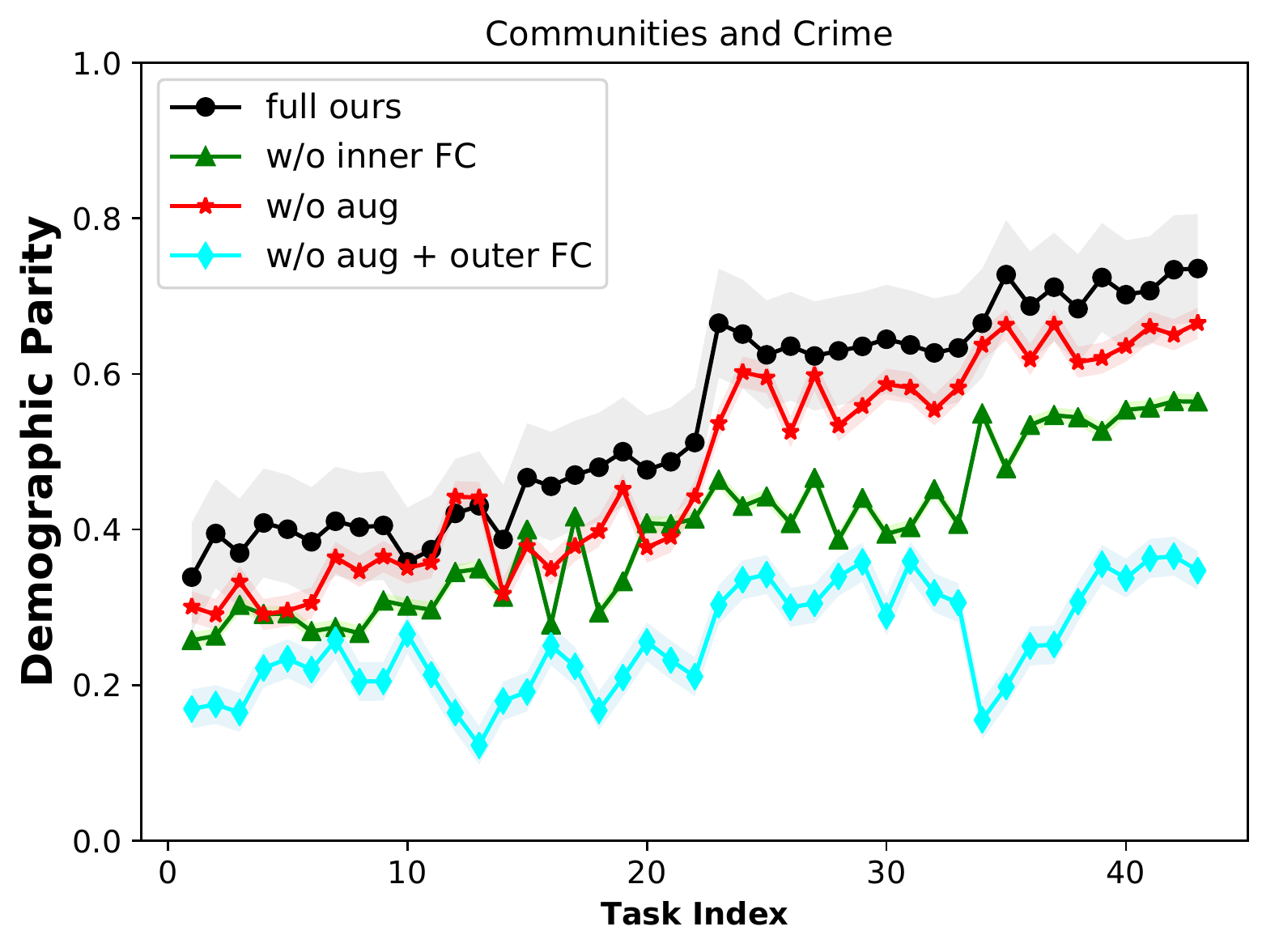}
        % \caption{}
    \end{subfigure}
    \begin{subfigure}[b]{0.235\textwidth}
        \includegraphics[width=\textwidth]{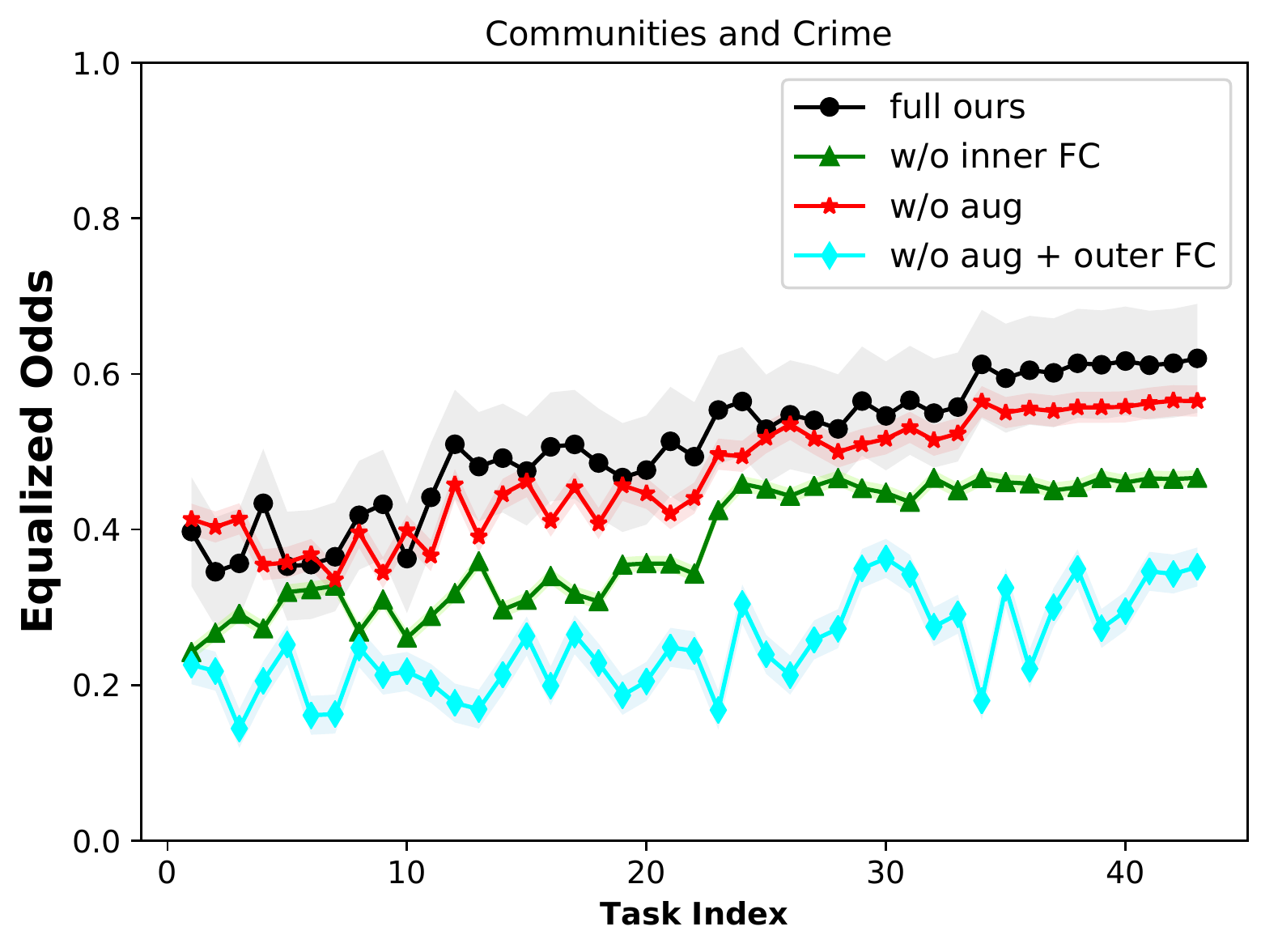}
        % \caption{}
    \end{subfigure}
    \vspace{-8mm}
    \caption{Ablation study of \sysname{} on Adult and Crime datasets.}
    \label{fig:as_adult_crime}
% \vspace{-5mm}
\end{figure}

More ablation study results on Adult and Crime datasets are given in Figure \ref{fig:as_adult_crime}.

%% file: main.bbl

\begin{thebibliography}{34}

%%% the Web address.

\ifx \showCODEN    \undefined \def \showCODEN     #1{\unskip}     \fi
\ifx \showDOI      \undefined \def \showDOI       #1{#1}\fi
\ifx \showISBNx    \undefined \def \showISBNx     #1{\unskip}     \fi
\ifx \showISBNxiii \undefined \def \showISBNxiii  #1{\unskip}     \fi
\ifx \showISSN     \undefined \def \showISSN      #1{\unskip}     \fi
\ifx \showLCCN     \undefined \def \showLCCN      #1{\unskip}     \fi
\ifx \shownote     \undefined \def \shownote      #1{#1}          \fi
\ifx \showarticletitle \undefined \def \showarticletitle #1{#1}   \fi
\ifx \showURL      \undefined \def \showURL       {\relax}        \fi
% invisible to TeX
\providecommand\bibfield[2]{#2}
\providecommand\bibinfo[2]{#2}
\providecommand\natexlab[1]{#1}
\providecommand\showeprint[2][]{arXiv:#2}

\bibitem[\protect\citeauthoryear{Bechavod, Ligett, Roth, Waggoner, and
  Wu}{Bechavod et~al\mbox{.}}{2019}]%
        {Bechavod-2019-NeurIPS}
\bibfield{author}{\bibinfo{person}{Yahav Bechavod}, \bibinfo{person}{Katrina
  Ligett}, \bibinfo{person}{Aaron Roth}, \bibinfo{person}{Bo Waggoner}, {and}
  \bibinfo{person}{Zhiwei Wu}.} \bibinfo{year}{2019}\natexlab{}.
\newblock \showarticletitle{Equal Opportunity in Online Classification with
  Partial Feedback}.
\newblock \bibinfo{journal}{\emph{NeurIPS}} (\bibinfo{date}{02}
  \bibinfo{year}{2019}).
\newblock


\bibitem[\protect\citeauthoryear{Calders, Kamiran, and Pechenizkiy}{Calders
  et~al\mbox{.}}{2009}]%
        {calders-2009-icdmw}
\bibfield{author}{\bibinfo{person}{Toon Calders}, \bibinfo{person}{Faisal
  Kamiran}, {and} \bibinfo{person}{Mykola Pechenizkiy}.}
  \bibinfo{year}{2009}\natexlab{}.
\newblock \showarticletitle{Building Classifiers with Independency
  Constraints}. In \bibinfo{booktitle}{\emph{2009 IEEE International Conference
  on Data Mining Workshops}}. \bibinfo{pages}{13--18}.
\newblock
\urldef\tempurl%
\url{https://doi.org/10.1109/ICDMW.2009.83}
\showDOI{\tempurl}


\bibitem[\protect\citeauthoryear{Cesa-Bianchi and Lugosi}{Cesa-Bianchi and
  Lugosi}{2006}]%
        {Cambridge-book-2006}
\bibfield{author}{\bibinfo{person}{Nicolò Cesa-Bianchi} {and}
  \bibinfo{person}{Gábor Lugosi}.} \bibinfo{year}{2006}\natexlab{}.
\newblock \showarticletitle{Prediction, Learning, and Games}.
\newblock \bibinfo{journal}{\emph{Cambridge University Press}}
  (\bibinfo{year}{2006}).
\newblock


\bibitem[\protect\citeauthoryear{Dwork, Hardt, Pitassi, Reingold, and
  Zemel}{Dwork et~al\mbox{.}}{2011}]%
        {Dwork-2011-CoRR}
\bibfield{author}{\bibinfo{person}{Cynthia Dwork}, \bibinfo{person}{Moritz
  Hardt}, \bibinfo{person}{Toniann Pitassi}, \bibinfo{person}{Omer Reingold},
  {and} \bibinfo{person}{Rich Zemel}.} \bibinfo{year}{2011}\natexlab{}.
\newblock \showarticletitle{Fairness Through Awareness}.
\newblock \bibinfo{journal}{\emph{CoRR}} (\bibinfo{year}{2011}).
\newblock


\bibitem[\protect\citeauthoryear{Finn, Abbeel, and Levine}{Finn
  et~al\mbox{.}}{2017}]%
        {Finn-ICML-2017-(MAML)}
\bibfield{author}{\bibinfo{person}{Chelsea Finn}, \bibinfo{person}{Pieter
  Abbeel}, {and} \bibinfo{person}{Sergey Levine}.}
  \bibinfo{year}{2017}\natexlab{}.
\newblock \showarticletitle{Model-Agnostic Meta-Learning for Fast Adaptation of
  Deep Networks}.
\newblock \bibinfo{journal}{\emph{ICML}} (\bibinfo{year}{2017}).
\newblock


\bibitem[\protect\citeauthoryear{Finn, Rajeswaran, Kakade, and Levine}{Finn
  et~al\mbox{.}}{2019}]%
        {Finn-ICML-2019}
\bibfield{author}{\bibinfo{person}{Chelsea Finn}, \bibinfo{person}{Aravind
  Rajeswaran}, \bibinfo{person}{Sham Kakade}, {and} \bibinfo{person}{Sergey
  Levine}.} \bibinfo{year}{2019}\natexlab{}.
\newblock \showarticletitle{Online Meta-Learning.}
\newblock \bibinfo{journal}{\emph{ICML}} (\bibinfo{year}{2019}).
\newblock


\bibitem[\protect\citeauthoryear{Finn, Xu, and Levine}{Finn
  et~al\mbox{.}}{2018}]%
        {Finn-NIPS-2018}
\bibfield{author}{\bibinfo{person}{Chelsea Finn}, \bibinfo{person}{Kelvin Xu},
  {and} \bibinfo{person}{Sergey Levine}.} \bibinfo{year}{2018}\natexlab{}.
\newblock \showarticletitle{Probabilistic model-agnostic meta-learning}. In
  \bibinfo{booktitle}{\emph{NeurIPS}}. \bibinfo{pages}{9516--9527}.
\newblock


\bibitem[\protect\citeauthoryear{Gillen, Jung, Kearns, and Roth}{Gillen
  et~al\mbox{.}}{2018}]%
        {Stephen-2018-NeurIPS}
\bibfield{author}{\bibinfo{person}{Stephen Gillen},
  \bibinfo{person}{Christopher Jung}, \bibinfo{person}{Michael Kearns}, {and}
  \bibinfo{person}{Aaron Roth}.} \bibinfo{year}{2018}\natexlab{}.
\newblock \showarticletitle{Online Learning with an Unknown Fairness Metric}.
\newblock \bibinfo{journal}{\emph{NeurIPS}} (\bibinfo{year}{2018}).
\newblock


\bibitem[\protect\citeauthoryear{Hannan}{Hannan}{1957}]%
        {Hannan-1957-online}
\bibfield{author}{\bibinfo{person}{James Hannan}.}
  \bibinfo{year}{1957}\natexlab{}.
\newblock \showarticletitle{Approximation to bayes risk in repeated play.}
\newblock \bibinfo{journal}{\emph{Contributions to the Theory of Games}}
  (\bibinfo{year}{1957}).
\newblock


\bibitem[\protect\citeauthoryear{Hardt, Price, and Srebro}{Hardt
  et~al\mbox{.}}{2016}]%
        {Hardt-NIPS-2016}
\bibfield{author}{\bibinfo{person}{Moritz Hardt}, \bibinfo{person}{Eric Price},
  {and} \bibinfo{person}{Nathan Srebro}.} \bibinfo{year}{2016}\natexlab{}.
\newblock \showarticletitle{Equality of opportunity in supervised learning.}
\newblock \bibinfo{journal}{\emph{NeurIPS}} (\bibinfo{year}{2016}).
\newblock


\bibitem[\protect\citeauthoryear{Hoi, Sahoo, Lu, and Zhao}{Hoi
  et~al\mbox{.}}{2018}]%
        {Hoi-2018-survey}
\bibfield{author}{\bibinfo{person}{Steven Hoi}, \bibinfo{person}{Doyen Sahoo},
  \bibinfo{person}{Jing Lu}, {and} \bibinfo{person}{Peilin Zhao}.}
  \bibinfo{year}{2018}\natexlab{}.
\newblock \showarticletitle{Online Learning: A Comprehensive Survey}.
\newblock \bibinfo{journal}{\emph{arXiv:1802.02871 [cs.LG]}}
  (\bibinfo{date}{02} \bibinfo{year}{2018}).
\newblock


\bibitem[\protect\citeauthoryear{Jenatton, Huang, and Archambeau}{Jenatton
  et~al\mbox{.}}{2016}]%
        {AdpOLC-2016-ICML}
\bibfield{author}{\bibinfo{person}{Rodolphe Jenatton}, \bibinfo{person}{Jim
  Huang}, {and} \bibinfo{person}{Cedric Archambeau}.}
  \bibinfo{year}{2016}\natexlab{}.
\newblock \showarticletitle{Adaptive Algorithms for Online Convex Optimization
  with Long-term Constraints}.
\newblock \bibinfo{journal}{\emph{ICML}} (\bibinfo{year}{2016}).
\newblock


\bibitem[\protect\citeauthoryear{Kohavi and Becker}{Kohavi and Becker}{1994}]%
        {AdultDataSet-UCI-1994}
\bibfield{author}{\bibinfo{person}{Ronny Kohavi} {and} \bibinfo{person}{Barry
  Becker}.} \bibinfo{year}{1994}\natexlab{}.
\newblock \showarticletitle{UCI Machine Learning Repository}.
\newblock  (\bibinfo{year}{1994}).
\newblock


\bibitem[\protect\citeauthoryear{Lahoti, Weikum, and Gummadi}{Lahoti
  et~al\mbox{.}}{2019}]%
        {Lahoti-2019-ICDE}
\bibfield{author}{\bibinfo{person}{Preethi Lahoti}, \bibinfo{person}{Gerhard
  Weikum}, {and} \bibinfo{person}{Krishna Gummadi}.}
  \bibinfo{year}{2019}\natexlab{}.
\newblock \showarticletitle{iFair: Learning Individually Fair Data
  Representations for Algorithmic Decision Making}.
\newblock \bibinfo{journal}{\emph{ICDE}}.
\newblock


\bibitem[\protect\citeauthoryear{Lichman}{Lichman}{2013}]%
        {CommunitiesandCrimeDataSet-UCI-1994}
\bibfield{author}{\bibinfo{person}{Moshe Lichman}.}
  \bibinfo{year}{2013}\natexlab{}.
\newblock \showarticletitle{UCI Machine Learning Repository}.
\newblock  (\bibinfo{year}{2013}).
\newblock


\bibitem[\protect\citeauthoryear{Lohaus, Perrot, and Luxburg}{Lohaus
  et~al\mbox{.}}{2020}]%
        {Lohaus-2020-ICML}
\bibfield{author}{\bibinfo{person}{Michael Lohaus}, \bibinfo{person}{Michael
  Perrot}, {and} \bibinfo{person}{Ulrike~Von Luxburg}.}
  \bibinfo{year}{2020}\natexlab{}.
\newblock \showarticletitle{Too Relaxed to Be Fair}. In
  \bibinfo{booktitle}{\emph{ICML}}.
\newblock


\bibitem[\protect\citeauthoryear{Mahdavi, Jin, and Yang}{Mahdavi
  et~al\mbox{.}}{2012}]%
        {OGDLC-2012-JMLR}
\bibfield{author}{\bibinfo{person}{Mehrdad Mahdavi}, \bibinfo{person}{Rong
  Jin}, {and} \bibinfo{person}{Tianbao Yang}.} \bibinfo{year}{2012}\natexlab{}.
\newblock \showarticletitle{Trading regret for efficiency: online convex
  optimization with long term constraints}.
\newblock \bibinfo{journal}{\emph{JMLR}} (\bibinfo{year}{2012}).
\newblock


\bibitem[\protect\citeauthoryear{Miller}{Miller}{2020}]%
        {Miller-2020-NYTimes}
\bibfield{author}{\bibinfo{person}{Jennifer Miller}.}
  \bibinfo{year}{2020}\natexlab{}.
\newblock \showarticletitle{Is an Algorithm Less Racist Than a Loan Officer?}
\newblock
  \bibinfo{journal}{\emph{www.nytimes.com/2020/09/18/business/digital-mortgages.html}}
  (\bibinfo{year}{2020}).
\newblock


\bibitem[\protect\citeauthoryear{Moro, Cortez, and Rita}{Moro
  et~al\mbox{.}}{2014}]%
        {Moro-bank-marketing-dataset-2014}
\bibfield{author}{\bibinfo{person}{Sérgio Moro}, \bibinfo{person}{Paulo
  Cortez}, {and} \bibinfo{person}{Paulo Rita}.}
  \bibinfo{year}{2014}\natexlab{}.
\newblock \showarticletitle{A data-driven approach to predict the success of
  bank telemarketing.}
\newblock \bibinfo{journal}{\emph{DSS}} (\bibinfo{year}{2014}).
\newblock


\bibitem[\protect\citeauthoryear{Nagabandi, Finn, and Levine}{Nagabandi
  et~al\mbox{.}}{2019}]%
        {Anusha-2019-ICLR}
\bibfield{author}{\bibinfo{person}{Anusha Nagabandi}, \bibinfo{person}{Chelsea
  Finn}, {and} \bibinfo{person}{Sergey Levine}.}
  \bibinfo{year}{2019}\natexlab{}.
\newblock \showarticletitle{Deep Online Learning via Meta-Learning: Continual
  Adaptation for Model-Based RL}.
\newblock \bibinfo{journal}{\emph{ICLR}} (\bibinfo{year}{2019}).
\newblock


\bibitem[\protect\citeauthoryear{Patil, Ghalme, Nair, and Narahari}{Patil
  et~al\mbox{.}}{2020}]%
        {Vishakha-2020-AAAI}
\bibfield{author}{\bibinfo{person}{Vishakha Patil}, \bibinfo{person}{Ganesh
  Ghalme}, \bibinfo{person}{Vineet Nair}, {and} \bibinfo{person}{Yadati
  Narahari}.} \bibinfo{year}{2020}\natexlab{}.
\newblock \showarticletitle{Achieving Fairness in the Stochastic Multi-Armed
  Bandit Problem}.
\newblock \bibinfo{journal}{\emph{AAAI}} (\bibinfo{year}{2020}).
\newblock


\bibitem[\protect\citeauthoryear{Rusu, Rao, Sygnowski, Vinyals, Pascanu,
  Osindero, and Hadsell}{Rusu et~al\mbox{.}}{2019}]%
        {Rusu-ICLR-2019}
\bibfield{author}{\bibinfo{person}{Andrei~A Rusu}, \bibinfo{person}{Dushyant
  Rao}, \bibinfo{person}{Jakub Sygnowski}, \bibinfo{person}{Oriol Vinyals},
  \bibinfo{person}{Razvan Pascanu}, \bibinfo{person}{Simon Osindero}, {and}
  \bibinfo{person}{Raia Hadsell}.} \bibinfo{year}{2019}\natexlab{}.
\newblock \showarticletitle{Meta-learning with latent embedding optimization}.
\newblock \bibinfo{journal}{\emph{ICLR}} (\bibinfo{year}{2019}).
\newblock


\bibitem[\protect\citeauthoryear{Schmidhuber}{Schmidhuber}{1987}]%
        {schmidhuber-1987-srl}
\bibfield{author}{\bibinfo{person}{Jurgen Schmidhuber}.}
  \bibinfo{year}{1987}\natexlab{}.
\newblock \showarticletitle{Evolutionary Principles in Self-Referential
  Learning}.
\newblock  (\bibinfo{year}{1987}).
\newblock


\bibitem[\protect\citeauthoryear{Slack, Friedler, and Givental}{Slack
  et~al\mbox{.}}{2020}]%
        {Slack-FAT-2019}
\bibfield{author}{\bibinfo{person}{Dylan Slack}, \bibinfo{person}{Sorelle
  Friedler}, {and} \bibinfo{person}{Emile Givental}.}
  \bibinfo{year}{2020}\natexlab{}.
\newblock \showarticletitle{Fairness Warnings and Fair-MAML: Learning Fairly
  with Minimal Data}.
\newblock \bibinfo{journal}{\emph{ACM FAccT}} (\bibinfo{year}{2020}).
\newblock


\bibitem[\protect\citeauthoryear{Wang, Chen, Zhao, Lin, Zhao, Tao, Wang, and
  Khan}{Wang et~al\mbox{.}}{2021}]%
        {wang-2021-WWW}
\bibfield{author}{\bibinfo{person}{Zhuoyi Wang}, \bibinfo{person}{Yuqiao Chen},
  \bibinfo{person}{Chen Zhao}, \bibinfo{person}{Yu Lin},
  \bibinfo{person}{Xujiang Zhao}, \bibinfo{person}{Hemeng Tao},
  \bibinfo{person}{Yigong Wang}, {and} \bibinfo{person}{Latifur Khan}.}
  \bibinfo{year}{2021}\natexlab{}.
\newblock \showarticletitle{CLEAR: Contrastive-Prototype Learning with Drift
  Estimation for Resource Constrained Stream Mining}. In
  \bibinfo{booktitle}{\emph{WWW}}.
\newblock


\bibitem[\protect\citeauthoryear{Yao, Zhou, Mahdavi, Li, Socher, and Xiong}{Yao
  et~al\mbox{.}}{2020}]%
        {Yao-2020-NeurIPS}
\bibfield{author}{\bibinfo{person}{Huaxiu Yao}, \bibinfo{person}{Yingbo Zhou},
  \bibinfo{person}{Mehrdad Mahdavi}, \bibinfo{person}{Zhenhui Li},
  \bibinfo{person}{Richard Socher}, {and} \bibinfo{person}{Caiming Xiong}.}
  \bibinfo{year}{2020}\natexlab{}.
\newblock \showarticletitle{Online Structured Meta-learning}.
\newblock \bibinfo{journal}{\emph{NeurIPS}} (\bibinfo{year}{2020}).
\newblock


\bibitem[\protect\citeauthoryear{Yu, Neely, and Wei}{Yu et~al\mbox{.}}{2017}]%
        {Yu-2017-NIPS}
\bibfield{author}{\bibinfo{person}{Hao Yu}, \bibinfo{person}{Michael~J. Neely},
  {and} \bibinfo{person}{Xiaohan Wei}.} \bibinfo{year}{2017}\natexlab{}.
\newblock \showarticletitle{Online Convex Optimization with Stochastic
  Constraints}. In \bibinfo{booktitle}{\emph{NeurIPS}}.
\newblock


\bibitem[\protect\citeauthoryear{Yuan and Lamperski}{Yuan and
  Lamperski}{2018}]%
        {GenOLC-2018-NeurIPS}
\bibfield{author}{\bibinfo{person}{Jianjun Yuan} {and} \bibinfo{person}{Andrew
  Lamperski}.} \bibinfo{year}{2018}\natexlab{}.
\newblock \showarticletitle{Online convex optimization for cumulative
  constraints}.
\newblock \bibinfo{journal}{\emph{NeurIPS}} (\bibinfo{year}{2018}).
\newblock


\bibitem[\protect\citeauthoryear{Zafar, Valera, Rodriguez, and Gummadi}{Zafar
  et~al\mbox{.}}{2017}]%
        {Zafar-AISTATS-2017}
\bibfield{author}{\bibinfo{person}{Muhammad~Bilal Zafar},
  \bibinfo{person}{Isabel Valera}, \bibinfo{person}{Manuel~Gomez Rodriguez},
  {and} \bibinfo{person}{Krishna~P. Gummadi}.} \bibinfo{year}{2017}\natexlab{}.
\newblock \showarticletitle{Fairness Constraints: Mechanisms for Fair
  Classification}.
\newblock \bibinfo{journal}{\emph{AISTATS}} (\bibinfo{year}{2017}).
\newblock


\bibitem[\protect\citeauthoryear{Zemel, Wu, Swersky, Pitassi, and Dwork}{Zemel
  et~al\mbox{.}}{2013}]%
        {Zemel-ICML-2013}
\bibfield{author}{\bibinfo{person}{Richard Zemel}, \bibinfo{person}{Yu Wu},
  \bibinfo{person}{Kevin Swersky}, \bibinfo{person}{Toniann Pitassi}, {and}
  \bibinfo{person}{Cynthia Dwork}.} \bibinfo{year}{2013}\natexlab{}.
\newblock \showarticletitle{Learning Fair Representations}.
\newblock \bibinfo{journal}{\emph{ICML}} (\bibinfo{year}{2013}).
\newblock


\bibitem[\protect\citeauthoryear{Zhao and Chen}{Zhao and Chen}{2019}]%
        {Zhao-ICDM-2019}
\bibfield{author}{\bibinfo{person}{Chen Zhao} {and} \bibinfo{person}{Feng
  Chen}.} \bibinfo{year}{2019}\natexlab{}.
\newblock \showarticletitle{Rank-Based Multi-task Learning For Fair
  Regression}.
\newblock \bibinfo{journal}{\emph{IEEE International Conference on Data Mining
  (ICDM)}} (\bibinfo{year}{2019}).
\newblock


\bibitem[\protect\citeauthoryear{Zhao and Chen}{Zhao and Chen}{2020}]%
        {Zhao-ICKG-1-2020}
\bibfield{author}{\bibinfo{person}{Chen Zhao} {and} \bibinfo{person}{Feng
  Chen}.} \bibinfo{year}{2020}\natexlab{}.
\newblock \showarticletitle{Unfairness Discovery and Prevention For Few-Shot
  Regression.}
\newblock \bibinfo{journal}{\emph{ICKG}} (\bibinfo{year}{2020}).
\newblock


\bibitem[\protect\citeauthoryear{Zhao, Chen, Wang, and Khan}{Zhao
  et~al\mbox{.}}{2020a}]%
        {zhao-2020-pdfm}
\bibfield{author}{\bibinfo{person}{Chen Zhao}, \bibinfo{person}{Feng Chen},
  \bibinfo{person}{Zhuoyi Wang}, {and} \bibinfo{person}{Latifur Khan}.}
  \bibinfo{year}{2020}\natexlab{a}.
\newblock \showarticletitle{A Primal-Dual Subgradient Approach for Fair Meta
  Learning}. In \bibinfo{booktitle}{\emph{ICDM}}.
\newblock


\bibitem[\protect\citeauthoryear{Zhao, Li, Li, and Chen}{Zhao
  et~al\mbox{.}}{2020b}]%
        {Zhao-ICKG-2-2020}
\bibfield{author}{\bibinfo{person}{Chen Zhao}, \bibinfo{person}{Changbin Li},
  \bibinfo{person}{Jincheng Li}, {and} \bibinfo{person}{Feng Chen}.}
  \bibinfo{year}{2020}\natexlab{b}.
\newblock \showarticletitle{Fair Meta-Learning For Few-Shot Classification.}
\newblock \bibinfo{journal}{\emph{ICKG}} (\bibinfo{year}{2020}).
\newblock


\end{thebibliography}
